\documentclass{article}

    \PassOptionsToPackage{numbers, compress}{natbib}

\usepackage[final]{neurips_2021}

\usepackage[utf8]{inputenc}             
\usepackage[T1]{fontenc}                
\usepackage{hyperref}                   
\usepackage{url}                        
\usepackage{booktabs}                   
\usepackage{amsfonts}                   
\usepackage{nicefrac}                   
\usepackage{microtype}                  
\usepackage[dvipsnames]{xcolor}         
\usepackage{graphicx}
\usepackage{subcaption}
\usepackage{amsmath}
\usepackage{algorithm}
\usepackage{algorithmic}
\usepackage{comment}
\usepackage{upquote,textcomp}
\usepackage{listings}
\usepackage[most]{tcolorbox}

\newtcbox{\dashedbox}[1][]{
  math upper,
  baseline=0.4\baselineskip,
  equal height group=dashedbox,
  nobeforeafter,
  colback=white,
  fontupper= \color{white},
  boxrule=0pt,
  enhanced jigsaw,
  boxsep=0pt,
  top=2pt,
  bottom=2pt,
  left=2pt,
  right=2pt,
  borderline horizontal={0.5pt}{0pt}{dashed},
  borderline vertical={0.5pt}{0pt}{dashed},
  #1
}

\title{Predify: Augmenting deep neural networks with brain-inspired predictive coding dynamics}

\author{%
  Bhavin~Choksi\thanks{Equal Contribution}  \\
  CerCo~CNRS, UMR~5549 \& \\ 
  Universit\'e de Toulouse\\
  \texttt{bhavin.choksi@cnrs.fr} \\

   \And
   Milad~Mozafari\textsuperscript{*} \\
  CerCo~CNRS, UMR~5549 \& \\
  IRIT~CNRS, UMR~5505 \\
  \texttt{milad.mozafari@cnrs.fr} \\
   \And
   Callum~Biggs O'May \\
  CerCo~CNRS\\
  UMR~5549 \\

   \AND
   Benjamin Ador \\
  CerCo~CNRS\\
  UMR~5549 \\

   \And
   Andrea~Alamia \\
  CerCo~CNRS\\
  UMR~5549 \\

   \And
   Rufin~VanRullen\\
  CerCo~CNRS, UMR~5549 \& \\
  ANITI, Universit\'e de Toulouse  \\
  \texttt{rufin.vanrullen@cnrs.fr} \\
}

\begin{document}

\maketitle

\begin{abstract}

Deep neural networks excel at image classification, but their performance is far less robust to input perturbations than human perception. In this work we explore whether this shortcoming may be partly addressed by incorporating brain-inspired recurrent dynamics in deep convolutional networks. We take inspiration from a popular framework in neuroscience: ``predictive coding''.~At each layer of the hierarchical model, generative feedback ``predicts'' (i.e., reconstructs) the pattern of activity in the previous layer. The reconstruction errors are used to iteratively update the network’s representations across timesteps, and to optimize the network's feedback weights over the natural image dataset--a form of unsupervised training. We show that implementing this strategy into two popular networks, VGG16 and EfficientNetB0, improves their robustness against various corruptions and adversarial attacks. We hypothesize that other feedforward networks could similarly benefit from the proposed framework. To promote research in this direction, we provide an open-sourced PyTorch-based package called \textit{Predify}, which can be used to implement and investigate the impacts of the predictive coding dynamics in any convolutional neural network. 

\end{abstract}

\section{Introduction}
Deep convolutional neural networks (DCNNs), initially inspired by the primate visual cortex architecture, have taken big strides in solving computer vision tasks in the last decade. State-of-the-art networks can learn to classify images with high accuracy from huge labeled datasets~\cite{krizhevsky2012imagenet,simonyan2014very,he2015deep,huang2016densely,al2021reconstructing,tan2019efficientnet}. This rapid progress and the resulting interest in these techniques have also highlighted their various shortcomings. Most widely studied is the sensitivity of neural networks, not only to perturbations specifically designed to fool them (so-called ``adversarial examples'') but also to regular noises typically observed in natural scenes~\cite{szegedy2013intriguing,hendrycks2019benchmarking,nguyen2014deep}. These shortcomings indicate that there is still room for improvement in current techniques.

One possible way to improve the robustness of artificial neural networks could be to take further inspiration from the brain. In particular, one major aspect of the cerebral cortex that is missing from standard feedforward DCNNs is the presence of feedback connections. Recent studies have stressed the importance of feedback connections in the brain~\cite{kietzmann2019recurrence,kar2019evidence}, and have shown how artificial neural networks can take advantage of such feedback for various tasks such as object recognition with occlusion~\cite{ernst2019recurrent}, or panoptic segmentation~\cite{linsley2020stable}. Feedback connections convey contextual information about the state of the higher layers down to the lower layers of the hierarchy; in this way, they can constrain lower layers to represent inputs in meaningful ways. In theory, this could make neural representations more robust to image degradation~\cite{wyatte2012limits}. Merely including feedback in the pattern of connections, however, may not always be sufficient; rather, it should be combined with proper mechanistic principles.

To that end, we explore the potential of recurrent dynamics for augmenting deep neural networks with brain-inspired predictive coding (supported by ample neuroscience evidence~\cite{bastos2012canonical,huang2011predictive,heilbron2018great,aitchison2017or,walsh2020evaluating}). We build large-scale hierarchical networks with both feedforward and feedback connections that can be trained using error backpropagation. Several prior studies have explored this interesting avenue of research~\cite{chalasani2013deep,lotter2016deep,wen2018deep,boutin2019sparse}, but with important differences with our approach (see Section~\ref{prior_work}). We demonstrate that our proposed method adds desirable properties to feedforward DCNNs, especially when viewed from the perspective of robustness. 

Our contributions can be summarized as follows:
\begin{itemize}
    \item We propose a novel strategy for effectively incorporating recurrent feedback connections based on the neuroscientific principle of predictive coding.
    \item We implement this strategy in two pre-trained feedforward architectures with unsupervised training of the feedback weights, and show that this improves their robustness against different types of natural and adversarial noise.
    \item We suggest and verify that an emergent property of the network is to iteratively shift noisy representations towards the corresponding clean representations---a form of ``projection towards the learned manifold'' as implemented in certain adversarial defense methods.
    \item To facilitate research aimed at using such neuroscientific principles in machine learning, we provide a Python package called \textit{Predify} that can easily implement the proposed predictive coding dynamics in any convolutional neural network with a few lines of code.
\end{itemize}

\begin{figure}[t]
    \centering
    \includegraphics[width=\textwidth]{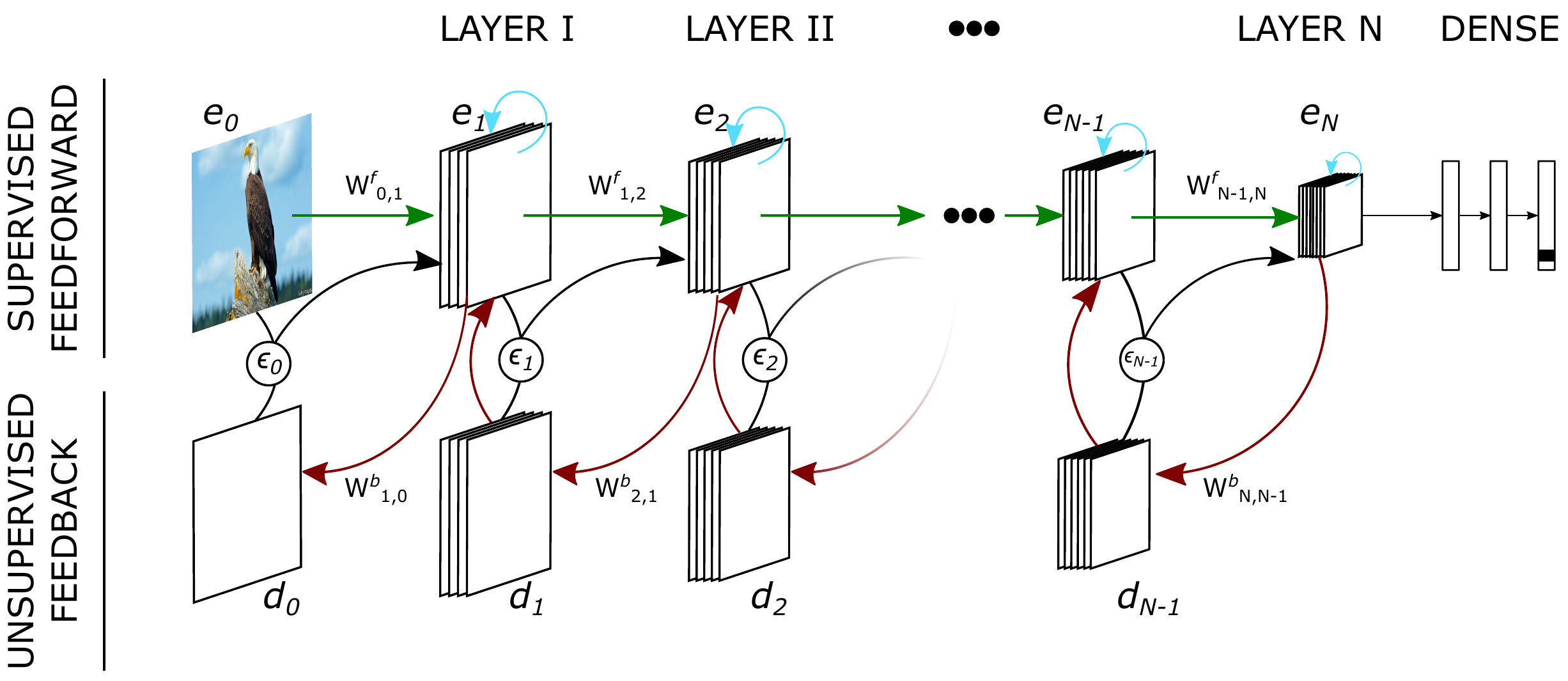}
    \caption{\textbf{General overview of our predictive coding strategy} as implemented in a feedforward hierarchical network with generative feedback connections. The architecture (roughly similar to stacked auto-encoders) consists of $N$ encoding layers $e_{n}$ and $N$ decoding layers $d_{n}$. $W_{m,n}$ denotes the connection weights from layer $m$ to layer $n$, with $W^{f}$ and $W^{b}$ for feedforward and feedback connections, respectively. The reconstruction errors at each layer are denoted by $\epsilon_{n}$. The feedforward connections (green arrows) are trained for image classification (in a supervised fashion), while the feedback weights (red arrows) are optimized for a prediction (i.e. reconstruction) objective (unsupervised). Predictive coding minimizes the reconstruction errors in each layer by updating activations in the next layer accordingly (black arrows). Self-connections (memory) are represented by blue arrows.}
    \label{fig:pc_schematic}
\end{figure}

\section{Our Approach}
\label{our_approach}
\subsection{The proposed predictive coding dynamics}

Predictive coding, as introduced by~\cite{rao1999predictive}, is a neurocomputational theory positing that the brain maintains an internal model of the world, which it uses to actively predict the observed stimulus. Within a hierarchical architecture, each higher layer attempts to predict the activity of the layer immediately below, and the errors made in this prediction are then utilized to correct the higher-layer activity.

To establish our notation, let us consider a hierarchical feedforward network equipped with generative feedback connections, as represented in Figure~\ref{fig:pc_schematic}. The network contains $N$ encoding layers $e_{n}$ ($n \in \mathbb{N}$) and $N$ corresponding decoding layers $d_{n-1}$. The feedforward weights connecting layer $n-1$ to layer $n$ are denoted by $W^{f}_{n-1,n}$, and the feedback weights from layer $n+1$ to $n$ by $W^{b}_{n+1,n}$. For a given input image, we first initiate the activations of all encoding layers with a feedforward pass. Then, over successive recurrent iterations (referred to as timesteps $t$), both the decoding and encoding layer representations are updated using the following equations (also refer to Pseudocode~\ref{pseudocode}): 

\begin{equation}
    \boldsymbol{d}_{n}(t) =  W^{b}_{n+1,n}\boldsymbol{e}_{n+1}(t) 
\end{equation}
\begin{equation}
\label{eq:pc_equation}
    \boldsymbol{e}_{n}(t+1) = \beta_{n} W^{f}_{n-1,n}\boldsymbol{e}_{n-1}(t+1) + \lambda_{n} \boldsymbol{d}_{n}(t) + (1 - \beta_{n} - \lambda_{n}) \boldsymbol{e}_{n}(t) - \alpha_{n} \nabla{\epsilon_{n-1}(t)} ,
\end{equation}

where $\beta_{n}$, $\lambda_{n}$ ($0\leq \beta_n + \lambda_n \leq 1$), and $\alpha_{n}$ act as layer-dependent balancing coefficients for the feedforward, feedback, and error-correction terms, respectively. $\epsilon_{n-1}(t)$ denotes the reconstruction error at layer $n-1$ and is defined as the mean squared error (MSE) between the representation $e_{n-1}(t)$ and the predicted reconstruction $d_{n-1}(t)$ at that particular timestep. Layer $e_0$ is defined as the input image and remains constant over timesteps. All the weights $W^{f}_{n-1,n}$ and $W^{b}_{n+1,n}$ are fixed during these iterations.

\floatname{algorithm}{\scriptsize Pseudocode}
\begin{algorithm}[b!]
\scriptsize
\algsetup{linenosize=\scriptsize}
\caption{\scriptsize Predictive Coding Iterations}
\label{pseudocode}
\begin{algorithmic}[1]

\STATE{Input image: $e_{0}$}
\FOR{$n = 1$ to $N$}
\STATE{$e_{n} \gets Conv(e_{n-1})$}
\STATE{$d_{n-1} \gets deConv(e_{n})$}
\STATE{$\epsilon_{n-1} \gets ||d_{n-1} - e_{n-1}||_{2}^{2}$ }
\ENDFOR

\FOR{$t=1$ to $T$}
    \FOR{$n = 1$ to $N$}
        \STATE{$ff \gets \beta_{n}\cdot  Conv(e_{n-1})$}
        \STATE{$fb \gets 0$}
        \IF{$n < N$}
            \STATE $fb \gets \lambda_{n}\cdot d_{n}$
        \ENDIF
        \STATE{$e_{n} \gets ff+fb+(1 - \beta_{n} - \lambda_{n})\cdot e_{n}-\alpha_{n} \cdot  \nabla{\epsilon_{n-1}}$}
        \STATE{$d_{n-1} \gets deConv(e_{n})$}
        \STATE{$\epsilon_{n-1} \gets ||d_{n-1} - e_{n-1}||_{2}^{2}$ }
    \ENDFOR
\ENDFOR
\end{algorithmic}
\end{algorithm}

Each of the four terms in Equation~\ref{eq:pc_equation} contributes different signals, reflected by different arrow colors in Figure \ref{fig:pc_schematic}: (i) the feedforward term (green arrows; controlled by parameter $\beta$) provides information about the (constant) input and changing representations in the lower layers, (ii) the feedback correction term (red arrows; parameter $\lambda$), as proposed in~\cite{rao1999predictive,heeger2017theory}, guides activations towards their representations from the higher levels, thereby reducing the reconstruction errors over time, (iii) the memory term (blue arrows) acts as a time constant to retain the current representation over successive timesteps, and (iv) the feedforward error correction term (black arrows; controlled by parameter $\alpha$) corrects representations in each layer such that their generative feedback can better match the preceding layer. For this error correction term, we directly use the error gradient $\nabla{\epsilon_{n-1}}=[\frac{\partial\epsilon_{n-1}}{\partial e_{n}^{0}}, ..., \frac{\partial\epsilon_{n-1}}{\partial e_{n}^{k}}]$ to take full advantage of modern machine learning capabilities (where $k$ is the number of elements in $e_n$). While the direct computation of this error gradient is biologically implausible, it has been noted before that it is mathematically equivalent to propagating error residuals up through the (transposed) feedback connection weights $(W^{b})^{T}$, as often done in other predictive coding implementations~\cite{wen2018deep,rao1999predictive}. Together, the feedforward and feedback error correction terms fulfill the objective of predictive coding as laid out by Rao and Ballard~\cite{rao1999predictive}. We discuss the similarities and differences between our equations and those proposed in the original Rao and Ballard implementation in the Appendix~\ref{apndx:rao_ballard}.

While it is certainly possible to train such an architecture in an end-to-end fashion, by combining a classification objective for the feedforward weights $W^{f}$ with an unsupervised predictive coding objective (see Section~\ref{methods}) for the feedback weights $W^{b}$, we believe that the benefits of our proposed scheme are best demonstrated by focusing on the added value of the feedback pathway onto a pre-existing state-of-the-art feedforward network. Consequently, we implement the proposed strategy with two existing feedforward DCNN architectures as backbones: VGG16 and EfficientNetB0, both trained on ImageNet. We show that predictive coding confers higher robustness to these networks.

\subsection{Model architectures and training}
\label{methods}

We select VGG16 and EfficientNetB0, two different pre-trained feedforward networks on ImageNet, and augment them with the proposed predictive coding dynamics. The resulting models are called PVGG16 and PEfficientNetB0, respectively. The networks' ``bodies'' (without the classification head) are split into a cascade of $N$ sub-modules, where each plays the role of an $e_{n}$ in equation~(\ref{eq:pc_equation}). We then add deconvolutions as feedback layers $d_{n-1}$ connecting each $e_{n}$ to $e_{n-1}$, with kernel sizes accounting for the increased receptive fields of the neurons in $e_{n}$ or upsampling layers to match the size of the predictions and their targets (see Appendix~\ref{apndx:netarc}). We then train the parameters of the feedback deconvolution layers with an unsupervised reconstruction objective (with all feedforward parameters frozen). We minimize the reconstruction errors just after the first forward pass, and after a single deconvolution step (i.e. no error correction or predictive coding recurrent dynamics are involved at this stage):

\begin{equation}\label{eq:rec_loss}
    \mathcal{L} = \sum_{n=0}^{N-1}\parallel e_{n} - d_{n} \parallel_2^{2},
\end{equation}

where $e_n$ is the output of the $n^{th}$ encoder after the first forward pass and $d_n$ is the estimated reconstruction of $e_n$ via feedback/deconvolution (from $e_{n+1}$).

For both the networks, after training the feedback deconvolution layers, we freeze all of the weights, and set the values of hyperparameters to $\beta_{n}=0.8$, $\lambda_{n}=0.1$, and $\alpha_{n}=0.01$ for all the encoders/decoders in Equation~(\ref{eq:pc_equation}). We also explore various strategies for further tuning hyperparameters to improve the results (see Appendix~\ref{apndx:tuninghps} for the chosen hyperparameter values).

\subsection{\textit{Predify}}
\label{methods_predify}

To facilitate and automate the process of adding the proposed predictive coding dynamics to existing deep neural networks, we have developed an open-source Python package called \textit{Predify}. The package is developed based on PyTorch~\cite{NEURIPS2019_9015} and provides a flexible object oriented framework to convert any PyTorch-compatible network into a predictive network. While an advanced user may find it easy to integrate \textit{Predify} in their project manually, a simple text-based user interface (in TOML\footnote{\url{https://toml.io/en/}} format) is also provided to automate the steps. For the sake of improved performance and flexibility, \textit{Predify} generates the code of the predictive network rather than the Python object. Given the original network and a configuration file (e.g. \texttt{{\color{orange}\textquotesingle config.toml\textquotesingle}}) that indicates the intended source and target layers for the predictive feedback, three lines of code are enough to construct the corresponding predictive network:

\lstdefinestyle{mystyle}{
    backgroundcolor=\color{white},   
    commentstyle=\color{TealBlue},
    keywordstyle=\color{purple},
    stringstyle=\color{orange},
    basicstyle=\ttfamily\footnotesize,
    breakatwhitespace=false,         
    breaklines=true,                 
    captionpos=b,                    
    keepspaces=true,                 
    showspaces=false,                
    showstringspaces=false,
    showtabs=false,                  
    tabsize=2
}

\lstset{style=mystyle}

\begin{lstlisting}[language=Python]
from predify import predify

net = # load PyTorch network
predify(net,'./config.toml') # config file indicates the layers that
                             # will act as outputs of encoders.
\end{lstlisting}

The Appendix~\ref{apndx:predify} provides further details on the package, along with a sample config file and certain default behaviours. \textit{Predify} is an ongoing project available on GitHub\footnote{\url{https://github.com/miladmozafari/predify}} under GNU General Public License v3.0. Scripts for creating PVGG16 and PEfficientNetB0 from their feedforward instances and reproducing the results presented in this paper, as well as the pre-trained weights are also available on another GitHub repository\footnote{\url{https://github.com/bhavinc/predify2021}}.

\section{Related work}
\label{prior_work}

There is a long tradition of drawing inspiration from neuroscience knowledge to improve machine learning performance.~Some studies suggest using sparse coding, a concept closely related to predictive coding~\cite{huang2011predictive,olshausen1996emergence, oja1999image,chalk2018toward,paiton2020selectivity}, for image denoising~\cite{lu2013sparse} and robust deep learning~\cite{sulam2020adversarial,kim2020modeling}, while other studies focus on implementing feedback and horizontal recurrent pathways to tackle challenges beyond the core object recognition~\cite{linsley2020stable,wyatte2012limits,heeger2017theory,spoerer2017recurrent,linsley2018learning,Frosst2018DARCCCDA,krotov2018dense,kubilius2019brain,rajaei2019beyond}.

Here, we focus specifically on those studies that tried implementing predictive coding mechanisms in machine learning models~\cite{chalasani2013deep,lotter2016deep,wen2018deep,boutin2019sparse}. Out of these, our implementation is most similar to the Predictive Coding Networks (PCNs) of~\cite{wen2018deep}. These hierarchical networks were designed with a similar goal in mind: improving object recognition with predictive coding dynamics. However, their network (including the feedback connection weights) is solely optimized with a classification objective. As a result, their network does not learn to uniformly reduce reconstruction errors over timesteps, as the predictive coding theory would mandate. We also found that their network performs relatively poorly until the final timestep (see corresponding Figure~\ref{fig:prednet} in the Appendix~\ref{apndx:purdue_paper}), which does not seem biologically plausible: biological systems typically cannot afford to wait until the last iteration before detecting a prey or a predator. In the proposed method, we incorporate the feedforward drive into a similar PC dynamics and train the feedback weights in an unsupervised way using a reconstruction loss. We then show that these modifications help resolve PCNs' issues. We discuss these PCNs~\cite{wen2018deep} further in the Appendix~\ref{apndx:purdue_paper}, together with our own detailed exploration of their network's behavior.

Other approaches to predictive coding for object recognition include Boutin et al.~\cite{boutin2019sparse}, who used a PCN with an additional sparsity constraint. The authors showed that their framework can give rise to receptive fields which resemble those of neurons in areas V1 and V2 of the primate brain. They also demonstrated the robustness of the system to noisy inputs, but only in the context of reconstruction. Unlike ours, they did not show that their network can perform (robust) classification, and they did not extend their approach to deep neural networks.

Spratling~\cite{spratling2017hierarchical} also described PCNs designed for object recognition, and demonstrated that their network could effectively recognise digits and faces, and locate cars within images. Their update equations differed from ours in a number of ways: they used divisive/multiplicative error correction (rather than additive), and a form of biased competition to make the neurons ``compete'' in their explanatory power. The weights of the network were not trained by error backpropagation, making it difficult to scale it to address modern machine learning problems. Conversely, our proposed network architecture and PC dynamics are fully compatible with error backpropagation, making them a suitable option for large-scale problems. Indeed, the tasks on which they tested their network are simpler than ours, and the datasets are much smaller.

Huang et al.~\cite{huang2020neural} also aimed to extend the principle of predictive coding by incorporating feedback connections such that the network maximizes ``self consistency'' between the input image features, latent variables and label distribution. The iterative dynamics they proposed, though different from ours, improved the robustness of neural networks against gradient-based adversarial attacks on datasets such as Fashion-MNIST and CIFAR10.


\section{Results}

\begin{figure}
    \centering
    \includegraphics[width=\textwidth]{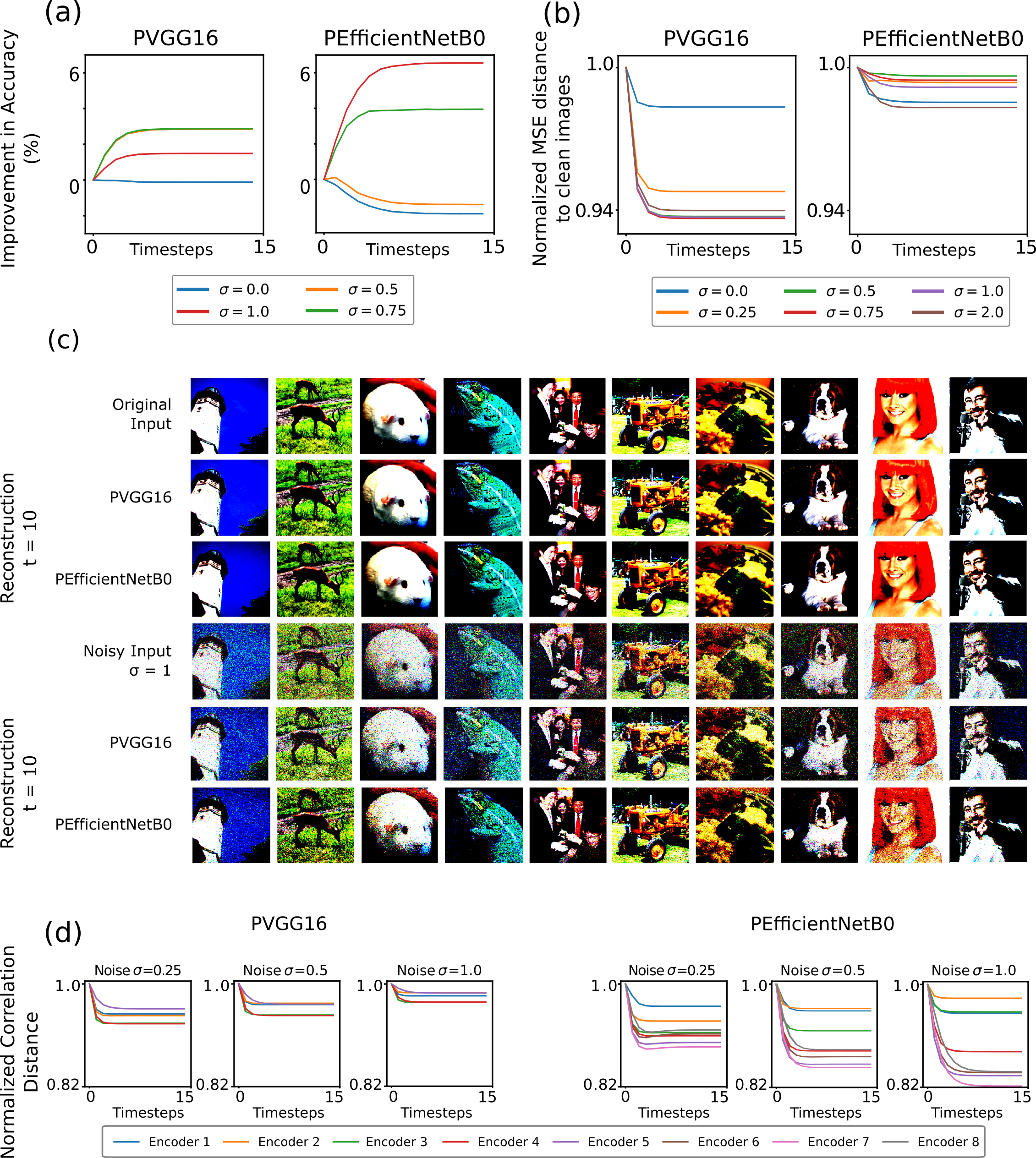}
    \caption{\textbf{Performance under Gaussian noise and projection towards the learned manifold.} (a) Improvement in recognition accuracy with reference to the feedforward baseline under various levels of Gaussian noise. Both networks demonstrate significant accuracy improvement across timesteps under noisy conditions, while maintaining a performance close to the feedforward level for clean images. (b) Normalized MSE distance between the image reconstruction ($d_0$) and the clean image ($e_0$). Irrespective of the noise level, image reconstruction consistently gets closer to the clean image across timesteps in both models. (c) Examples of clean and noisy input images together with their final reconstruction by the model (the row order from top to bottom is: original image, PVGG16 reconstruction, PEfficientNetB0 reconstruction; noisy image, PVGG16 reconstruction, PEfficientNetB0 reconstruction). For best viewing, we recommend zooming in on the electronic version. (d) Normalized correlation distance between representation of clean and noisy images for each encoder ($e_i$) across timesteps. The values are normalized with respect to the feedforward baseline (timestep 0). In both models and all encoders, the noisy representations tend to move toward the clean copies.}
    \label{fig:manifold_projection}
\end{figure}

Here we contrast the behavior of feedforward networks with their predictive coding augmentations. When considered at timestep 0 (i.e., after a single feedforward and feedback pass through the model), the deep predictive coding networks (DPCNs) and their accuracy are---by construction---exactly identical to their standard pretrained feedforward versions. Over successive timesteps, however, the influence of feedback and predictive coding iterations becomes visible. Here, we investigate for both DPCNs (PVGG16 and PEfficientNetB0): (i) how the PC dynamics update the networks' representations across timesteps, and in which direction relative to the learned manifold; (ii) how the networks benefit from PC under noisy conditions, or against adversarial attacks.

\subsection{Performance under Gaussian noise}
To understand the evolution of representations and the behavior of the proposed DPCNs, we first investigate their performance under the influence of different levels of Gaussian noise. To this end, we inject additive Gaussian noise to the ImageNet validation set, and monitor the models' performance across timesteps.

In Figure~\ref{fig:manifold_projection}a we provide the classification accuracy on these noisy images and  absolute values in the Table~\ref{tab:absolute_values}. We observed that both models progressively improve their recognition accuracy relative to their feedforward baseline (timestep 0) over successive iterations while imposing only a minor performance reduction on clean images. In other words, the networks are able to discard some of the noise by leveraging the predictive coding dynamics over timesteps.

\subsection{Projection towards the learned manifold}

In order to quantify DPCNs' denoising ability, we evaluate the quality of image reconstructions generated by each network using the mean squared error (MSE) between the clean image and its reconstruction generated by the first decoder. For each DPCN, we normalize these distances, by dividing them by the value obtained for the corresponding feedforward network (at t=0). We provide the absolute values in the Table~\ref{tab:mse_values_absolute}.  As Figures~\ref{fig:manifold_projection}b-c illustrate, the reconstructions become progressively cleaner over timesteps. It should be noted that the feedback connections were trained only to reconstruct clean images; therefore, this denoising property is an emerging feature of the PC dynamics.

Next, we test whether the higher layers of the proposed DPCNs also manifest this denoising property. Hence, we pass clean and noisy versions of all images from the ImageNet validation set through the networks, and measure the average correlation distance between the clean and noisy representations of each encoder at each timestep. As done above, these correlation distances are then normalized with the distance measured at timestep 0 (i.e., relative to the standard feedforward network). For both the networks, the correlation distances decrease consistently over timesteps across all layers (see Figure~\ref{fig:manifold_projection}d). This implies that predictive coding iterations help the networks to steer the noisy representations closer to the representations elicited by the corresponding (unseen) clean image. 

This is an important property for robustness.  When compared to clean images, noisy images can result in different representations at higher layers~\cite{xie2019feature} and consequently, produce significant classification errors. Various defenses have aimed to protect neural networks from perturbations and adversarial attacks by constraining the images to the ``original data manifold''. Accordingly, studies have used generative models such as GANs~\cite{samangouei2018defense,shen2017ape,meng2017magnet,jalal2017robust} or PixelCNNs~\cite{song2018pixeldefend} to constrain the input to the data manifold. Similarly, multiple efforts have been made to clean the representations in higher layers and keep them closer to the learned latent space~\cite{xie2019feature,tao2018attacks,orhan2019improving,roth2019odds}. 
Here, we demonstrate that feedback predictive coding iterations can achieve a similar goal by iteratively projecting noisy representations towards the manifolds learned during training, both in pixel (Figure~\ref{fig:manifold_projection}b-c) and representation spaces (Figure~\ref{fig:manifold_projection}d).

\begin{figure}[t!]
    \centering
    \includegraphics[width=0.7\textwidth]{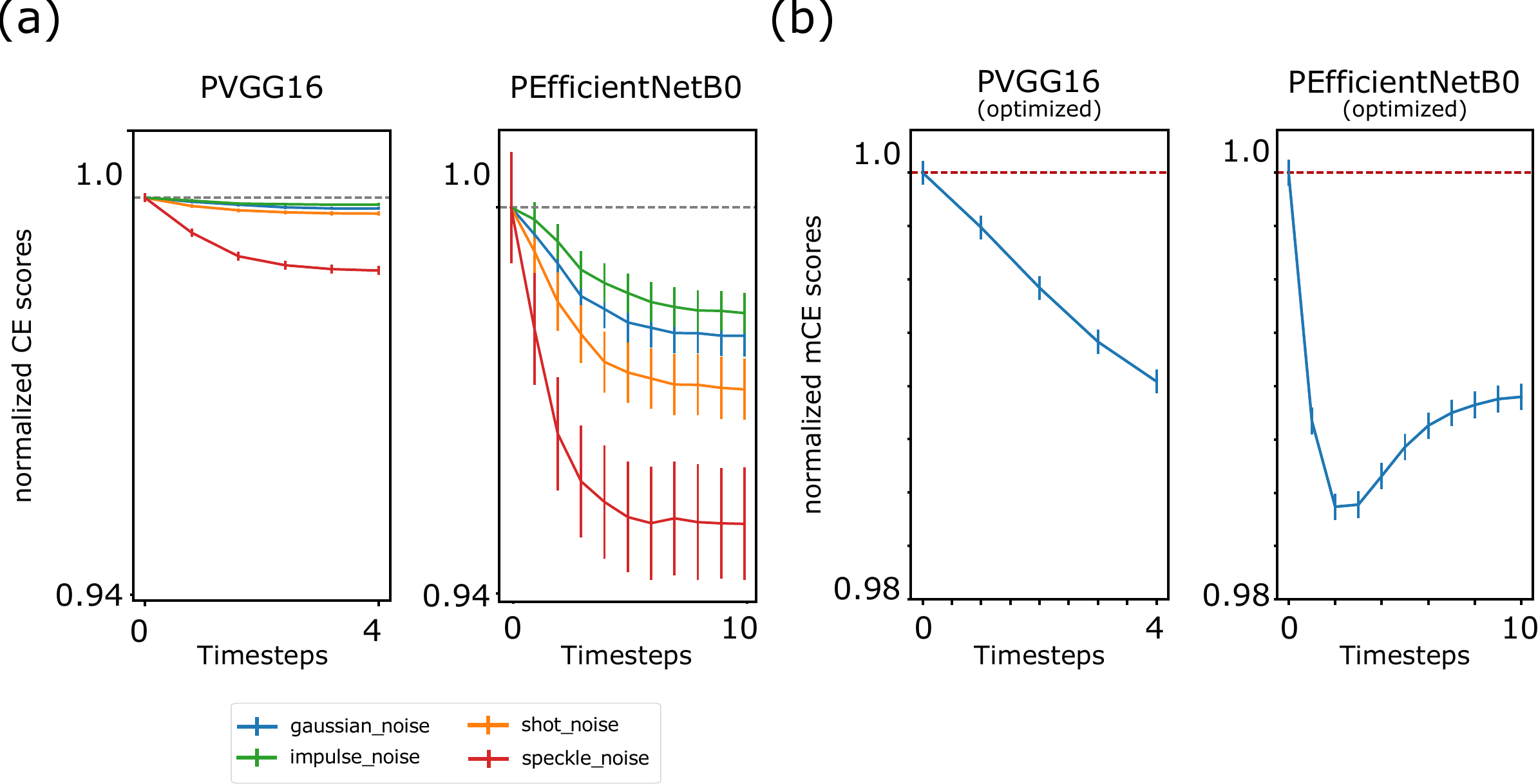}
    \caption{\textbf{Benchmarking robustness to ImageNet-C.} (a) Normalized corruption errors (CE) of PVGG16 and PEfficientNetB0 under four types of additive noise corruptions. The values are normalized with respect to the feedforward baseline. Both networks show consistent reductions in the errors across timesteps. (b) Normalized mean Corruption Error (mCE) scores for PVGG16 and PEfficientNetB0 on all the 19 corruptions available in the ImageNet-C dataset, when optimized hyperparameters are used (as described in the Appendix~\ref{apndx:tuninghps}). The values are normalized with respect to the feedforward baseline. In both the panels, error bars represent the standard deviation of the bootstrapped estimate of the mean value.}
    \label{fig:robustness_imagenet_c}
\end{figure}

\subsection{Benchmarking robustness to ImageNet-C}
Given the promising results with additive Gaussian noise (Figure~\ref{fig:manifold_projection}), we extend the noise variety and quantify the classification accuracy of the networks under different types of perturbations. We use ImageNet-C, a benchmarking dataset for noise robustness provided by~\cite{hendrycks2019benchmarking}, including 19 types of image corruptions across 5 severity levels each. To begin with, we evaluate DPCNs with pre-defined hyperparameter values (as provided in subsection~\ref{methods}). We observe that they improve the Corruption Error (CE) scores over timesteps for several of the additive-noise corruptions: Gaussian noise, shot noise, impulse noise or speckle noise (see Figure~\ref{fig:robustness_imagenet_c}), but fail to improve the overall mean Corruption Error, or mCE score (the recommended score for this benchmark~\cite{hendrycks2019benchmarking}).

Thus, instead of using pre-defined hyperparameter values, we fine-tune them using two different methods (see Appendix~\ref{apndx:tuninghps}), and repeat the above experiment. As shown in Figure~\ref{fig:robustness_imagenet_c}b, when the hyperparameters are more appropriately tuned for the task, the PC dynamics can increase noise robustness more generally across noise types, resulting in improvements of the mean Corruption Error (mCE) score. The CE plots for individual perturbations along with other recommended metrics (values normalized with AlexNet scores, Relative mCE scores) are provided in the Appendix~\ref{apndx:mce_alexnet}. 

Furthermore, in the Appendix~\ref{apndx:mce_robust_net}, we demonstrate that we can replicate these observations with a version of PEfficientNetB0 provided by~\citep{xie2020adversarial} that is robust to corruptions in the ImageNet-C dataset. We show that the recurrent dynamics we propose still help in further improving the mCE score of this already robust network.

\subsection{Benchmarking robustness to adversarial attacks}
\begin{figure}
    \centering
    \includegraphics[width=0.9\textwidth]{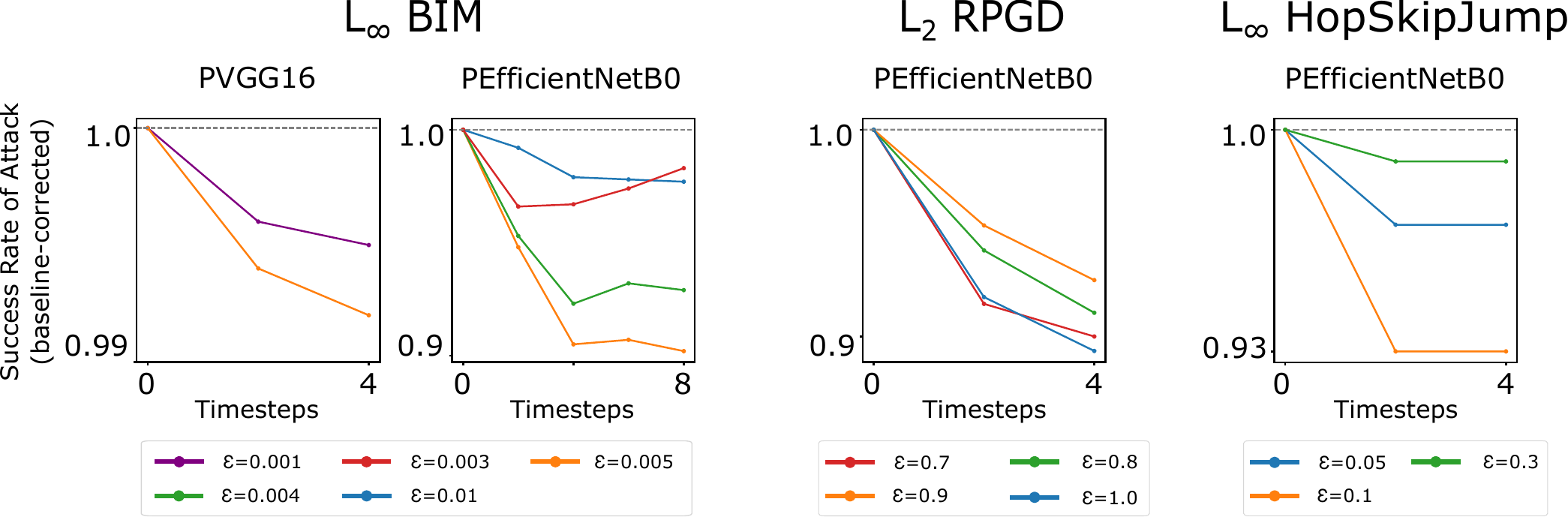}
    \caption{\textbf{Benchmarking robustness to adversarial attacks.} Plots show the success rate of targeted adversarial attacks against DPCNs across timesteps. The values are baseline-corrected, relative to the success rate at timestep 0 (feedforward baseline). Though we see some signs of reversals for a few perturbations, on average, both networks demonstrate improving robustness across timesteps to different types and/or levels of perturbations.}
    \label{fig:robustness_adversarial_attacks}
    \label{fig4}
\end{figure}

Finally, we evaluate the robustness of the networks across timesteps against adversarial attacks. The proposed DPCNs are recurrent models, meaning that their layer representations change on every timestep, and consequently, so do the classification boundaries in the last layer, leading to different accuracy and generalization errors across time (as seen above). To mitigate this effect and properly assess the changes in robustness due to the PC dynamics, for each network we start by selecting 1000 images from the ImageNet validation dataset such that they are correctly classified across all timesteps. Also, we only perform \emph{targeted} attacks so that for each image, the same attack target is given for all timesteps. Using the \textit{Foolbox} library~\cite{rauber2017foolbox}, we conduct targeted Basic\_Iterative\_Method attacks  (BIM, with $L_{\infty}$ norm)~\cite{goodfellow2014explaining} for both networks; although it would prove computationally prohibitive to systematically explore all standard types of adversarial attacks, we also evaluated random Projected\ Gradient\ Descent\ attacks (RPGD, with $L_2$ norm)~\cite{madry2018towards}, and non-gradient-based HopSkipJump attacks~\citep{chen2020hopskipjumpattack} on a subset of 100 images, specifically for PEfficientNetB0. Across various levels of allowed image perturbations (denoted as $\epsilon$s), the predictive coding iterations tend to decrease the success rate of the attacks across timesteps, for both networks and attacks (see Figure~\ref{fig4}). That is, DPCNs are more robust against these adversarial attacks than their feedforward counterparts.

\section{Discussion and Conclusion}

In this work, we explore the use of unsupervised recurrent predictive coding (PC) dynamics, based on neuroscientific principles, to augment modern deep neural networks. The resulting models have an initial feedforward sweep, compatible with visual processing in human and macaque brains~\cite{kar2019evidence,thorpe1996speed,hung2005fast,vanrullen2007power}. Following this feedforward sweep, consecutive layers iteratively exchange information regarding predictions and prediction errors, aiming to converge towards a stable explanation of the input. This dynamic system is inspired by, and reminiscent of, the ``canonical microcircuit'' (a central component of cortical structure~\cite{bastos2012canonical}) that relies on feedback signaling between hierarchically adjacent layers to update its activity. Overall, the augmented networks are closer to the architecture of biological visual systems, while gaining some desirable functional properties. For example, in~\cite{PANG2021164}, we also demonstrated that the proposed dynamics help the networks perceive illusory contours in a similar way to humans.

Here, we implemented these PC dynamics in two state-of-the-art DCNs, VGG16 and EfficientNetB0, and showed that they helped to improve the robustness of the networks against various corruptions (e.g. ImageNet-C). We demonstrated that this behavior, at least partly, stems from PC's ability to project both the corrupted image reconstructions and neural representations towards their clean counterparts.

We also tested the impact of our network augmentations against adversarial attacks; here again, we showed that PC helps to improve the robustness of the networks. So far, the most promising strategy for achieving robustness has been adversarial training, whereby adversarial datapoints are added to the training dataset. While efficient, this strategy was also shown to be strongly limited~\cite{schmidt2018adversarially,zhang2019defense}. Apart from factors like the choice of the norms used for training, or the high computation requirements, it is ultimately performed with a supervised loss function that can alter the decision boundaries in undesirable ways~\cite{zhang2019defense,tanay2016boundary}.
Most importantly, adversarial training shares very little, if any, resemblance to the way the brain achieves robustness. Instead, here we start from biological principles and show that they can lead to improved adversarial robustness. It is worth mentioning that both our networks achieved robustness totally via unsupervised training of the feedback connections (while of course, the backbone feed-forward networks that we used were pretrained in a supervised manner). We avoided using costly adversarial training, or tuning our hyperparameters specifically for classification under each attack. This likely explains why the models, while improving in robustness compared to their feedforward versions, remain far from state-of-the-art adversarial defenses. On the other hand, we believe that addition of these methods (adversarial training, hyperparameter tuning) to the training paradigm, in future work, could further improve the networks' adversarial robustness.

For the present experiments, we made a choice of using different objectives for training the feedforward and feedback weights: pre-trained feedforward weights optimized for classification, feedback weights trained with a reconstruction objective (computed after a single time-step). On the one hand, we note that it is perfectly feasible to train a similar predictive coding architecture with a single objective (classification, reconstruction, or otherwise) for both feedforward and feedback weights~\citep{PANG2021164,alamia2021role}. On the other hand, our choice has several advantages. First, using a feedforward backbone pretrained for classification allowed us to demonstrate the effect of our dynamics on pre-existing state-of-the-art neural networks. Some authors have tried training both feedforward and feedback connections together for classification~\citep{wen2018deep} at the final timestep for relatively smaller networks, but as we discussed in our explorations in the Appendix~\ref{apndx:purdue_paper}, we found that the resulting network ended up classifying correctly at the last timestep, with very poor performance during early timesteps. This problem could be addressed by training over time-averaged metrics, such as the average cross-entropy loss for $N$ timesteps. Nonetheless, training the feedback weights for reconstruction instead of classification has the additional advantage that it can be done entirely without supervision. We chose to train the feedback weights for a single time-step, because training with recurrence over multiple timesteps would have required unrolling the network over time. Hence, training a large network like PVGG16 for say 5 or 10 timesteps would incur significant computational challenges. Furthermore, our use of a one-step reconstruction objective allowed us to train the feedback weights independently of the various hyperparameters of our predictive coding dynamics ($\beta$, $\lambda$, and $\alpha$), which only influence the model behavior after the second timestep. Training these weights using recurrence would have required to (i) either fix the values of these hyperparameters beforehand, leading to constraints of expensive hyperparameter explorations; (ii) or directly train these hyperparameters as parameters of the model, probably with additional constraints to prevent the network from reaching trivial values (e.g., if all hyperparameters but the feedforward term $\beta$ converge to zero, the network performs identically to a feedforward one). Finally, from a neuroscience perspective, whether and how the brain combines discriminative and generative representations has been an open question addressed by many researchers, e.g.~\citep{al2021reconstructing,dicarlo2021does,huffman2014multivariate}. Our approach of a discriminative (classification-trained) feedforward coupled with generative (reconstruction-trained) feedback could be considered another attempt in this direction.

We speculate that the proposed PC dynamics could help improve robustness in most feedforward neural architectures. To facilitate further explorations in this direction, we provided a Python package, called \textit{Predify}, which allows users to implement recurrent PC dynamics in any feedforward DCN, with only a few lines of code. \textit{Predify} automates the network building, and thus simplifies experiments. On the other hand, there is as yet no established method or criteria to automate the process of identifying the appropriate number of encoding layers, their source and target layers in the DCN hierarchy, and the corresponding hyperparameter values. This remains an open research question, and a requirement for manual explorations and tuning from \textit{Predify} users. For instance, our own explorations with augmenting ResNets through \textit{Predify} proved difficult, and failed in some situations but succeeded in others. More specifically, as developed in~\citep{alamia2021role} using \textit{Predify}, ResNet augmentations always achieved noise robustness when the hyperparameter values (controlling the feedforward, feedback, and memory terms) could be tuned separately for each noise type; but we found it challenging to identify a single set of hyperparameters that could generalize to all noise types. Nonetheless, we are hopeful that the package will prove useful to the community.  The code is structured such that users can readily adapt it to test their hypotheses. In particular, it should allow both proponents and opponents of the predictive coding theory to investigate its effects on any DCN.

Overall, this work contributes to the general case for continuing to draw inspiration from biological visual systems in computer vision, both at the level of model architecture and dynamics. We believe that our user-friendly Python package \textit{Predify} can open new opportunities, even for neuroscience researchers with little background in machine learning, to investigate bio-inspired hypotheses in deep computational models, and thus bridge the gap between the two communities.

\section{Broader Impacts}
\label{broader_impacts}

The research discussed above proposes novel ways of using brain-inspired dynamics in current machine learning models. Specifically, it demonstrates a neuro-inspired method for improving the robustness of machine learning models. Given that such models are employed by the general public, and are simultaneously shown to be heavily vulnerable, research efforts to increase (even marginally) or to understand their robustness against mal-intentioned adversaries has high societal relevance. 

Importantly, the research also aims to bridge techniques between two different fields--neuroscience and machine learning, which can potentially open new avenues for studying the human brain. For example, it could help better understand the unexplained neural activities in patients, to improve their living conditions, and in the best case, in the treatment of their conditions. While this may also be associated with inherent risks (related to privacy or otherwise), there are clear potential benefits to society.

The likelihood of sentient AI arising from this line of research is estimated to be rather low.

\clearpage
\begin{ack}

This work was funded by an ANITI (Artificial and Natural Intelligence Toulouse Institute) Research Chair to RV (ANR grant ANR-19-PI3A-0004), as well as ANR grants AI-REPS (ANR-18-CE37-0007-01) and OSCI-DEEP (ANR-19-NEUC-0004). 

\end{ack}

\bibliography{references}
\bibliographystyle{unsrt}

\clearpage
\clearpage

\appendix
\renewcommand{\thefigure}{S\arabic{figure}}
\setcounter{figure}{0}

\renewcommand{\thetable}{S\arabic{table}}
\setcounter{table}{0}

\section{Appendix}

\subsection{Getting Started with Predify}
\label{apndx:predify}
Both VGG16 and EfficientNetB0 are converted to predictive coding networks PVGG16 and PEfficientNetB0, using the Predify package. The fastest and easiest way to convert a feedforward network into its predictive coding version is to use Predify's text-based interface which supports configuration files in TOML format.

The current version of Predify assumes that there is no gap between the encoders. Therefore, in the minimal case, one only needs to provide a list of sub-module names in the target feedforward network. Then, Predify takes care of the rest by converting each of them into an encoder and assigning default decoders. More precisely, let $x$ and $y$ denote the input and output of a layer (or complex sub-module, potentially including multiple layers) that is selected to be an encoder ($e_n$). If $x$ and $y$ respectively have the size $(c_{in}, h_{in}, w_{in})$ and $(c_{out}, h_{out}, w_{out})$; then, the default decoder's structure that predicts this encoder ($d_{n+1}$) is a 2D upscaling operation by the factor of $(h_{in}/h_{out}, w_{in}/w_{out})$ followed by a transposed convolutional layer with $c_{out}$ channels and $3\times3$ window size. The values of hyperparameters will be set to $\beta_n=0.3$, $\lambda_n=0.3$, and $\alpha_n=0.01$

In Predify, each encoder ($e_n$) and the decoder that uses its output to predict the activity of the encoder below ($d_{n-1}$) is called a \textit{PCoder}. To verify the functionality of Predify's default settings, we applied it for PEfficientNetB0 used in this work. Here is the corresponding minimal configuration file:

\definecolor{codegreen}{rgb}{0,0.6,0}
\definecolor{codegray}{rgb}{0.5,0.5,0.5}
\definecolor{codepurple}{rgb}{0.58,0,0.82}
\definecolor{backcolour}{rgb}{0.95,0.95,0.92}
\lstdefinestyle{mystyle}{
    backgroundcolor=\color{backcolour},   
    commentstyle=\color{codegreen},
    keywordstyle=\color{magenta},
    numberstyle=\tiny\color{codegray},
    stringstyle=\color{codepurple},
    basicstyle=\ttfamily\footnotesize,
    breakatwhitespace=false,         
    breaklines=true,                 
    captionpos=b,                    
    keepspaces=true,                 
    numbersep=5pt,                  
    showspaces=false,                
    showstringspaces=false,
    showtabs=false,                  
    tabsize=2
}

\lstset{style=mystyle}

\begin{lstlisting}[language=Python]
name = "PEfficientNetB0"

input_size = [3,224,224]
gradient_scaling = true
shared_hyperparameters = false

[[pcoders]]
module = "act1"
[[pcoders]]
module = "blocks[0]"
[[pcoders]]
module = "blocks[1]"
[[pcoders]]
module = "blocks[2]"
[[pcoders]]
module = "blocks[3]"
[[pcoders]]
module = "blocks[4]"
[[pcoders]]
module = "blocks[5]"
[[pcoders]]
module = "blocks[6]"
\end{lstlisting}

One can easily override the default setting by providing all the details for a PCoder. Here is the configuration corresponding to the PVGG16 used in this work:

\begin{lstlisting}[language=Python]
imports = [
"from torch.nn import Sequential, ReLU, ConvTranspose2d",
]

name = "PVGG16"

input_size = [3, 224, 224]
gradient_scaling = true
shared_hyperparameters = false

[[pcoders]]
module = "features[3]"
predictor = "ConvTranspose2d(64, 3, kernel_size=(5, 5), stride=(1, 1), padding=(2, 2))"
hyperparameters = {feedforward=0.2, feedback=0.05, pc=0.02}

[[pcoders]]
module = "features[8]"
predictor = "Sequential(ConvTranspose2d(128, 64, kernel_size=(10, 10), stride=(2, 2), padding=(4, 4)), ReLU(inplace=True))"
hyperparameters = {feedforward=0.4, feedback=0.1, pc=0.05}

[[pcoders]]
module = "features[15]"
predictor = "Sequential(ConvTranspose2d(256, 128, kernel_size=(14, 14), stride=(2, 2), padding=(6, 6)), ReLU(inplace=True))"
hyperparameters = {feedforward=0.4, feedback=0.1, pc=0.008}

[[pcoders]]
module = "features[22]"
predictor = "Sequential(ConvTranspose2d(512, 256, kernel_size=(14, 14), stride=(2, 2), padding=(6, 6)), ReLU(inplace=True))"
hyperparameters = {feedforward=0.5, feedback=0.1, pc=0.0024}

[[pcoders]]
module = "features[29]"
predictor = "Sequential(ConvTranspose2d(512, 512, kernel_size=(14, 14), stride=(2, 2), padding=(6, 6)), ReLU(inplace=True))"
hyperparameters = {feedforward=0.6, feedback=0.0, pc=0.006}
\end{lstlisting}

The network configuration files (in TOML format) are available to download on GitHub\footnote{\url{https://github.com/bhavinc/predify2021}}.

\subsection{Network Architectures}
\label{apndx:netarc}

VGG16 consists of five convolution blocks and a classification head. Each convolution block contains two or three convolution+ReLU layers with a max-pooling layer on top. For each $e_{n}$ in PVGG16, we selected the max-pooling layer in block $n-1$ and all the convolution layers in block $n$ of VGG16 (for $n\in\{1,2,3,4,5\}$) as the sub-module that provides the feedforward drive. Afterwards, to predict the activity of each $e_{n}$, a deconvolution layer $d_{n}$ is added which takes the $e_{n+1}$ as the input. Here, deconvolution kernel sizes are set by taking the increasing receptive field sizes into account.

In the case of PEfficientNetB0, we used PyTorch implementation of EfficientNetB0 provided in \url{https://github.com/rwightman/pytorch-image-models}. This implementation of EfficientNetB0 consists of eight blocks of layers (considering the first convolution and batch normalization layers as a separate block). Similar to PVGG16, we convert each of these blocks into an encoder ($e_n$) and add deconvolution layers accordingly. This time we set the kernel size of all deconvolution layers to 3x3 and use upsampling layers to compensate the shrinkage of layer size through the feedforward pathway (i.e. Predify's default setting).

Table~\ref{tab:netarc} summarizes PVGG16's architecture. Moreover, the hyperparameter values  are provided in Tables~\ref{tab:hyper1} and~\ref{tab:hyper2}.

\begin{table}[H]

    \scriptsize
    \centering

    \caption{Architectures of $e_n$s and $d_n$s for PVGG16 and PEfficientNetB0. Conv (channel, size, stride), MaxPool (size, stride), Deconv (channel, size, stride), Upsample (scale\_factor), BN is BatchNorm, $[\ ]_+$ is ReLU, and $[\ ]_*$ is SiLU. EfficientBlock corresponds to each block in PyTorch implementation of EfficientNetB0.}
    \begin{tabular}{|c|c|c||c|c|}
        \hline
        & \multicolumn{2}{|c||}{PVGG16}& \multicolumn{2}{|c|}{PEfficientNetB0}\\
         & \multicolumn{2}{|c||}{Input Size: 3x224x224} & \multicolumn{2}{|c|}{Input Size: 3x224x224}\\
        \cline{2-5}
        & $e_n$ & $d_{n-1}$ & $e_n$ & $d_{n-1}$ \\ 
        \hline
        PCoder1 &
        \begin{tabular}{c}
             $\big[$Conv (64, 3, 1)$\big]_+$ \\
             $\big[$Conv (64, 3, 1)$\big]_+$ \\
        \end{tabular}
        &
        \begin{tabular}{c}
             Deconv (3, 5, 1)
        \end{tabular}
        &
        \begin{tabular}{c}
             $[$BN (Conv (32, 3, 2))$]_*$
        \end{tabular}
        &
        \begin{tabular}{c}
             Upsample (2)\\
             Deconv (3, 3, 1)
        \end{tabular}
        \\
        \hline
        PCoder2 &
        \begin{tabular}{c}
             MaxPool (2, 2) \\
             $\big[$Conv (128, 3, 1)$\big]_+$ \\
             $\big[$Conv (128, 3, 1)$\big]_+$ \\
        \end{tabular}
        &
        \begin{tabular}{c}
             $\big[$Deconv (64, 10, 2)$\big]_+$
        \end{tabular}
        &
        \begin{tabular}{c}
             EfficientBlock0
        \end{tabular}
        &
        \begin{tabular}{c}
             Deconv (32, 3, 1)
        \end{tabular}
        \\
        \hline
        PCoder3 &
        \begin{tabular}{c}
             MaxPool (2, 2) \\
             $\big[$Conv (256, 3, 1)$\big]_+$ \\
             $\big[$Conv (256, 3, 1)$\big]_+$ \\
             $\big[$Conv (256, 3, 1)$\big]_+$ \\
        \end{tabular}
        &
        \begin{tabular}{c}
             $\big[$Deconv (128, 14, 2)$\big]_+$
        \end{tabular}
        &
        \begin{tabular}{c}
             EfficientBlock1
        \end{tabular}
        &
        \begin{tabular}{c}
             Upsample (2)\\
             Deconv (16, 3, 1)
        \end{tabular}
        \\
        \hline
        PCoder4 &
        \begin{tabular}{c}
             MaxPool (2, 2) \\
             $\big[$Conv (512, 3, 1)$\big]_+$ \\
             $\big[$Conv (512, 3, 1)$\big]_+$ \\
             $\big[$Conv (512, 3, 1)$\big]_+$ \\
        \end{tabular}
        &
        \begin{tabular}{c}
             $\big[$Deconv (256, 14, 2)$\big]_+$
        \end{tabular}
        &
        \begin{tabular}{c}
             EfficientBlock2
        \end{tabular}
        &
        \begin{tabular}{c}
             Upsample (2)\\
             Deconv (24, 3, 1)
        \end{tabular}
        \\
        \hline
        PCoder5 &
        \begin{tabular}{c}
             MaxPool (2, 2) \\
             $\big[$Conv (512, 3, 1)$\big]_+$ \\
             $\big[$Conv (512, 3, 1)$\big]_+$ \\
             $\big[$Conv (512, 3, 1)$\big]_+$ \\
        \end{tabular}
        &
        \begin{tabular}{c}
             $\big[$Deconv (512, 14, 2)$\big]_+$
        \end{tabular}
        &
        \begin{tabular}{c}
             EfficientBlock3
        \end{tabular}
        &
        \begin{tabular}{c}
             Upsample (2)\\
             Deconv (40, 3, 1)
        \end{tabular}
        \\
        \hline
        PCoder6 &
        -
        &
        -
        &
        \begin{tabular}{c}
             EfficientBlock4
        \end{tabular}
        &
        \begin{tabular}{c}
             Deconv (80, 3, 1)
        \end{tabular}
        \\
        \hline
        PCoder7 &
        -
        &
        -
        &
        \begin{tabular}{c}
             EfficientBlock5
        \end{tabular}
        &
        \begin{tabular}{c}
             Upsample (2)\\
             Deconv (112, 3, 1)
        \end{tabular}
        \\
        \hline
        PCoder8 &
        -
        &
        -
        &
        \begin{tabular}{c}
             EfficientBlock6
        \end{tabular}
        &
        \begin{tabular}{c}
             Deconv (192, 3, 1)
        \end{tabular}
        \\
        \hline
    \end{tabular}
    \label{tab:netarc}
\end{table}

\subsection{Execution Time}
\label{apndx:exec_time}

Since we used a variable number of GPUs for the different experiments, an exact execution time is hard to pinpoint. Briefly, depending on the number of timesteps, analysing mCE scores and adversarial attacks on PEfficientNetB0 took around 15-20 hours on an NVIDIA TitanV gpu. These numbers were about three to four times higher for experiments on PVGG16. For both the networks, training the feedback weights on the ImageNet dataset generally finished before 5 epochs, which took approximately 7-8 hours for a single GPU.

\subsection{Gradient Scaling}
\label{apndx:grad_scale}

In our dynamics, the error ($\epsilon_{n-1}$) is defined as a scalar quantity whose gradient is taken with respect to the activation of the higher layer ($e_{n}$). That is,

\begin{align}
	\nabla\epsilon_{n-1} &= \begin{bmatrix}
							\frac{\partial\epsilon_{n-1}}{\partial e_n^1} \\
							\vdots \\
							\frac{\partial\epsilon_{n-1}}{\partial e_n^L}
					\end{bmatrix}
\end{align}

where L denotes the number of elements in $e_{n}$. The partial derivative with respect to $e_n^j$ can then be written as, 

\begin{align}
\label{kceq6}
\frac{\partial\epsilon_{n-1}}{\partial e_n^j} & =  \frac{1}{K}\sum_i^K\frac{\partial(e^i_{n-1} - d^i_{n-1})^2}{\partial e^j_n}  \\
\end{align}

where K is the number of elements in $e_{n-1}$ ( = channels x width x height).
Equation~\ref{kceq6} highlights how the dimensionality of the prediction (equivalently the error term) affects the gradients, scaling them down by a factor K.

This can be easily seen by supposing that the gradients with respect to $e_n^j$ are i.i.d normally distributed around 0 with standard deviation $\sigma$,

\begin{align}
    \frac{\partial(e^i_{n-1} - d^i_{n-1})^2}{\partial e^j_{n}} \sim \mathcal{N}(0,\sigma^2)
\end{align}
\begin{align}
    \sum_i^K\frac{\partial(e^i_{n-1} - d^i_{n-1})^2}{\partial e^j_{n}} \sim \mathcal{N}(0,K\sigma^2)
\end{align}

Thus, 
\begin{align}
    \frac{\partial \epsilon_{n-1}}{\partial e^j_n}  &= \frac{1}{K}\sum_i^K\frac{\partial(e^i_{n-1} - d^i_{n-1})^2}{\partial e_n^j} \sim \mathcal{N}(0,\frac{\sigma^2}{K})
\end{align}

This scaling is further troublesome in DCNs, where most gradients are zero since they are not part of the receptive field of the element $e_n^j$. Hence assuming that there are only C elements (kernel*channels) that are part of the receptive field of $e_n^j$, 

\begin{align}
    \sum_i^K\frac{\partial(e^i_{n-1} - d^i_{n-1})^2}{\partial e_n^j} = \sum_i^C\frac{\partial(e^i_{n-1} - d^i_{n-1})^2}{\partial e_n^j} \sim \mathcal{N}(0,C\sigma^2)
\end{align}

Hence,

\begin{align}
\label{eq9}
    \frac{\partial \epsilon_{n-1}}{\partial e^j_n} = \frac{1}{K}\sum_i^C\frac{\partial(e^i_{n-1} - d^i_{n-1})^2}{\partial e^j_n} \sim \mathcal{N}(0,\frac{C\sigma^2}{K^2})
\end{align}

We use Equation~\ref{eq9} to, at least partly, counteract this effect due to the dimensionality of the prediction errors. We multiply the gradient by a factor of $\sqrt{K^2/C}$ to scale them in a way that is more comparable across layers, and thus apply a more meaningful step size for correcting the errors.

\subsection{Prior work: PCNs}
\label{apndx:purdue_paper}

To better understand the model proposed by Wen et al.~\cite{wen2018deep} and its differences to ours, we conducted several experiments using the code that they provided, and report here our most compelling observations. A first striking shortcoming was that the accuracy of their feedforward baseline was far from optimal. Using their code, with relatively minor tweaks to the learning rate schedule, we were able to bring it up from \textasciitilde60\% to 70\% -- just a few percentage points below their recurrent network. We expect that this could be further improved with a more extensive and systematic hyperparameter search. In other words, their training hyperparameters appeared to have been optimised for their predictive coding network, but not -- or not as much -- for their feedforward baseline. We further found that a minor change to the architecture - using group normalisation layers after each ReLU – leads to a feedforward network which performs on par with the recurrent network, with a mean over 6 runs of 72\% and best of 73\%. Adding the same layers to the recurrent network did not lead to a corresponding improvement in accuracy.

\begin{figure}

\centering
\begin{subfigure}{0.3\textwidth}
\includegraphics[width=\linewidth]{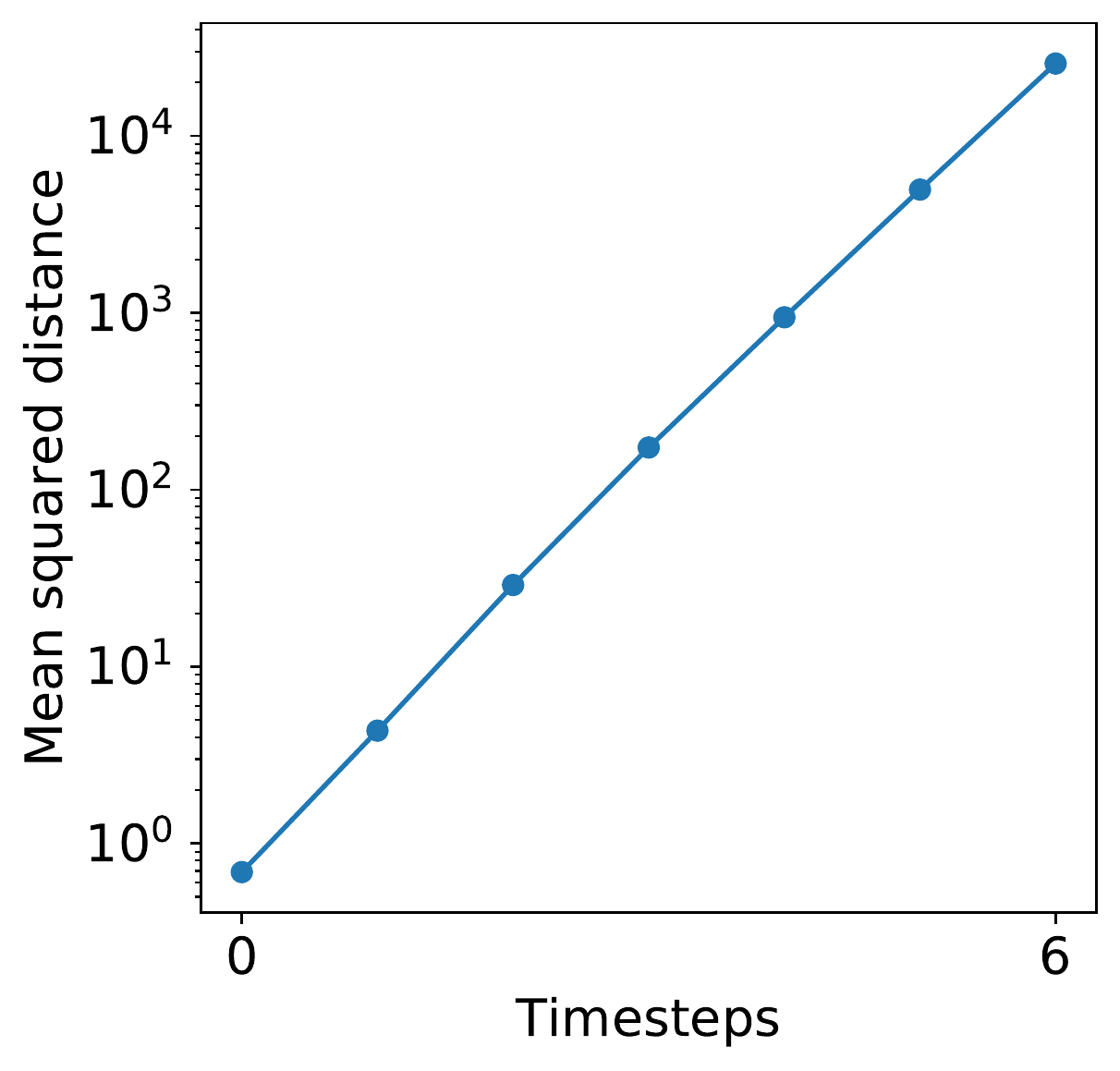}
\captionsetup{justification=centering}
\caption{Reconstruction error over timesteps}
\label{fig:prednet_reconstructions}
\end{subfigure}
\begin{subfigure}{0.45\textwidth}
\includegraphics[width=\linewidth]{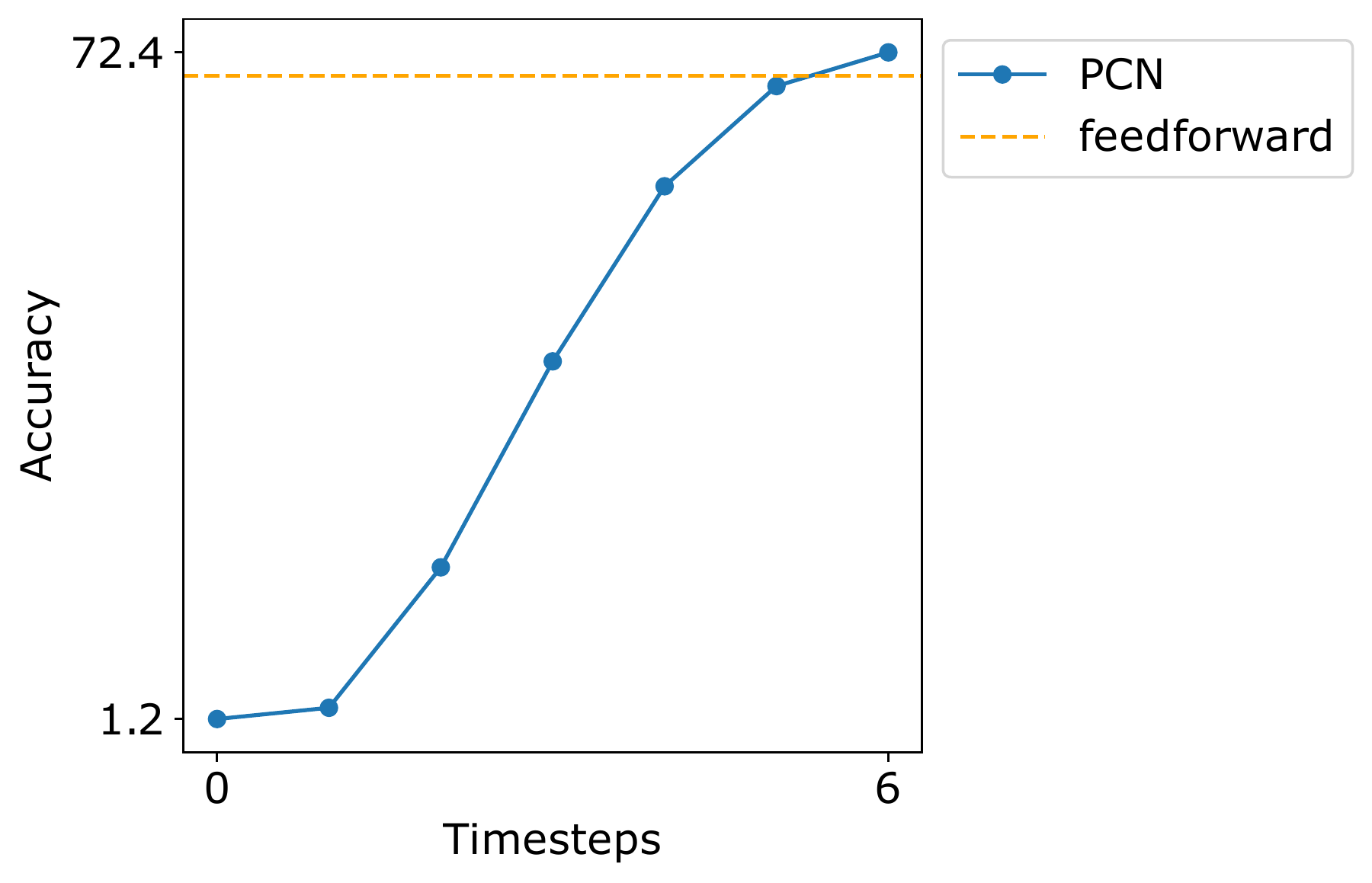}
\captionsetup{justification=centering}
\caption{Accuracy over timesteps \\ on CIFAR100 testset}
\label{fig:prednet_accuracy}
\end{subfigure}

\caption{\textbf{PCN}: Panel (a) shows the reconstruction errors of the model over timesteps. It does not decrease over timesteps, as would be expected in a predictive coding system. Panel (b) depicts the accuracy of the model on the CIFAR100 test dataset. The model performs at chance level at early timesteps and then becomes better in the last few timesteps.}
\label{fig:prednet}
\end{figure}

We also found that the network had poor accuracy (underperforming the optimized feedforward baseline) until the final timestep, as can been seen in Figure \ref{fig:prednet_accuracy}. This can be clarified by a closer reading of Figure 3 of their paper: the reported improvements over cycles from 60\% at timestep 0 to more than 70\% at timestep 6 are for seven distinct networks, each evaluated only at the timestep they were trained for. So in fact, in their model the predictive coding updates do not gradually improve on an already reasonably guess. This is clearly not biologically plausible: visual processing would be virtually useless if the correct interpretation of a scene only crystallised after a number of ``timesteps''. By the time a person has identified an object that object is likely to have disappeared or, in a worst case scenario, eaten them. We also experimented with feeding the classification error at each timestep into an aggregate loss function, but this lead to a network which, while performing well, essentially did not improve over timesteps.

Figure \ref{fig:prednet_reconstructions} shows that the network does not uniformly minimise reconstruction errors over time for all layers, and thus is not performing correct predictive coding updates. In fact the total reconstruction error (across all layers) increases exponentially over timesteps. There are a number of possible explanations for this. Firstly, in the case of the network with untied weights, the authors choose to make a strong assumption in the update equations (seen as the equivalence of their Equations 5 and 6): that the feedback weights can be assumed to be the transpose of the feedforward weights, i.e. $W^{b}=(W^{f})^{T}$. They thus propagate the feedforward error through the feedforward weights. However, it might be that the network learns feedback weights which essentially invert the feedforward transformation as assumed, but this is not guaranteed, and nor is it explicitly motivated through the classification loss function. Indeed, because the network is not motivated to learn a representation at earlier timesteps which produces a good prediction, it does not necessarily need to learn the inverse transformation: it can instead learn some other transformation which, when applied with the update equations, leads the network to \textit{end up} in the right place. That being said, this assumption is valid for the network with tied weights, and this network also does not uniformly reduce reconstruction error over timesteps. Possibly, the presence of ReLU non-linearities means that the forward convolution may still not be perfectly invertible by a transposed convolution. Finally, in line with this unexpected increase of reconstruction errors over time, we have also failed to extract good image reconstructions from the network as seen in Figure 5 of their paper, although in private communication the authors indicated that this was possible with some other form of normalisation.

In short, while the ideas put forward in~\cite{wen2018deep} share similarities with our own, their exact implementation did not support the claims of the authors, and the question of whether predictive coding can benefit deep neural networks remained an open one. We hope that our approach detailed in the present study can help resolve this question.

\clearpage
\subsection{Comparing with Rao and Ballard}
\label{apndx:rao_ballard}

This section aims to start from the equations initially provided in Rao and Ballard~\cite{rao1999predictive} and compare them to ours. The parallels drawn will help to highlight the similarities and the differences between both the approaches. 

Rao and Ballard consider a two-layer system, and start with the assumption that the brain possesses a set of internal causes, denoted as $\mathbf{r}$ (in matrix notation), that it uses to predict the visual stimulus, for example an input image $\mathbf{I}$, such that

\begin{equation}
    \mathbf{I} \approx f(U \mathbf{r})
\end{equation}

where $f(.)$ is some nonlinear activation function. This $\mathbf{r}$ can be equalled to encoding layer $e_{1}$ in our equations, with $\mathbf{I}$ being the input image $e_{0}$ or its reconstruction $d_{0}$. $U$ here, represents the top-down weight matrix (equivalent to $W^{b}_{1,0}$) that helps to make a prediction about the input image. That is, 

\begin{equation}
    \mathbf{I} \approx f(U \mathbf{r})  \equiv e_0 \approx d_0 = W^b_{1,0}e_1
\end{equation}

In this two-layer hierarchical architecture, $\mathbf{r}$ itself is predicted by the higher layer $\mathbf{r}^{h}$ using the weight matrix $U^{h}$, equivalent to how $e_1$ is predicted by $e_2$ using $W^{b}_{2,1}$ in our model. This prediction denoted as $\mathbf{r}^{td}$ in Rao and Ballard's original implementation can be equalled to $d_{1}$ in our equations. 

\begin{equation}
    \mathbf{r}^{td} = f(U^{h}\mathbf{r}^{h}) \equiv d_1 = W^b_{2,1}e_2
\end{equation}

The errors made in making the predictions are defined, like ours, as the mean squared distance,

\begin{align}
\label{error_equation}
    \epsilon_0 &= (\mathbf{I} - f(U\mathbf{r}))^{T}(\mathbf{I} - f(U\mathbf{r})) \\
    \epsilon_1 &= (\mathbf{r} - \mathbf{r}^{td})^{T}(\mathbf{r} - \mathbf{r}^{td})
\end{align}

Please note that differentiating the prediction error $\epsilon_0$ with respect to $\mathbf{r}$ (similar to taking the gradient of $\epsilon_{n-1}$ with respect to $e_n$ as done in our error-correction term) gives us,

\begin{align}
    \nabla \mathbf{\epsilon_0} &= -2U^{T} {\frac{\partial f}{\partial U\mathbf{r}}}^{T}(\mathbf{I} - f{U\mathbf{r}} ) \\ 
    \label{sds}
    &=  -k U^{T}(\mathbf{I} - f(U\mathbf{r}))
\end{align}

which will be useful later.

As per the predictive coding theory, the brain tries both to learn parameters ($U$ and $U^h$) over a dataset of natural inputs, and tries to modify its neural activations ($\mathbf{r}$ and $\mathbf{r}^h$) over time given a particular input, in such a way as to minimize the total error $E$, defined as:

\begin{align}
\label{eq_energy}
    E = a \cdot \underbrace{(\mathbf{I} - f(U\mathbf{r}))^{T}(\mathbf{I} - f(U\mathbf{r}))}_{\epsilon_0} + b \cdot \underbrace{(\mathbf{r} - \mathbf{r}^{td})^{T}(\mathbf{r} - \mathbf{r}^{td}) }_{\epsilon_1}
\end{align}

Here $a$ and $b$ act as constants that weigh the errors in this two-level hierarchichal network. Equation~\ref{eq_energy} is reflected as Equation 4 on Page 86 of the original paper~\cite{rao1999predictive}. The original implementation also contains terms that account for the prior probability distributions of $\mathbf{r}$ and $U$; these terms can be equated to regularization terms, and thus we omit them for the sake of simplicity.

Equation~\ref{eq_energy} represents the overall error, calculated as sum of the mean squared errors across the hierarchy of the network. It should be noted that we use this same objective function ($-E$) to train the feedback weights of our networks.

As stated above, the predictive coding dynamics aim to modify neural representations $\mathbf{r}$ so as to minimize the error $E$, i.e., differentiating the above equation: 
\begin{equation}
\label{eq6}
\frac{d\mathbf{r}}{dt} = -\frac{\partial E}{\partial \mathbf{r}} = a \cdot U^{T}\frac{\partial{f}^{T}}{\partial U\mathbf{r}} (\mathbf{I} - f(U\mathbf{r})) + b \cdot (\mathbf{r}^{td} - \mathbf{r})
\end{equation} 

Barring a regularization term, the above equation is equivalent to Equation 7 on page 86 of~\cite{rao1999predictive}. One can see that the first term in the RHS of equation~\ref{eq6} can be substituted with our error-correction term $\nabla\epsilon_0$ (see Eq.~\ref{sds}). Hence, Equation~\ref{eq6} after simultaneously expanding the LHS becomes,

\begin{equation}
 \frac{\mathbf{r}(t+dt) - \mathbf{r}(t)}{dt} =  - a_1 \cdot \nabla \epsilon_{r}(t) + b \cdot (\mathbf{r}^{td}(t) - \mathbf{r}(t))
\end{equation}

We use subscript $r$ for $\epsilon$ to emphasize that this error can be calculated at any level/stage $\mathbf{r}$ represents in a multi-layer hierarchical system, and is not restricted to just the first layer of the hierarchy. Similarly, the time resolution $dt$ can be equated to 1 timestep (of arbitrary duration) for simulations. Hence, rearranging the equation further,

\begin{equation}
    \mathbf{r}(t+1) = \underbrace{ b \cdot \mathbf{r}^{td}(t)}_{feedback} + \underbrace{(1 - b) \mathbf{r}(t)}_{memory} - \underbrace{a_{1} \nabla \epsilon_{r}(t)}_{error-correction}
\end{equation}

In the above equation, the first term corresponds to our feedback term, the second term corresponds to our memory term and the last term corresponds to our feedforward error-correction term. That is, exchanging constants to match our notation:

\begin{equation}
\label{last_eq}
    \mathbf{r}(t+1) =  \underbrace{\dashedbox{feedforward}}_{feedforward} + \underbrace{ \lambda \cdot \mathbf{r}^{td}(t)}_{feedback} + \underbrace{(1 - \lambda) \mathbf{r}(t)}_{memory} - \underbrace{a_{1} \nabla \epsilon_{r}(t)}_{error-correction}
\end{equation}

This can be directly compared to our main Equation~\ref{eq:pc_equation}.

Equation~\ref{last_eq} also highlights the fact that our approach has an extra feedforward term that is not present in the original Rao and Ballard proposal. We believe that such a modification allows for rethinking the role of error-correction in network dynamics; where error-correction constituted the predominant mode of feed-forward communication in the Rao and Ballard implementation, it plays a more supporting role in our implementation, iteratively correcting the errors made by the feedforward convolutional layers. We empirically found that the feedforward term helped to improve the stability of the training. Interestingly, a common criticism of predictive coding lies in its inability to explain the dominance of feedforward brain activity compared to prediction error signals~\cite{heilbron2018great,aitchison2017or}. We believe that our proposed implementation allows for a flexible modulation of these two terms, and thus systematic investigation of these factors--as done in~\citep{alamia2021role}.

From a practical perspective, we expect that our framework can be readily used by both proponents and opponents of the predictive coding theory. Setting the feedforward term $\beta$ equal to zero produces a pure predictive coding network as proposed in Rao and Ballard~\cite{rao1999predictive}. Alternatively, one can set the error-correction term $\alpha$ equal to zero to study a bidirectional network with feedback and feedforward drives, in the style of Heeger~\cite{heeger2017theory}. The framework has been implemented such that the basic update rule (as class \texttt{Pcoder} in the package) is easily adaptable, allowing one to try other complex interactions between these terms; for example, one could easily include multiplicative interactions between feedback and feedforward terms to emulate forms of biased competition (see~\cite{spratling2017hierarchical,spratling2008predictive}).  

\clearpage
\subsection{Tuning hyperparameters}
\label{apndx:tuninghps}

In addition to the fixed set of hyperparameters used in our initial experiments (Figures~\ref{fig:manifold_projection},~\ref{fig:robustness_imagenet_c}a and~\ref{fig:robustness_adversarial_attacks}), we also experimented with optimizing our hyperparameters. To tune the hyperparameters for the models, we applied two different strategies for both the models--tuning hyperparameters for the whole network vs tuning hyperparameters for each pcoder separately. After a few initial explorations on clean images, we discovered that the hyperparameters dictate where the network dynamics converge, and consequently its performance for noisy situations. This effect is characterized and investigated thoroughly in~\citep{alamia2021role}. Thus, in this study, we decide to use gaussian noise of standard deviation 0.5 to tune the hyperparameters and test it on all other types of noises from the ImageNet-C dataset.

For PVGG16, we start by fixing the value of alpha for each layer to zero and only search for $\beta_{n}$'s and $\lambda_{n}$'s. We calculate the average cross-entropy loss for 4 timesteps on 2000 images and use it as a metric for choosing the hyperparameters. The hyperparameters chosen are as follows : 

\begin{table}[H]
    \scriptsize
    \centering
    
    \caption{Values of the Hyperparameters}
    \begin{tabular}{|c|c c c|}
        \hline
        $n$ & $\beta_n$ & $\lambda_n$ & $\alpha_n$\\
        \hline
        \hline
        $1$ & $0.2$ & $0.05$ & $0.01$ \\
        \hline
        $2$ & $0.4$ & $0.10$ & $0.01$ \\
        \hline
        $3$ & $0.4$ & $0.10$ & $0.01$ \\
        \hline
        $4$ & $0.5$ & $0.10$ & $0.01$ \\
        \hline
        $5$ & $0.6$ & $0.00$ & $0.01$ \\
        \hline
    \end{tabular}
    \label{tab:hyper1}
\end{table}

For PEfficientNetB0, we take a different approach. Instead of the whole network, we start by finetuning each pcoder using the same metric (average crossentropy for 4 timesteps) on 4050 images. We then combine all hyperparameters found for each pcoder. The hyperparameters chosen are as follows :

\begin{table}[h!]
    \scriptsize
    \centering
    
    \caption{Values of the Hyperparameters}
    \begin{tabular}{|c|c c c|}
        \hline
        $n$ & $\beta_n$ & $\lambda_n$ & $\alpha_n$\\
        \hline
        \hline
        $1$ & $0.77$ & $0.08$ & $0.01$ \\
        \hline
        $2$ & $0.76$ & $0.11$ & $0.01$ \\
        \hline
        $3$ & $0.83$ & $0.03$ & $0.01$ \\
        \hline
        $4$ & $0.94$ & $0.01$ & $0.01$ \\
        \hline
        $5$ & $0.73$ & $0.25$ & $0.01$ \\
        \hline
        $6$ & $0.81$ & $0.01$ & $0.01$ \\
        \hline
        $7$ & $0.85$ & $0.10$ & $0.01$ \\
        \hline
        $8$ & $1.0$ & $0.00$ & $0.01$ \\
        \hline
    \end{tabular}
    \label{tab:hyper2}
\end{table}

We then, calculate the mCE scores using all the 19 noises for both the networks. The CE scores for each noise are shown below : 
\clearpage

\begin{figure}[h!]

    \centering
    \begin{subfigure}{0.24\textwidth}
        \includegraphics[scale=0.25]{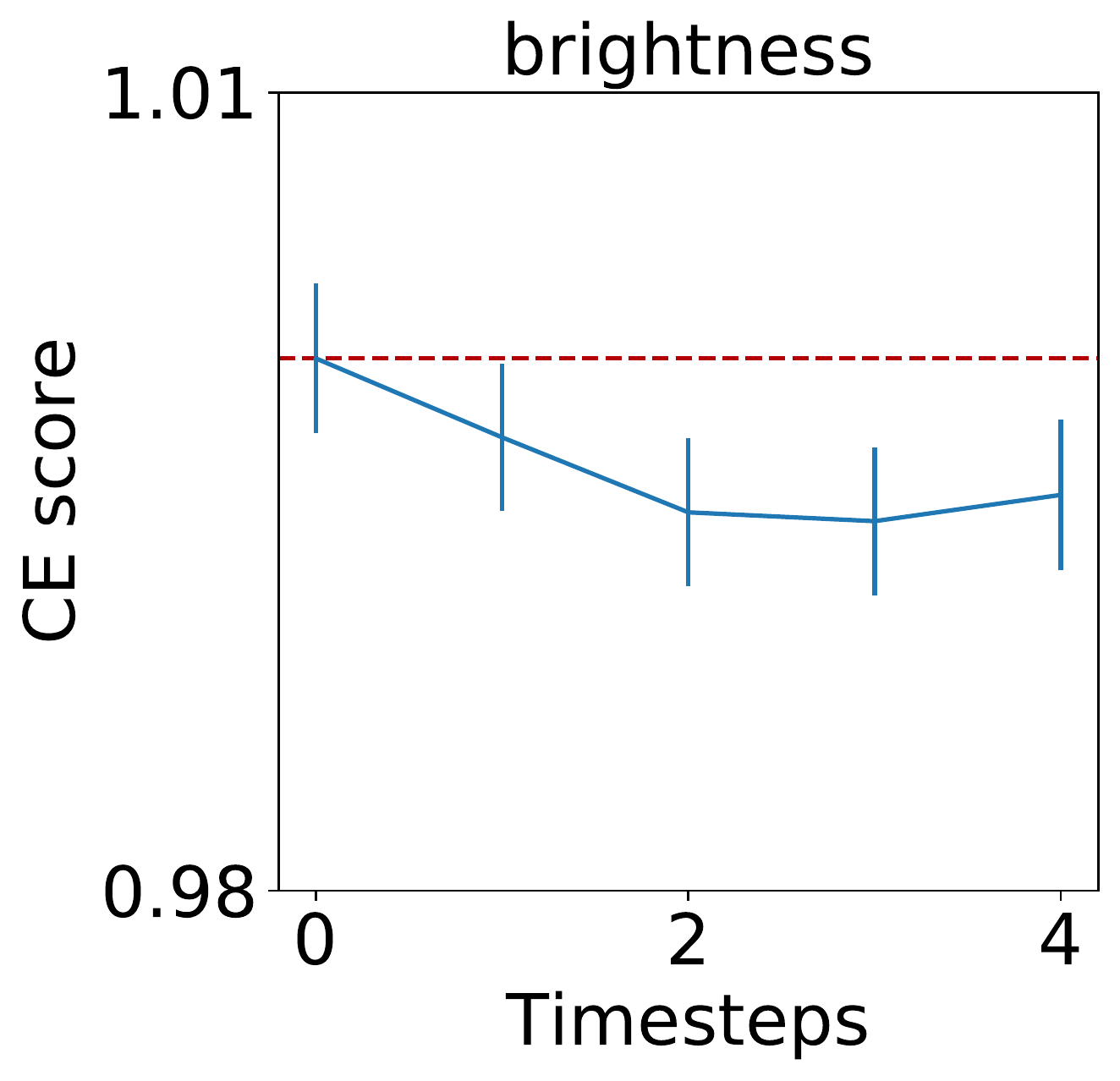}
    \end{subfigure}%
        \begin{subfigure}{0.24\textwidth}
        \includegraphics[scale=0.25]{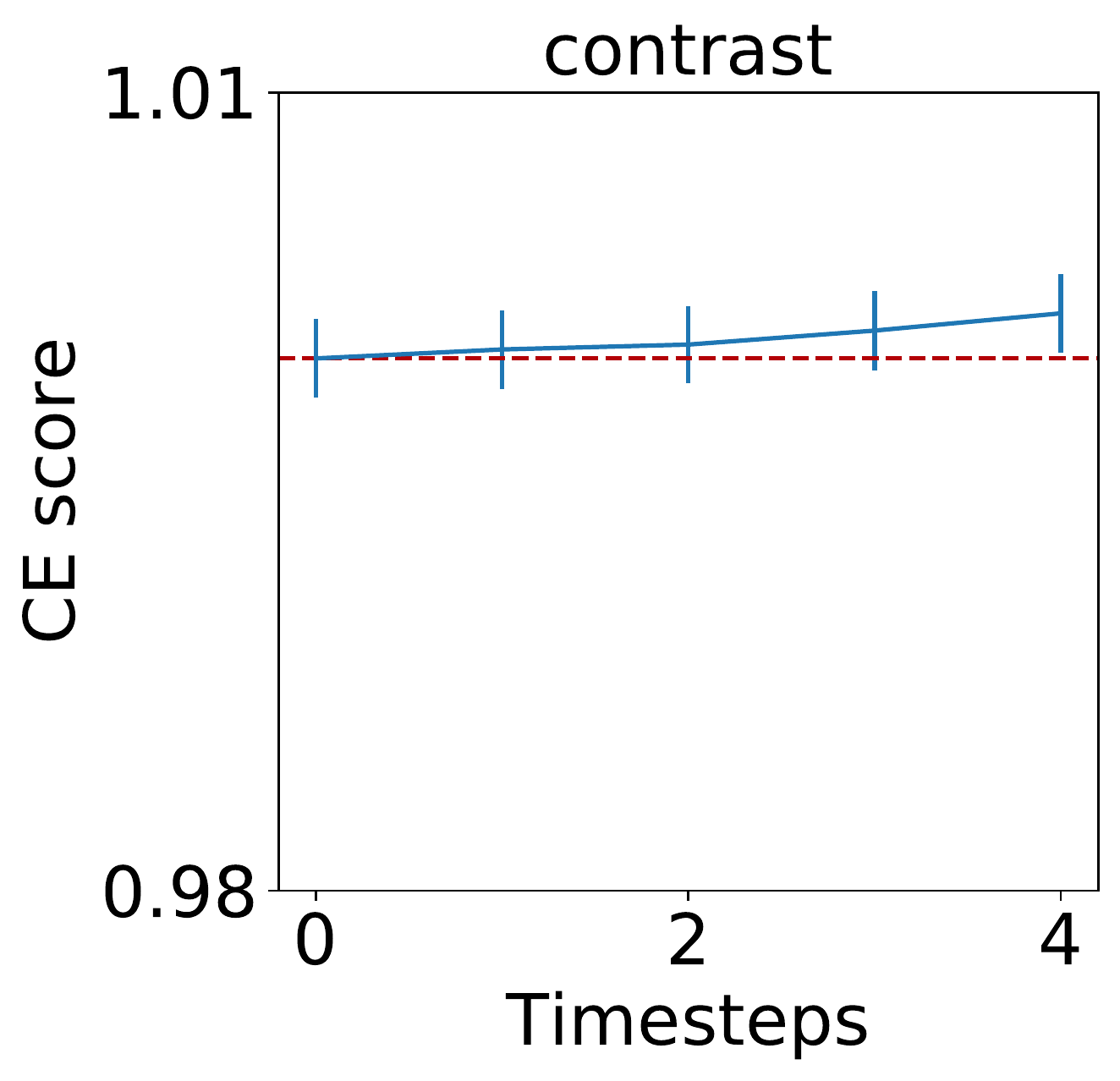}
    \end{subfigure}%
        \begin{subfigure}{0.24\textwidth}
        \includegraphics[scale=0.25]{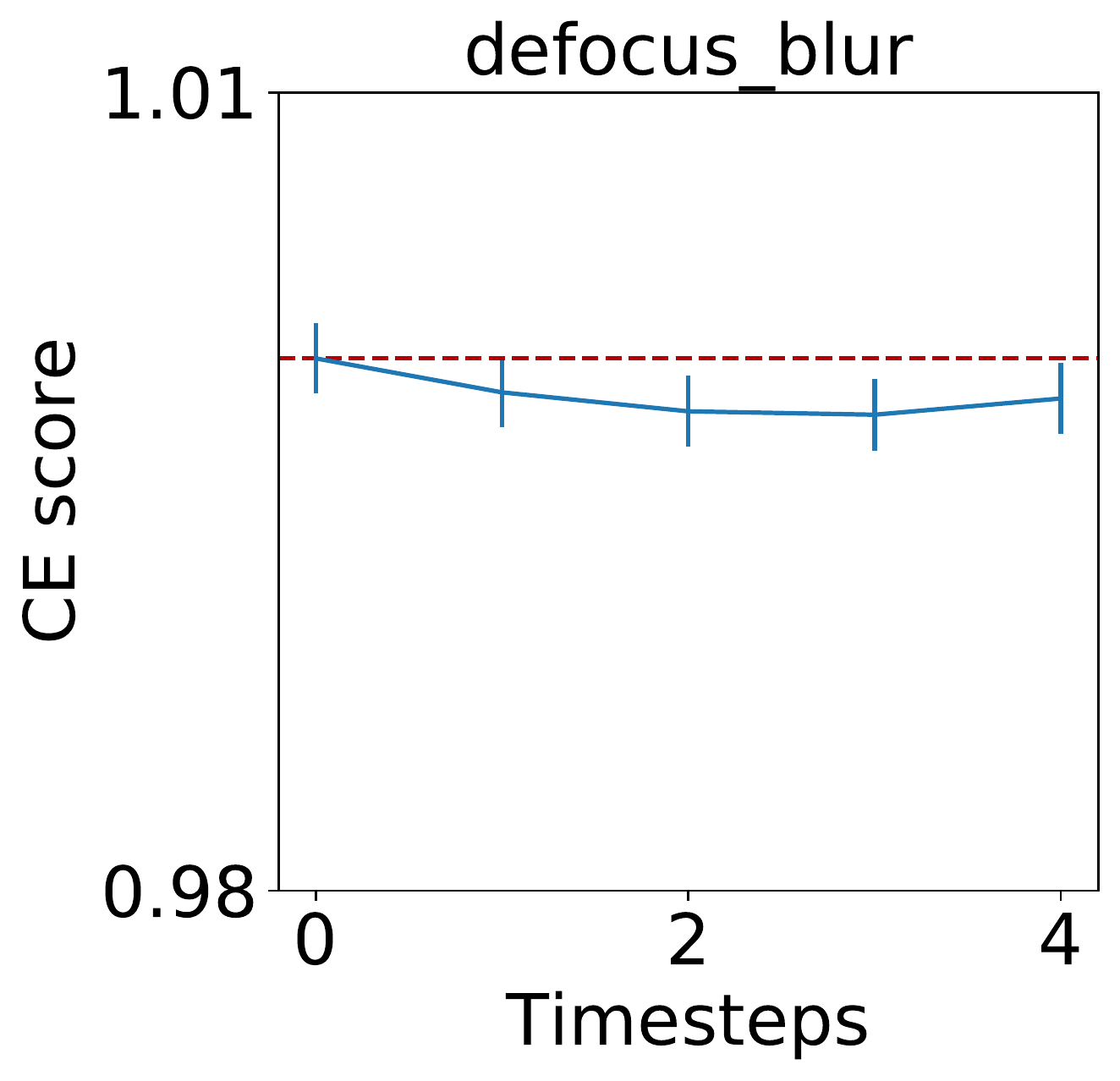}
    \end{subfigure}%
        \begin{subfigure}{0.24\textwidth}
        \includegraphics[scale=0.25]{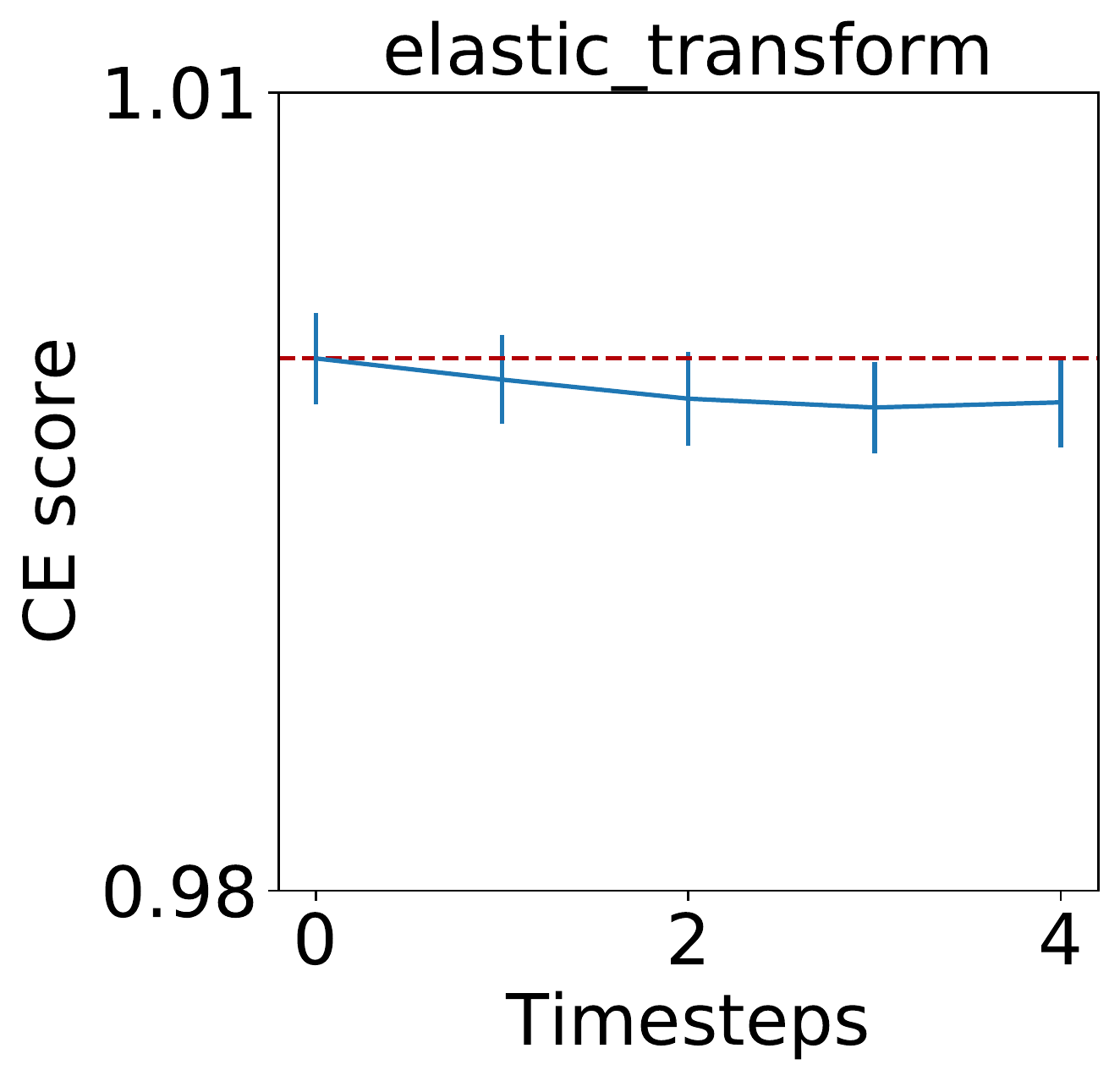}
    \end{subfigure}%

    \centering
    \begin{subfigure}{0.24\textwidth}
        \includegraphics[scale=0.25]{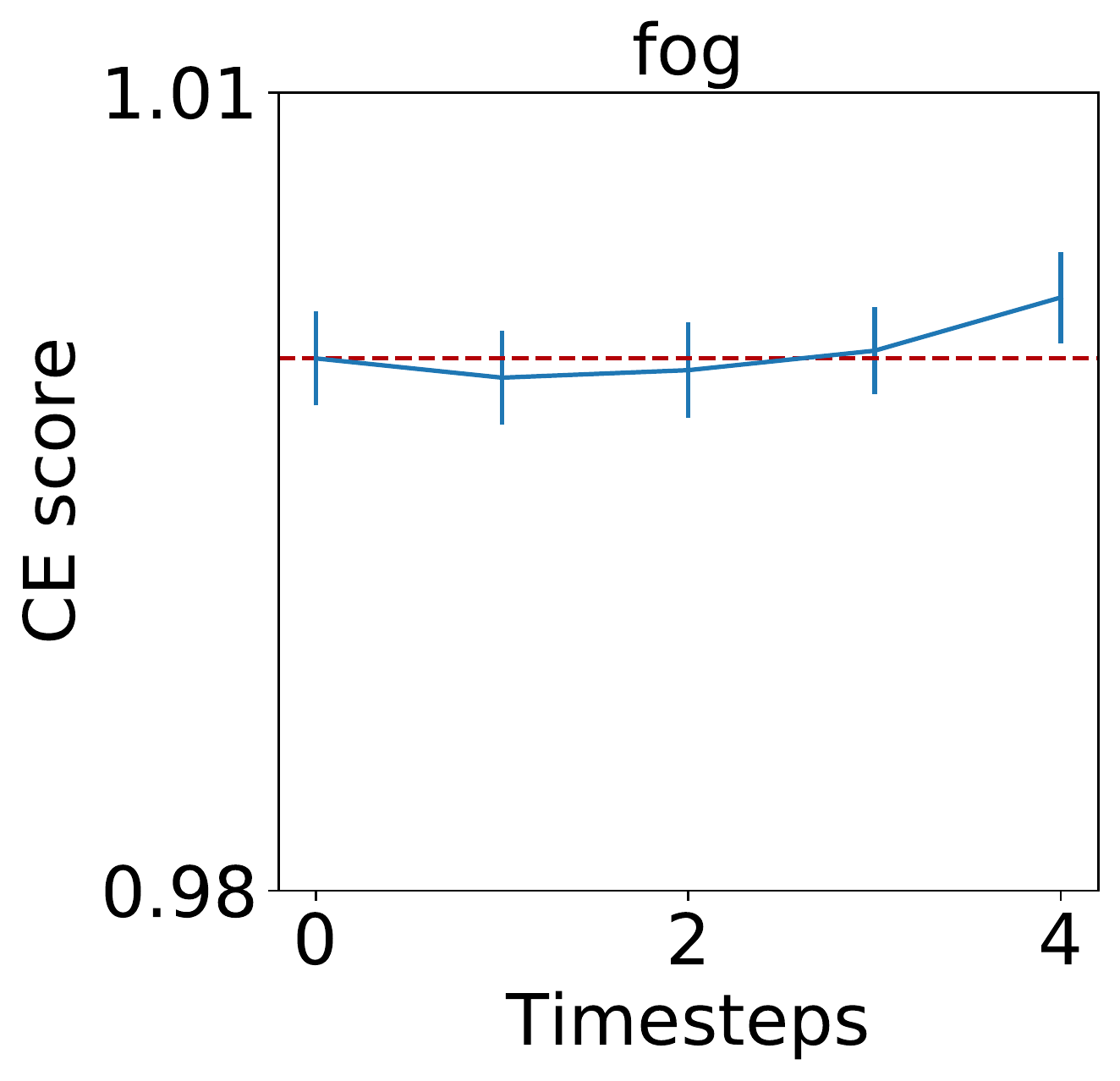}
    \end{subfigure}%
        \begin{subfigure}{0.24\textwidth}
        \includegraphics[scale=0.25]{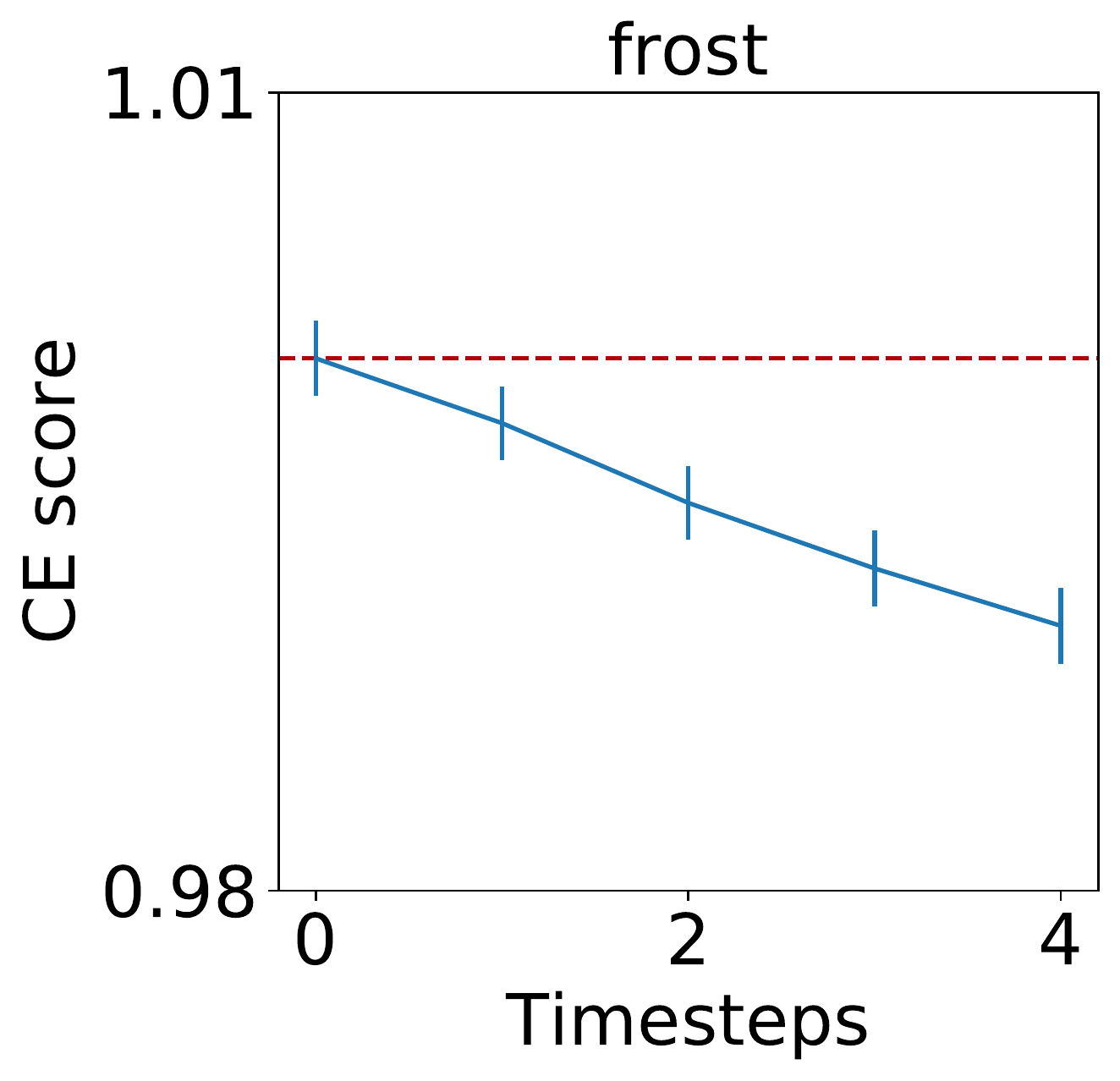}
    \end{subfigure}%
        \begin{subfigure}{0.24\textwidth}
        \includegraphics[scale=0.25]{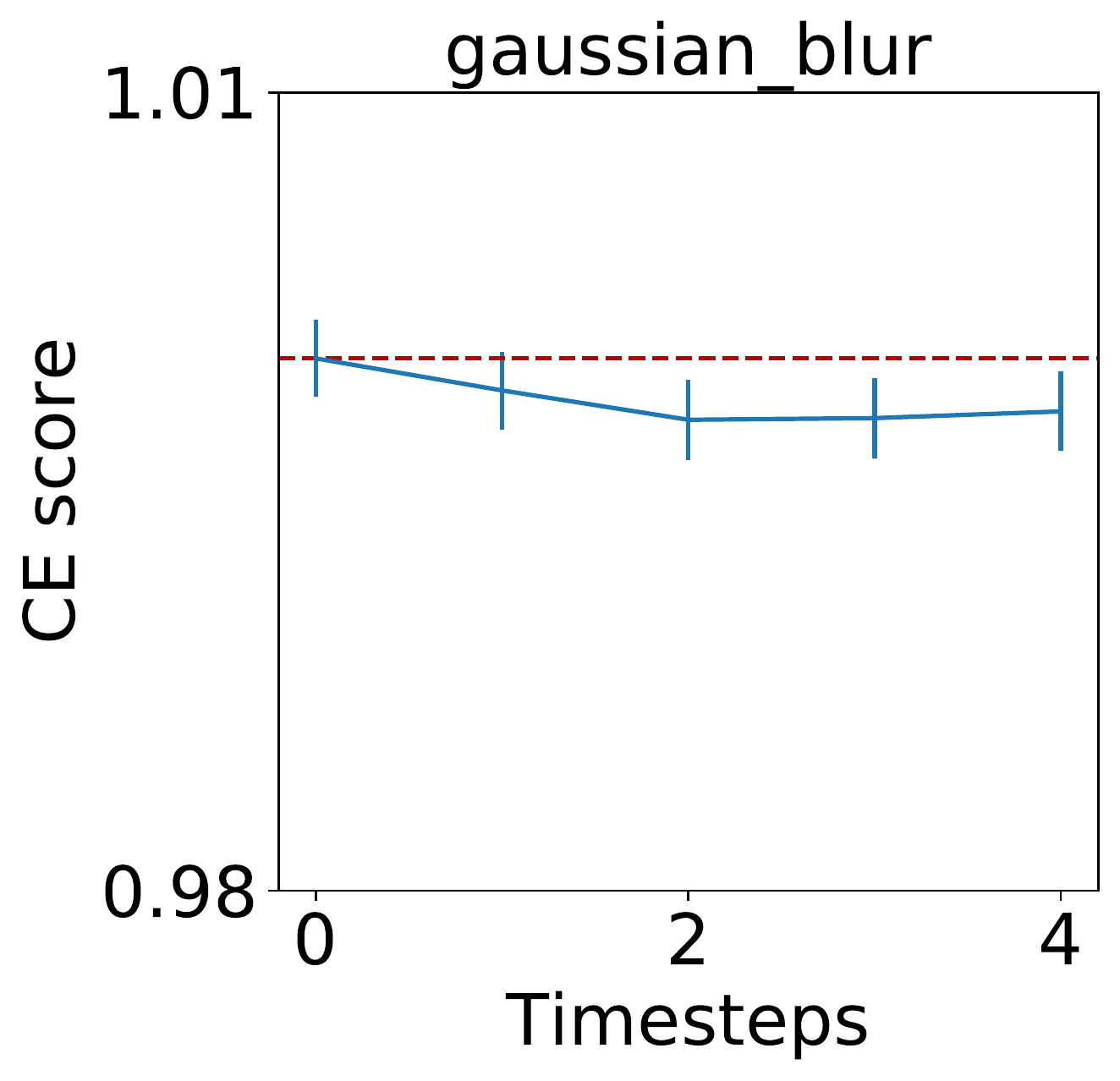}%
    \end{subfigure}%
        \begin{subfigure}{0.24\textwidth}
        \includegraphics[scale=0.25]{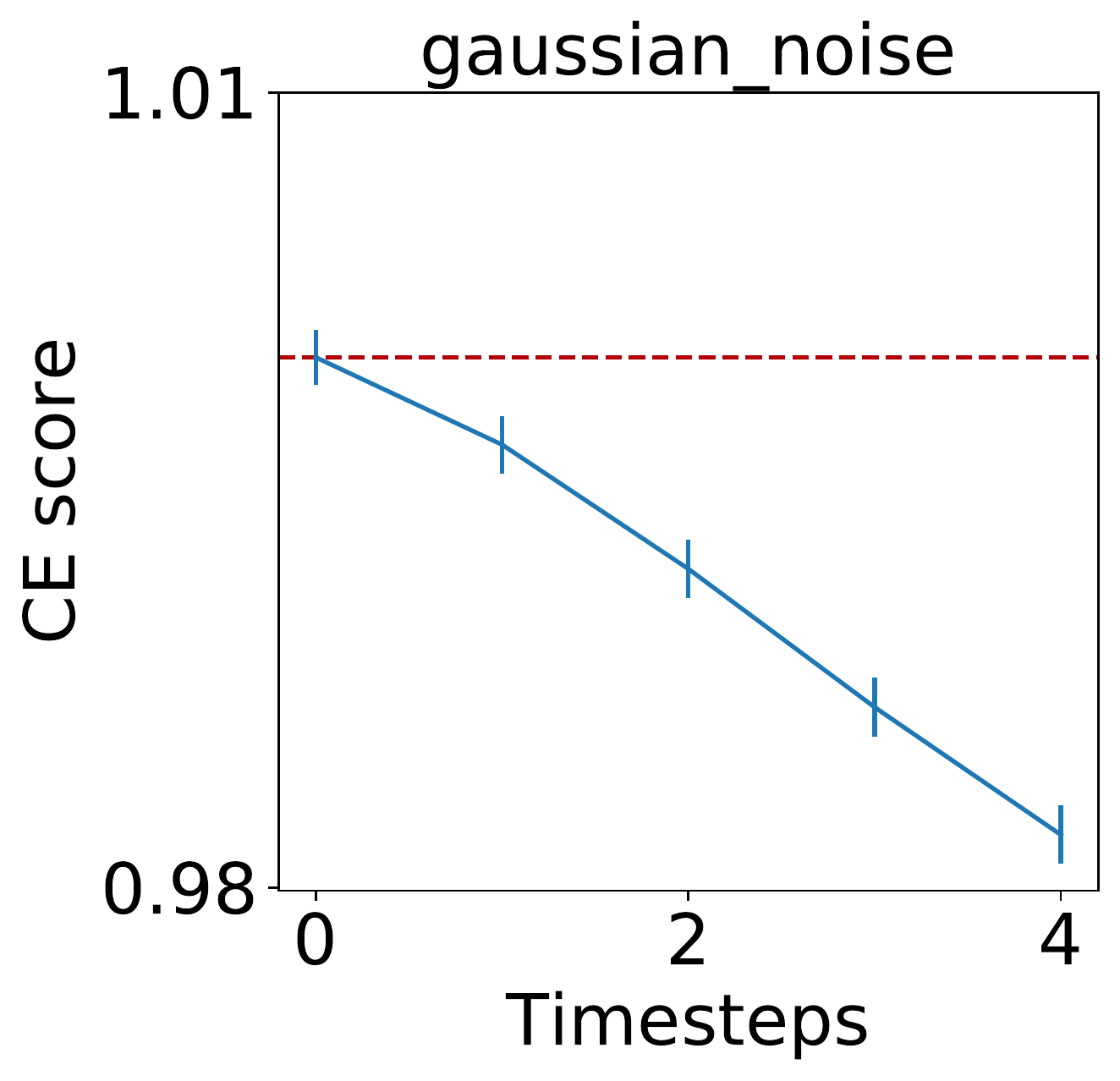}
    \end{subfigure}%
    
    \centering
    \begin{subfigure}{0.24\textwidth}
        \includegraphics[scale=0.25]{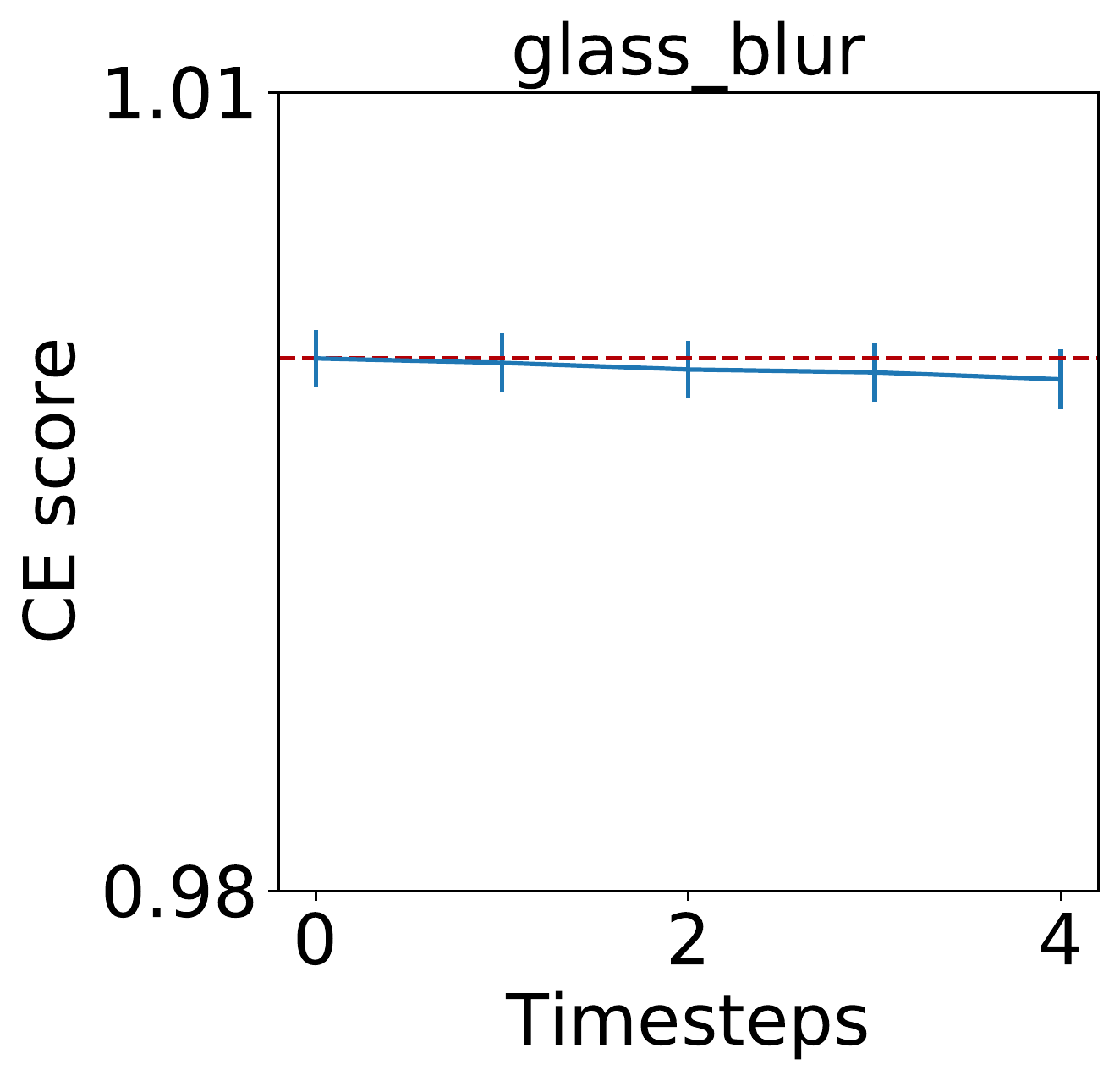}
    \end{subfigure}%
        \begin{subfigure}{0.24\textwidth}
        \includegraphics[scale=0.25]{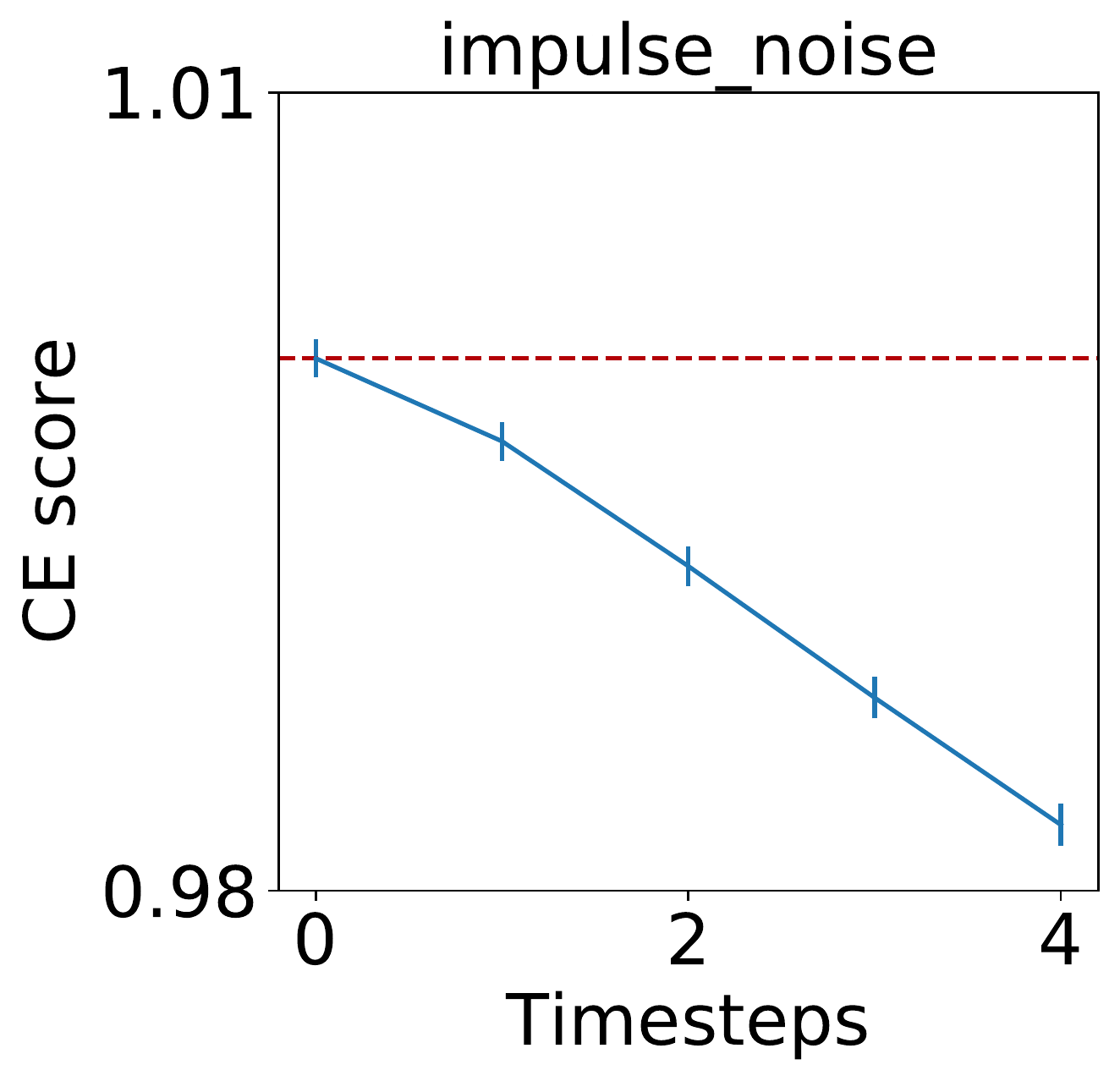}
    \end{subfigure}%
    \begin{subfigure}{0.24\textwidth}
        \includegraphics[scale=0.25]{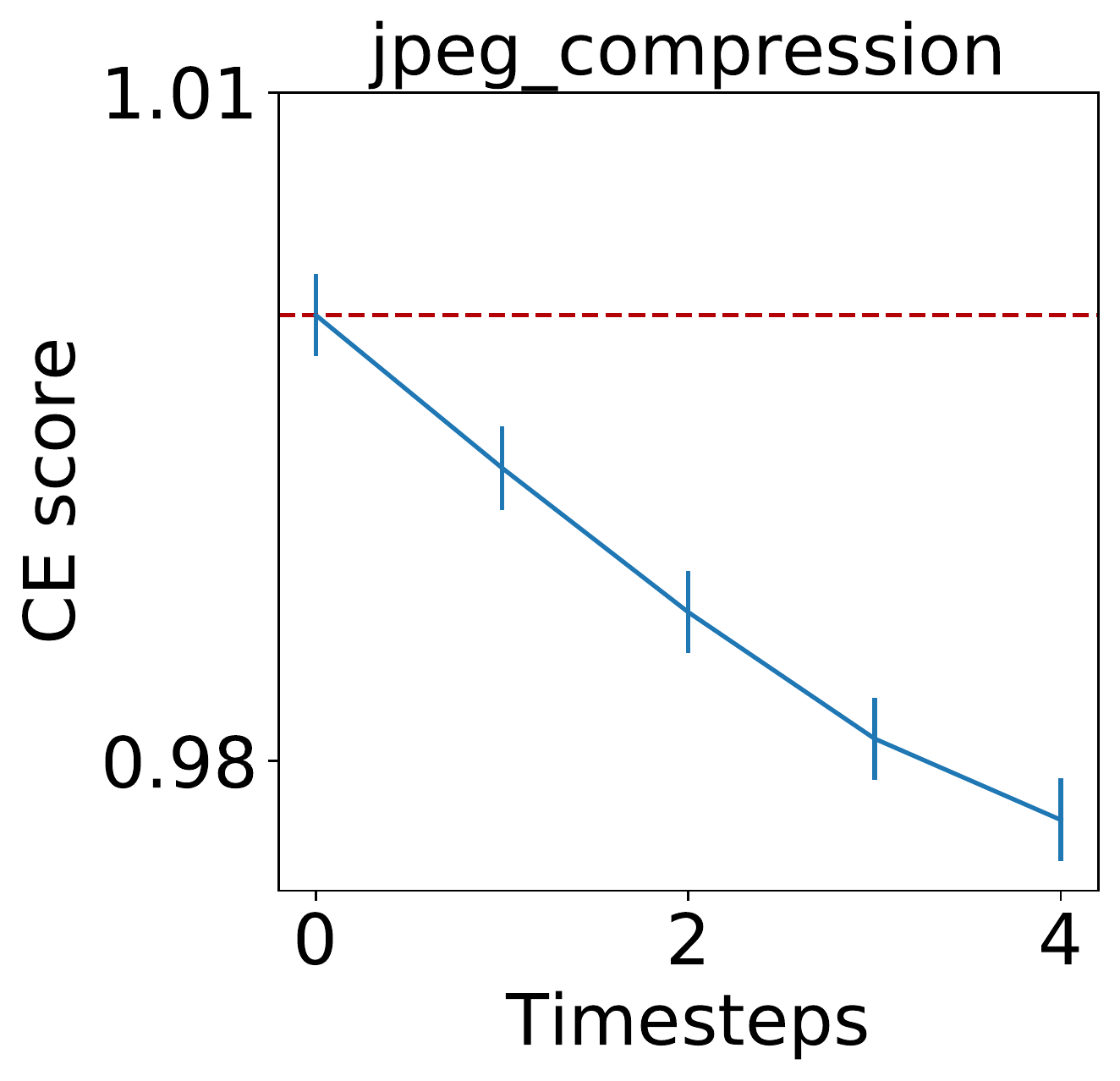}
    \end{subfigure}%
        \begin{subfigure}{0.24\textwidth}
        \includegraphics[scale=0.25]{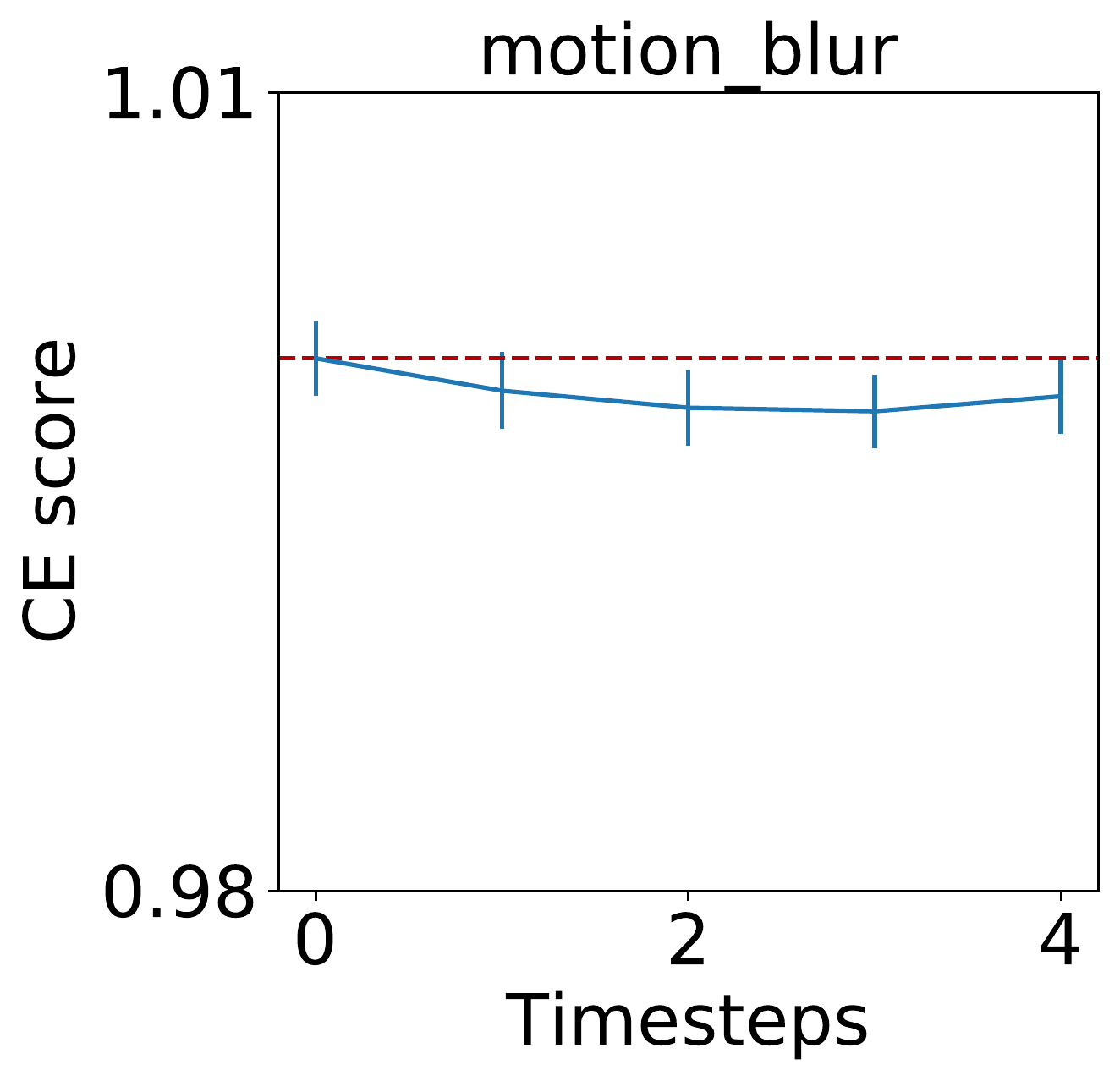}
    \end{subfigure}%

    \centering
    \begin{subfigure}{0.24\textwidth}
        \includegraphics[scale=0.25]{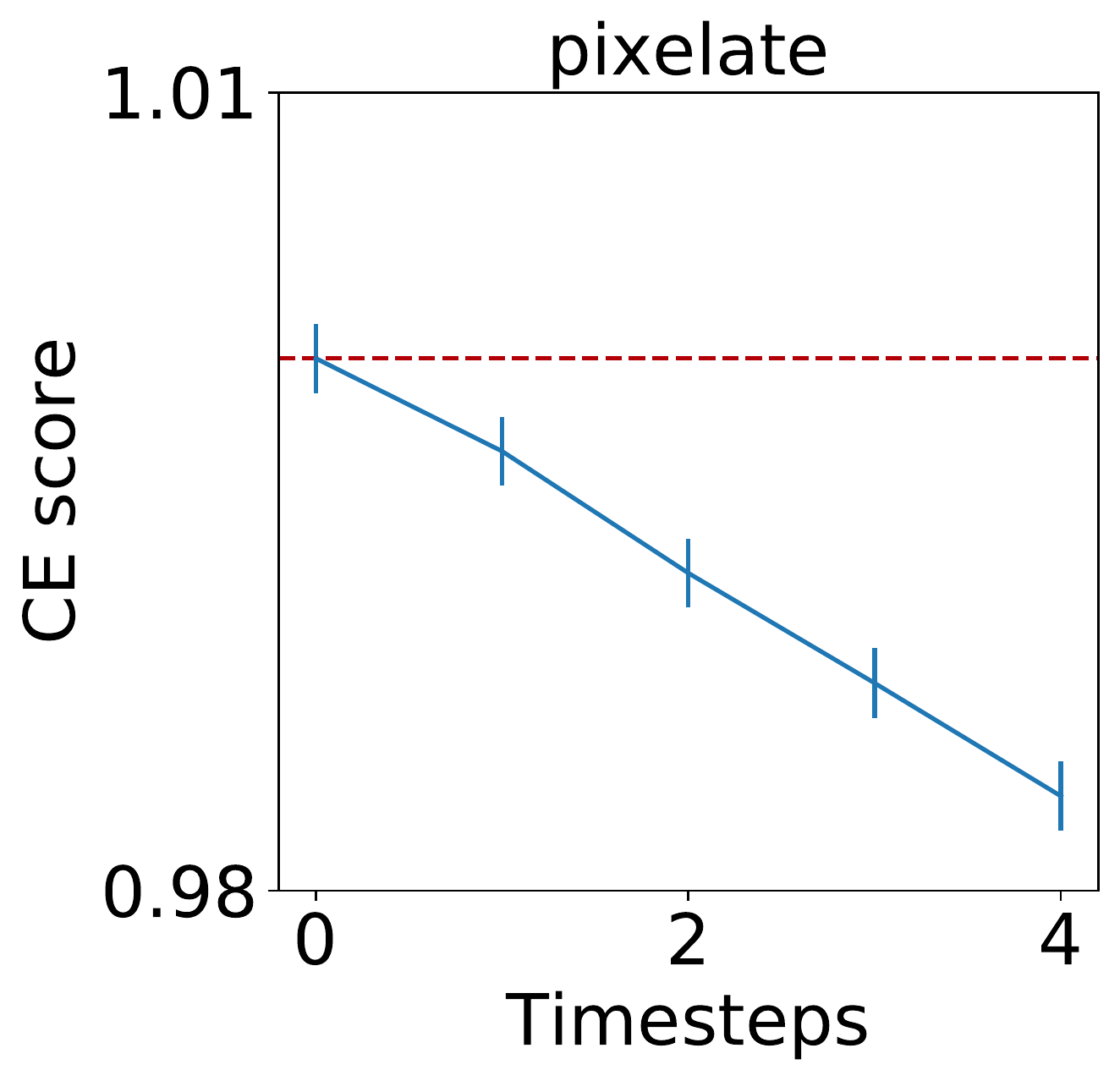}
    \end{subfigure}%
        \begin{subfigure}{0.24\textwidth}
        \includegraphics[scale=0.25]{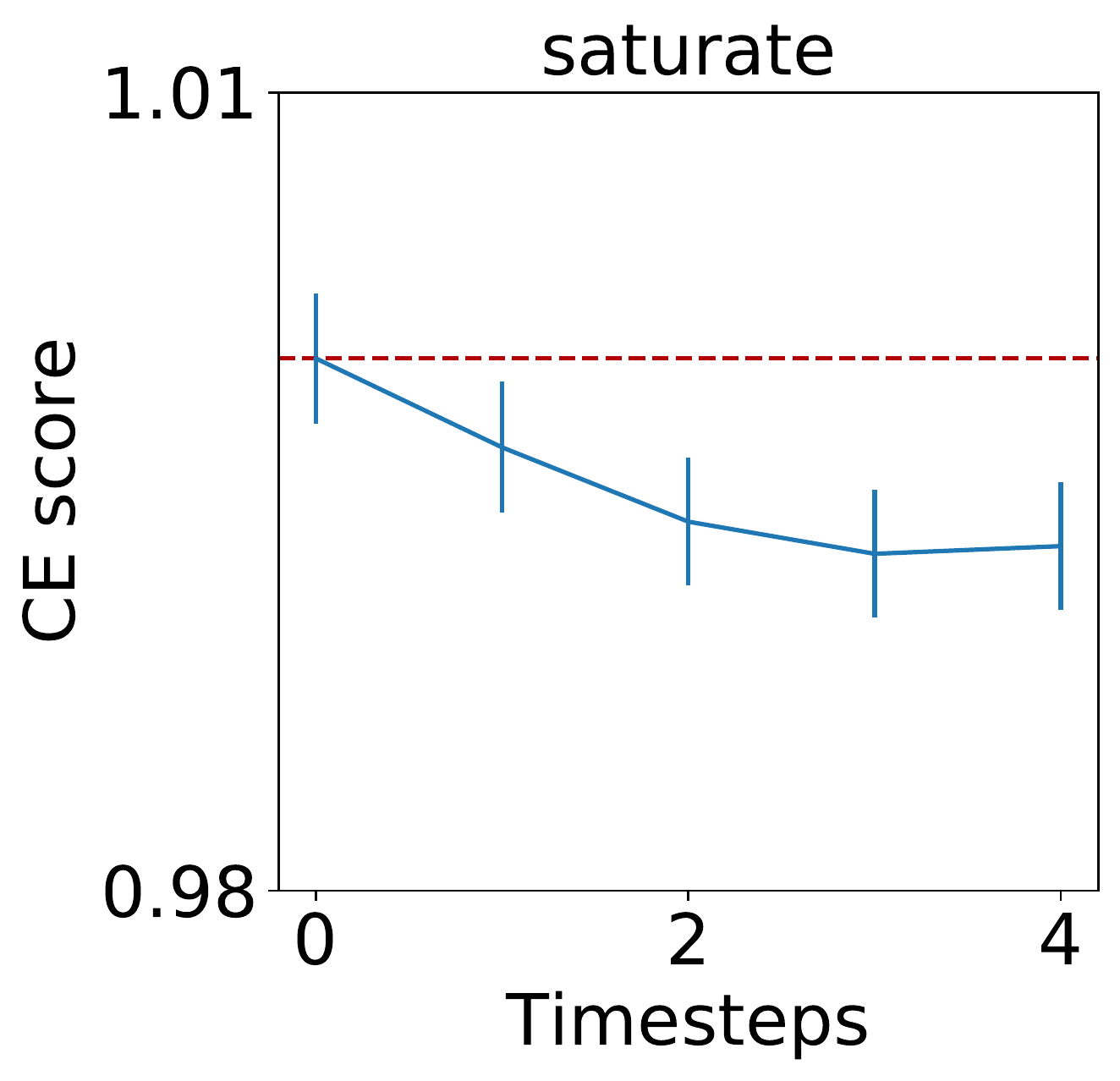}
    \end{subfigure}%
    \begin{subfigure}{0.24\textwidth}
        \includegraphics[scale=0.25]{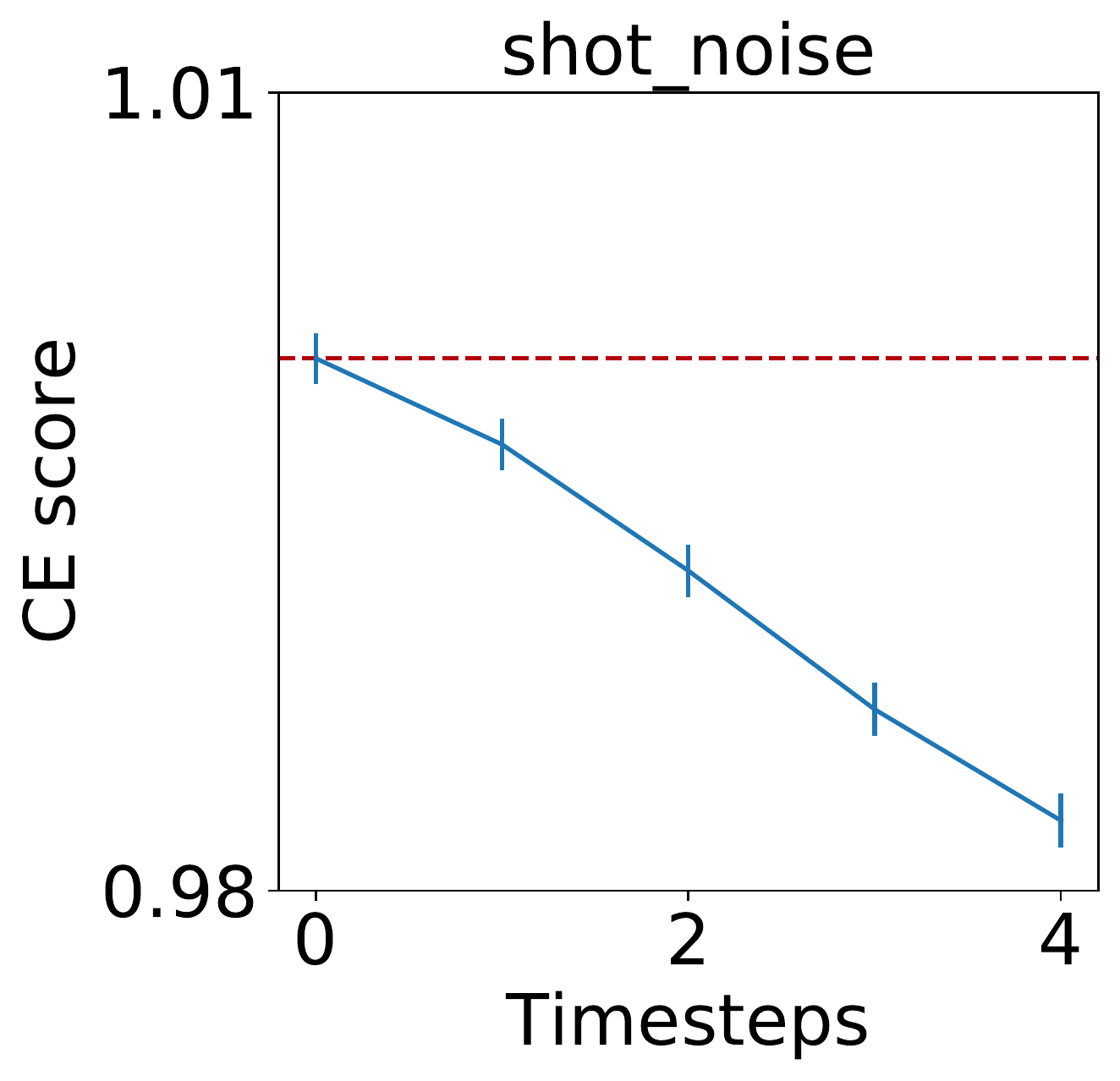}
    \end{subfigure}%
        \begin{subfigure}{0.24\textwidth}
        \includegraphics[scale=0.25]{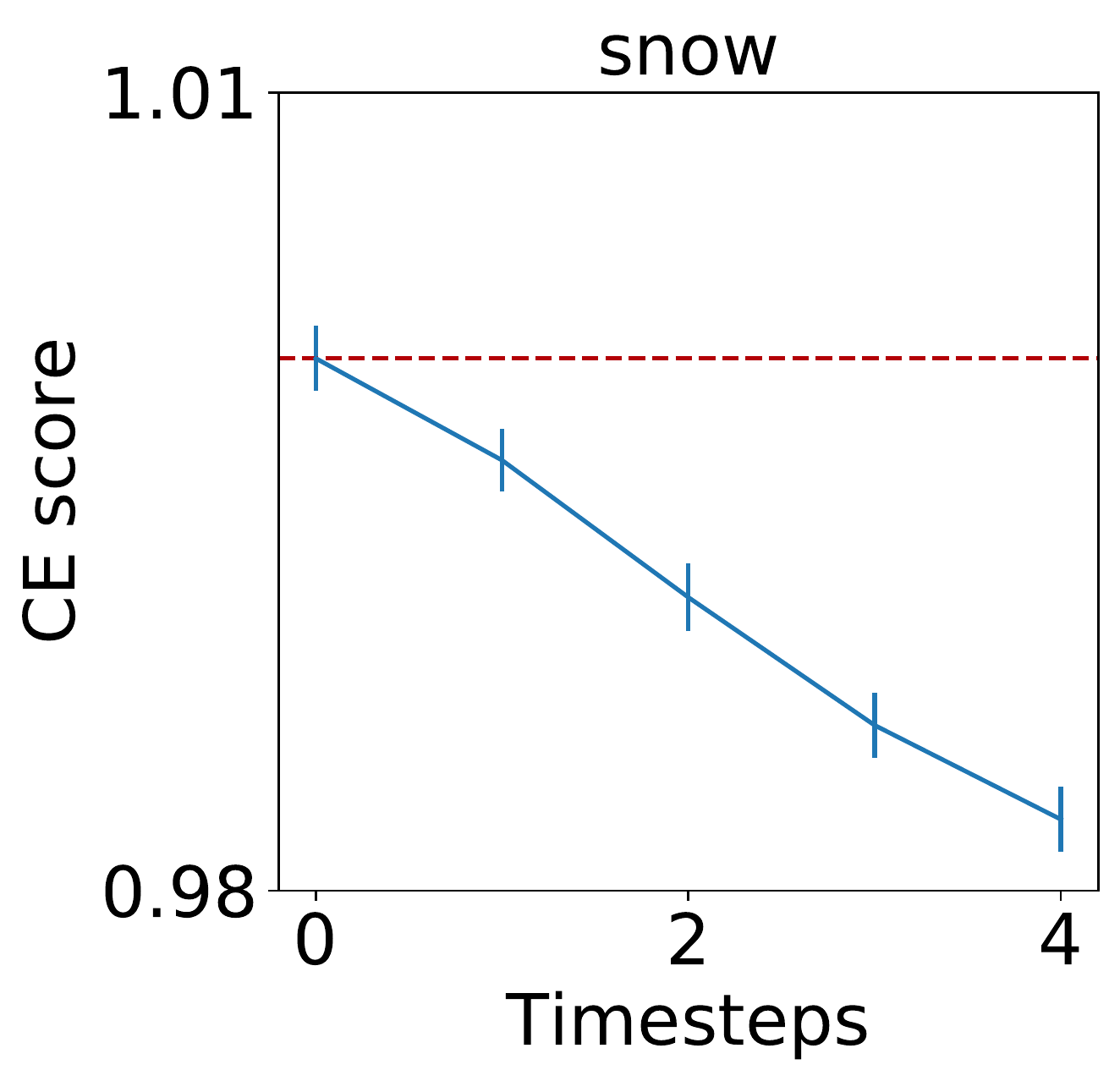}
    \end{subfigure}%
    
    \flushleft
    \hspace{0.19 cm}
    \begin{subfigure}{0.24\textwidth}
        \includegraphics[scale=0.25]{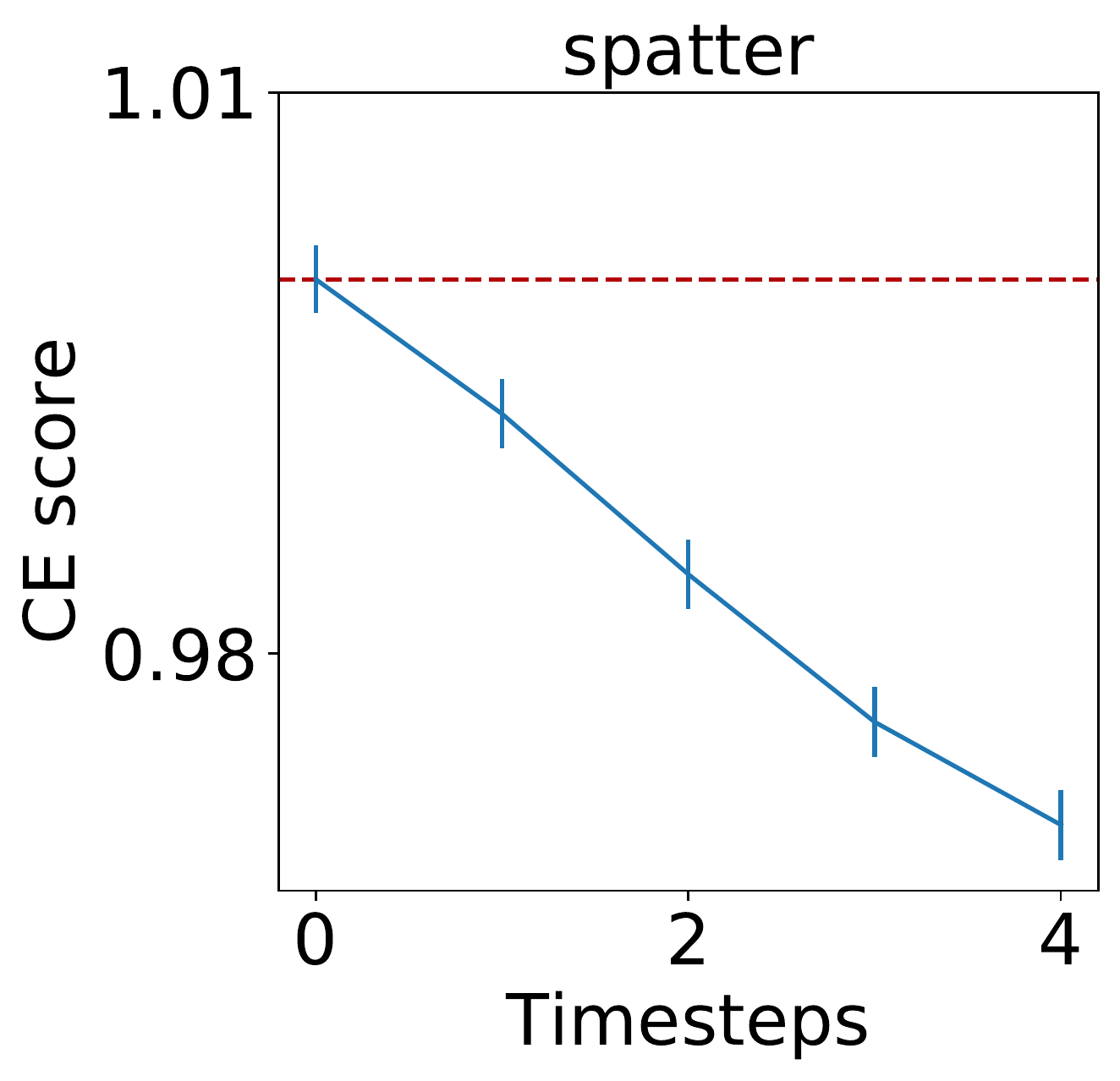}
    \end{subfigure}%
        \begin{subfigure}{0.24\textwidth}
        \includegraphics[scale=0.25]{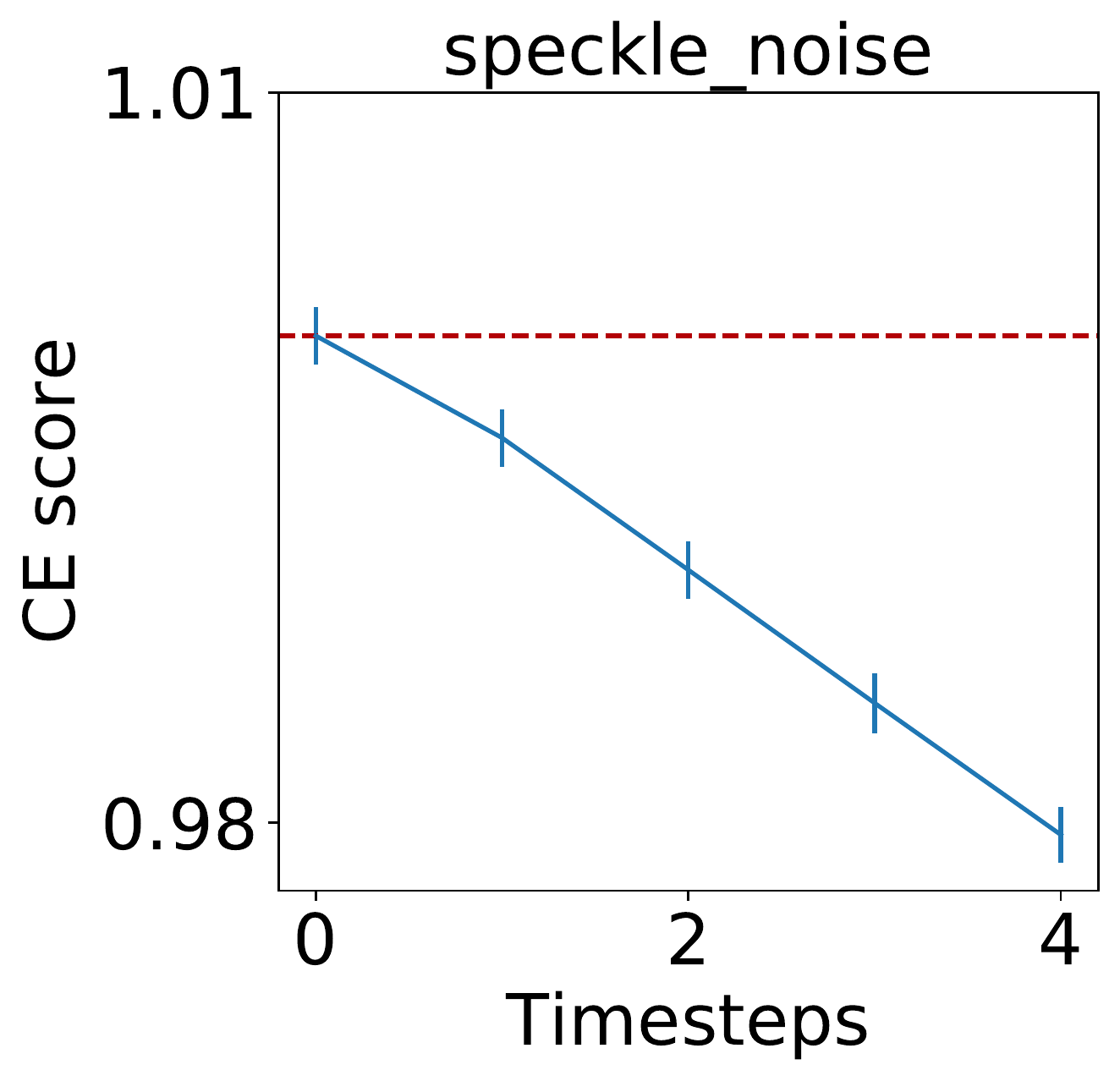}
    \end{subfigure}%
        \begin{subfigure}{0.24\textwidth}
        \includegraphics[scale=0.25]{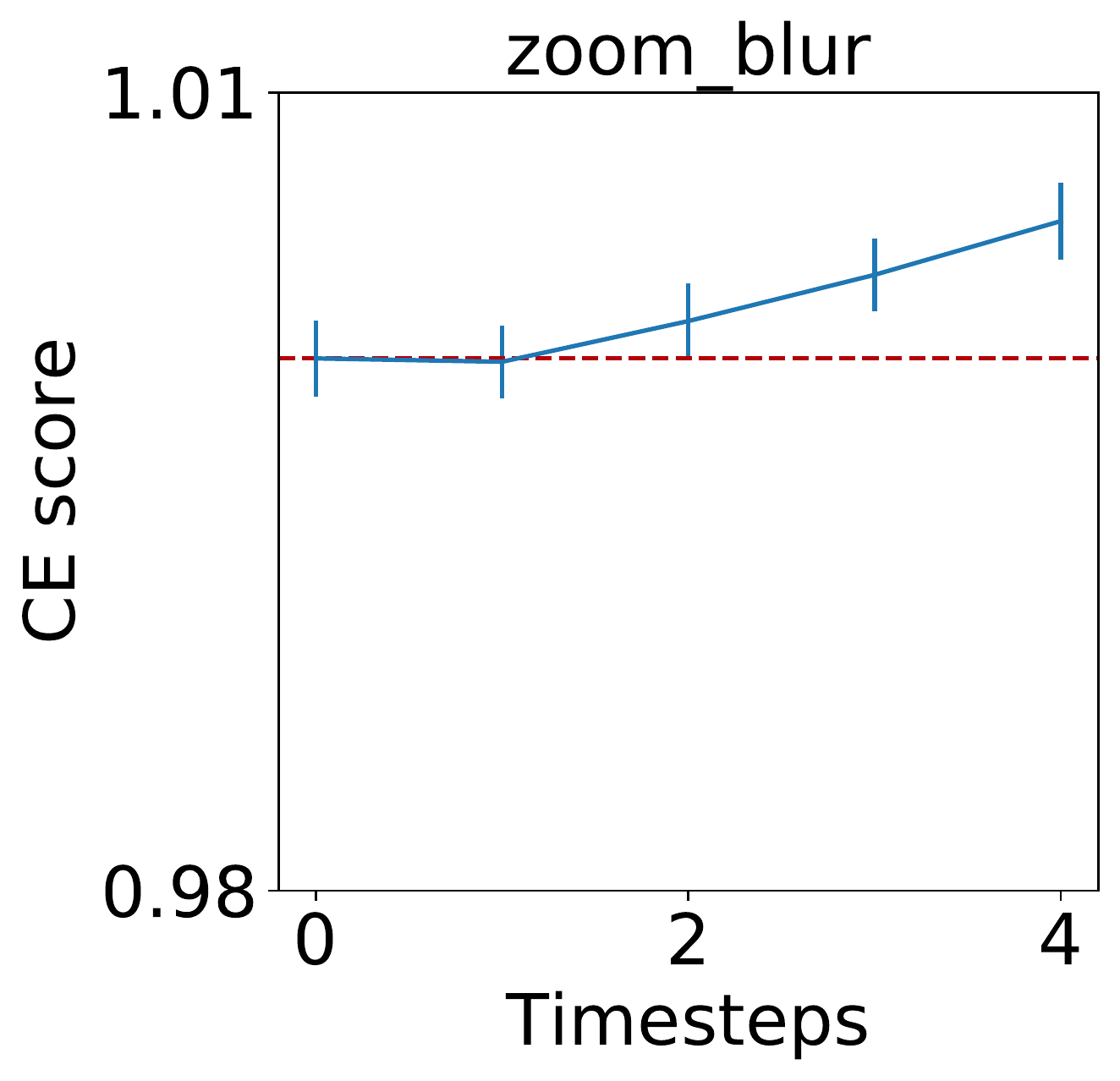}
    \end{subfigure}%

\caption{\textbf{PVGG16 (optimised) Corruption Error (CE) scores for all distortions:} The panel shows the CE scores calculated on the distorted images provided in the ImageNet-C dataset. The values are normalized with the CE score obtained for the feedforward VGG. The error bars denote the standard deviation of the means obtained from bootstrapping (resampling multiple binary populations across all severities.) }    
\label{ce_scores_appdx}
\end{figure}

\begin{figure}[h!]

    \centering
    \begin{subfigure}{0.24\textwidth}
        \includegraphics[scale=0.25]{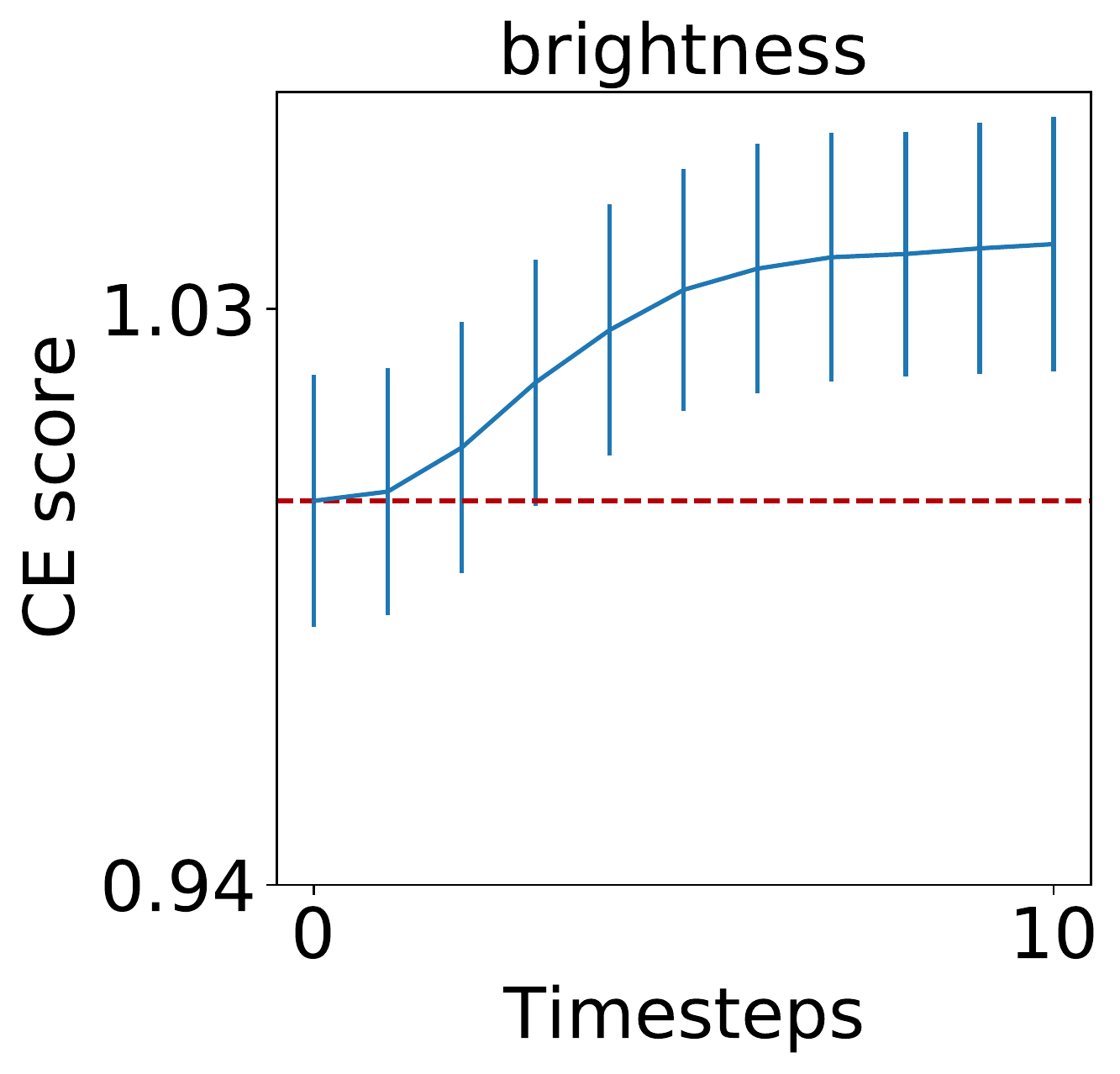}
    \end{subfigure}%
        \begin{subfigure}{0.24\textwidth}
        \includegraphics[scale=0.25]{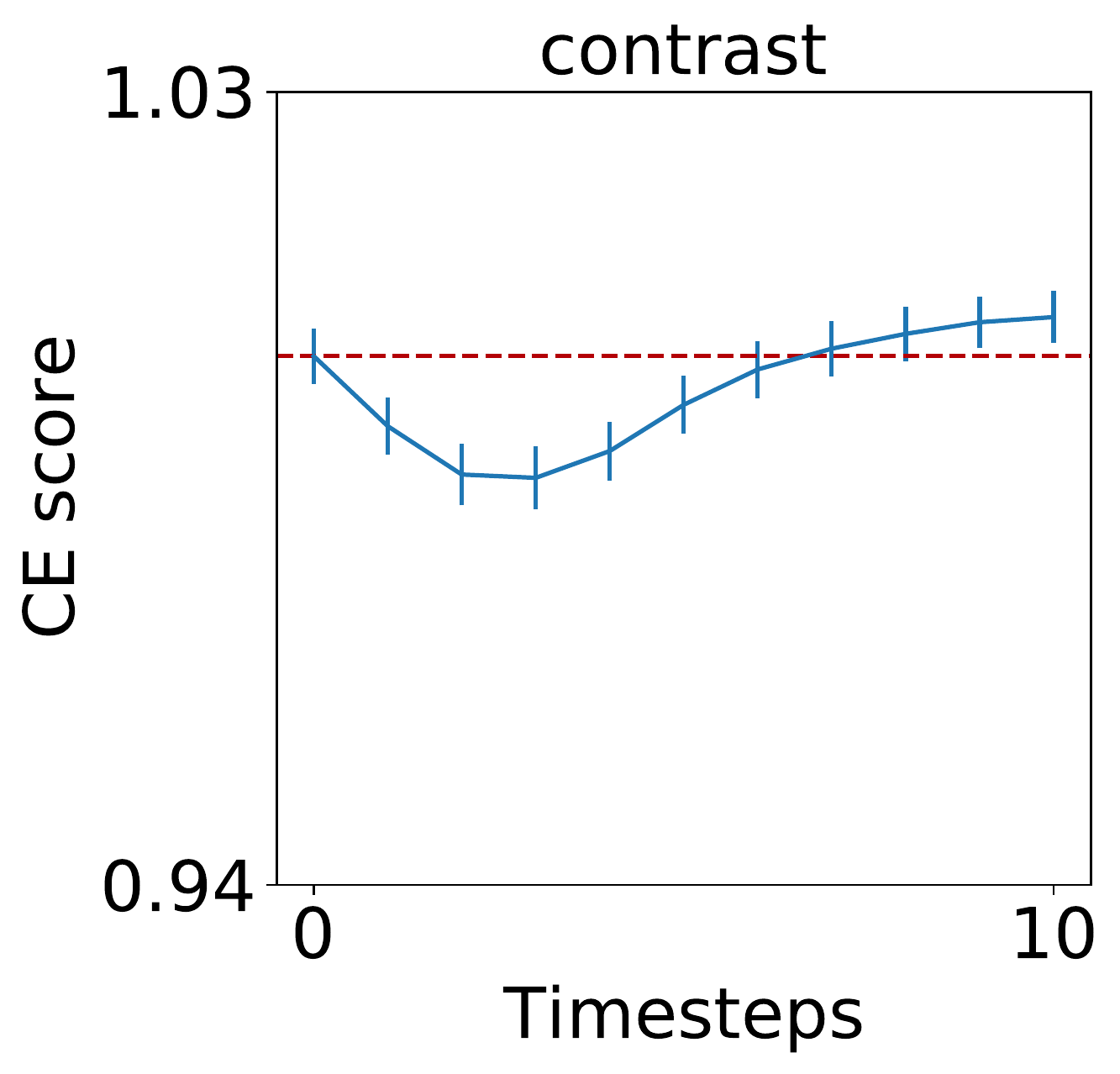}
    \end{subfigure}%
        \begin{subfigure}{0.24\textwidth}
        \includegraphics[scale=0.25]{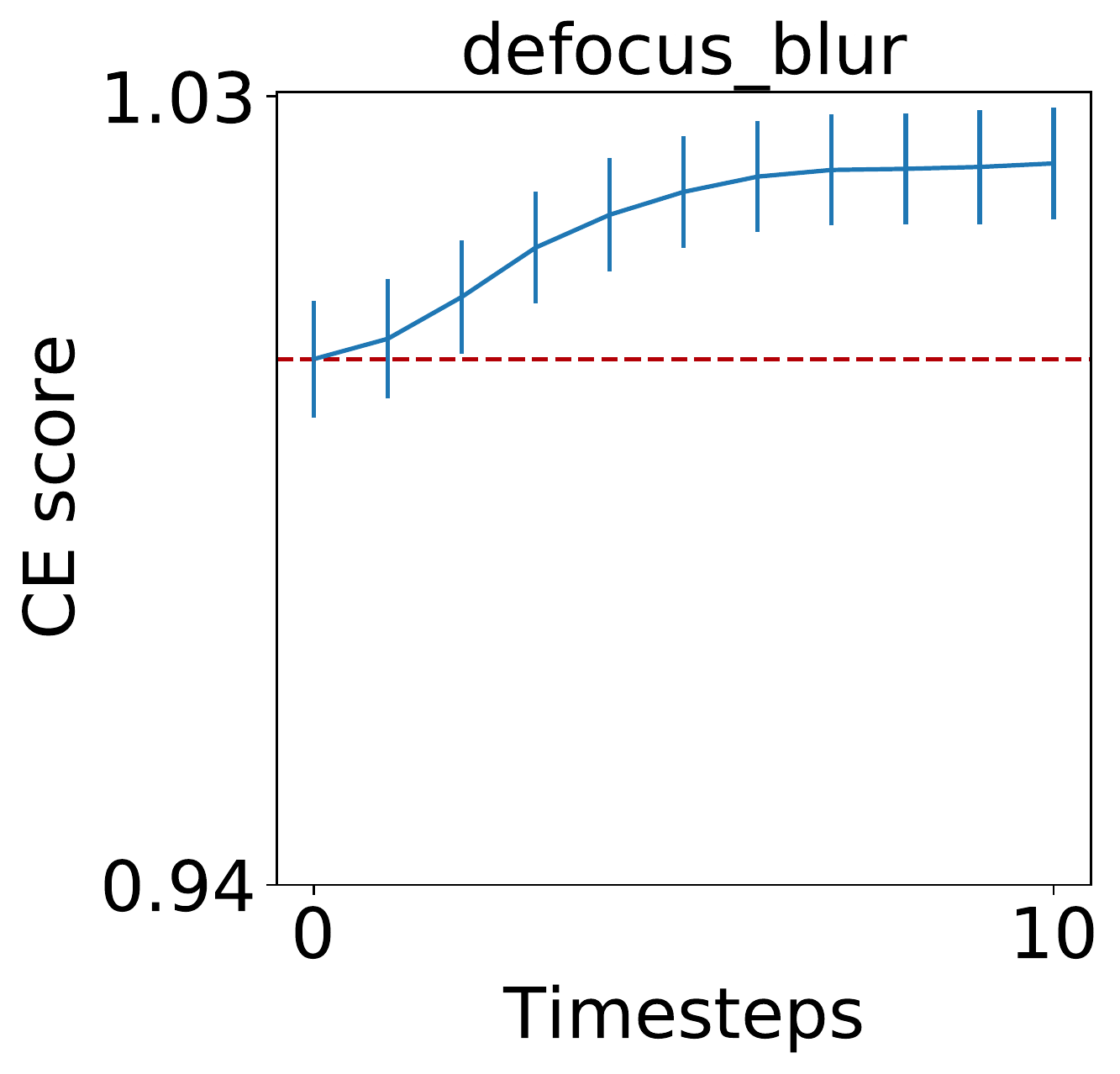}
    \end{subfigure}%
        \begin{subfigure}{0.24\textwidth}
        \includegraphics[scale=0.25]{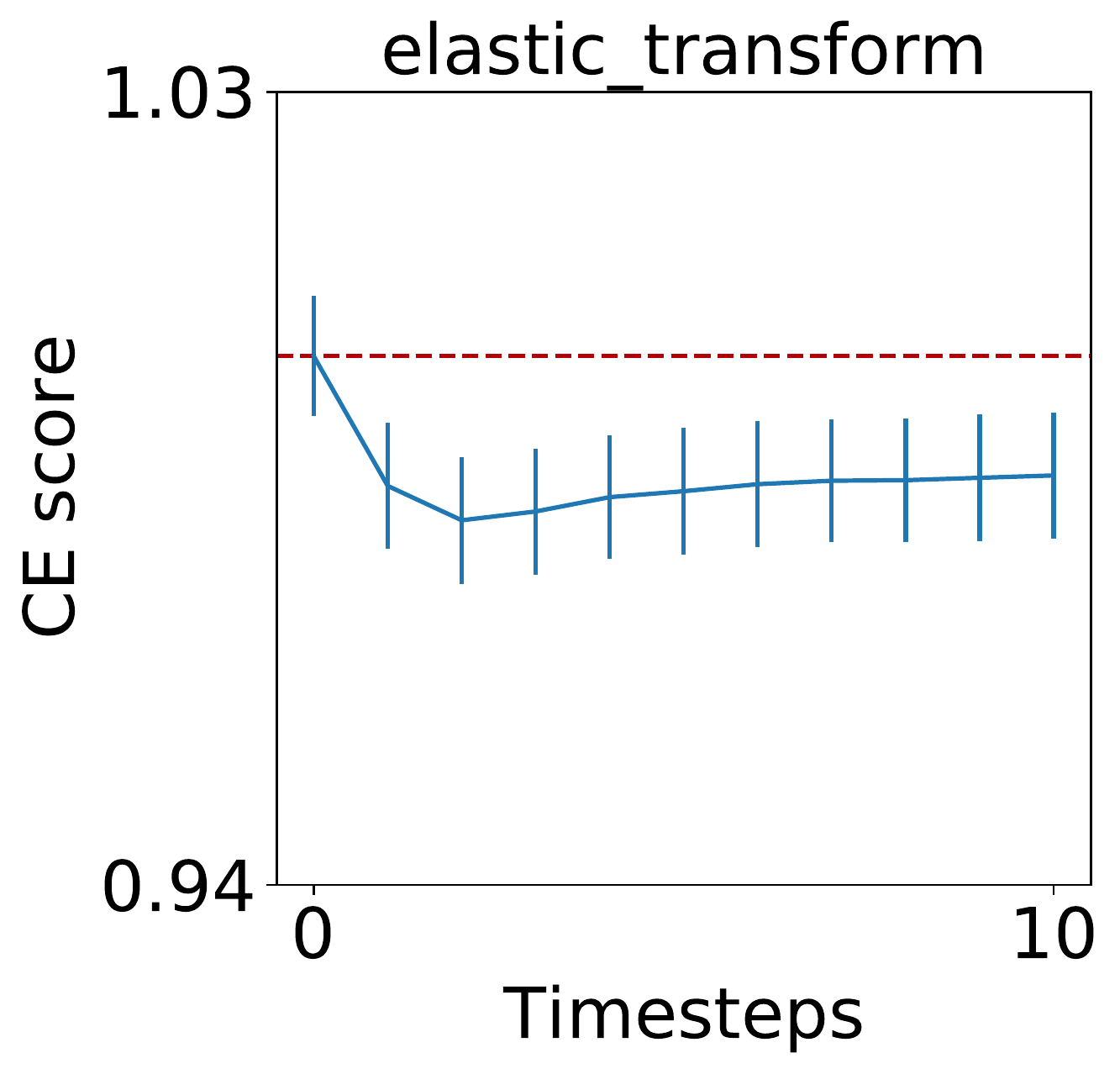}
    \end{subfigure}%

    \centering
    \begin{subfigure}{0.24\textwidth}
        \includegraphics[scale=0.25]{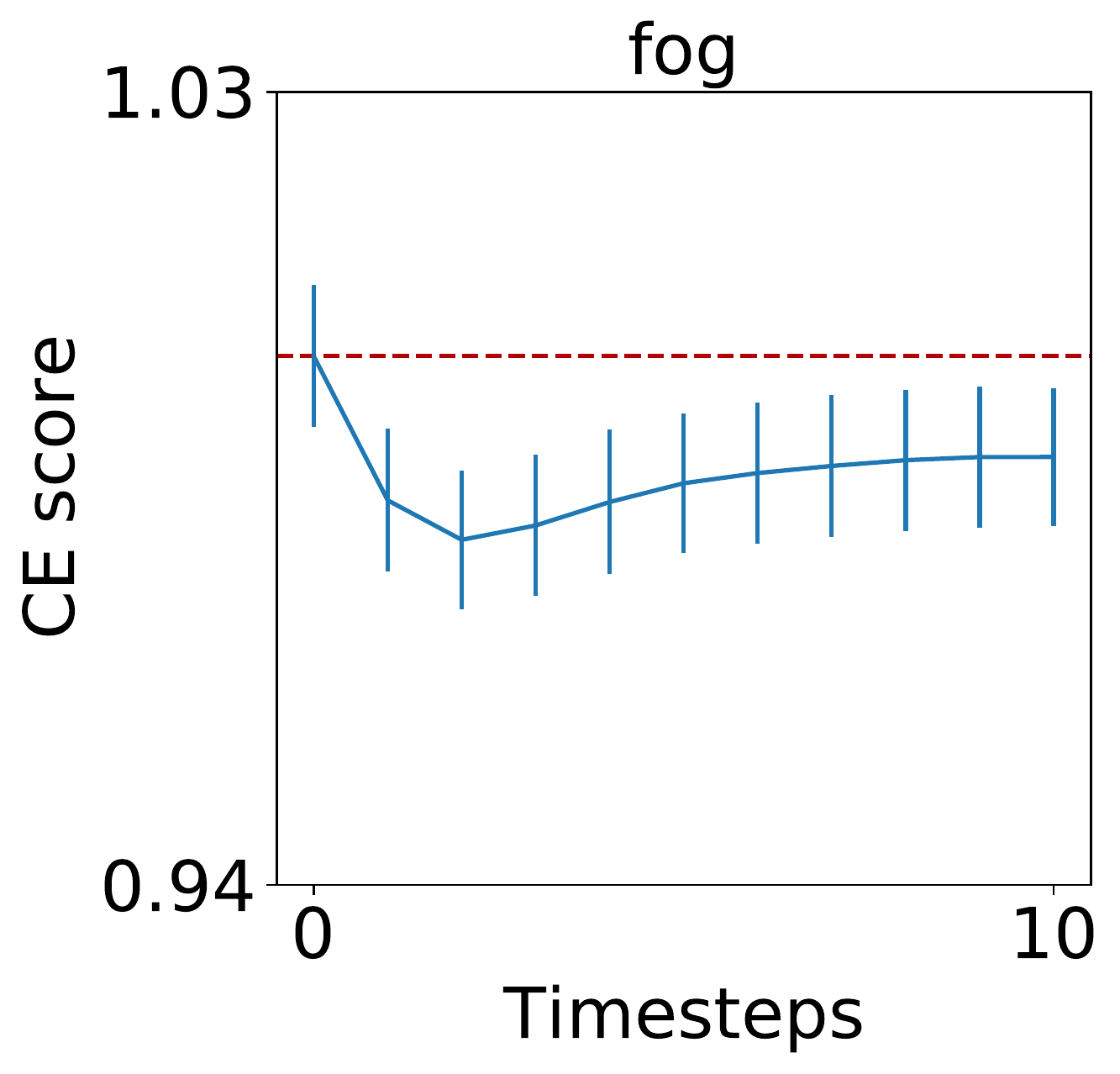}
    \end{subfigure}%
        \begin{subfigure}{0.24\textwidth}
        \includegraphics[scale=0.25]{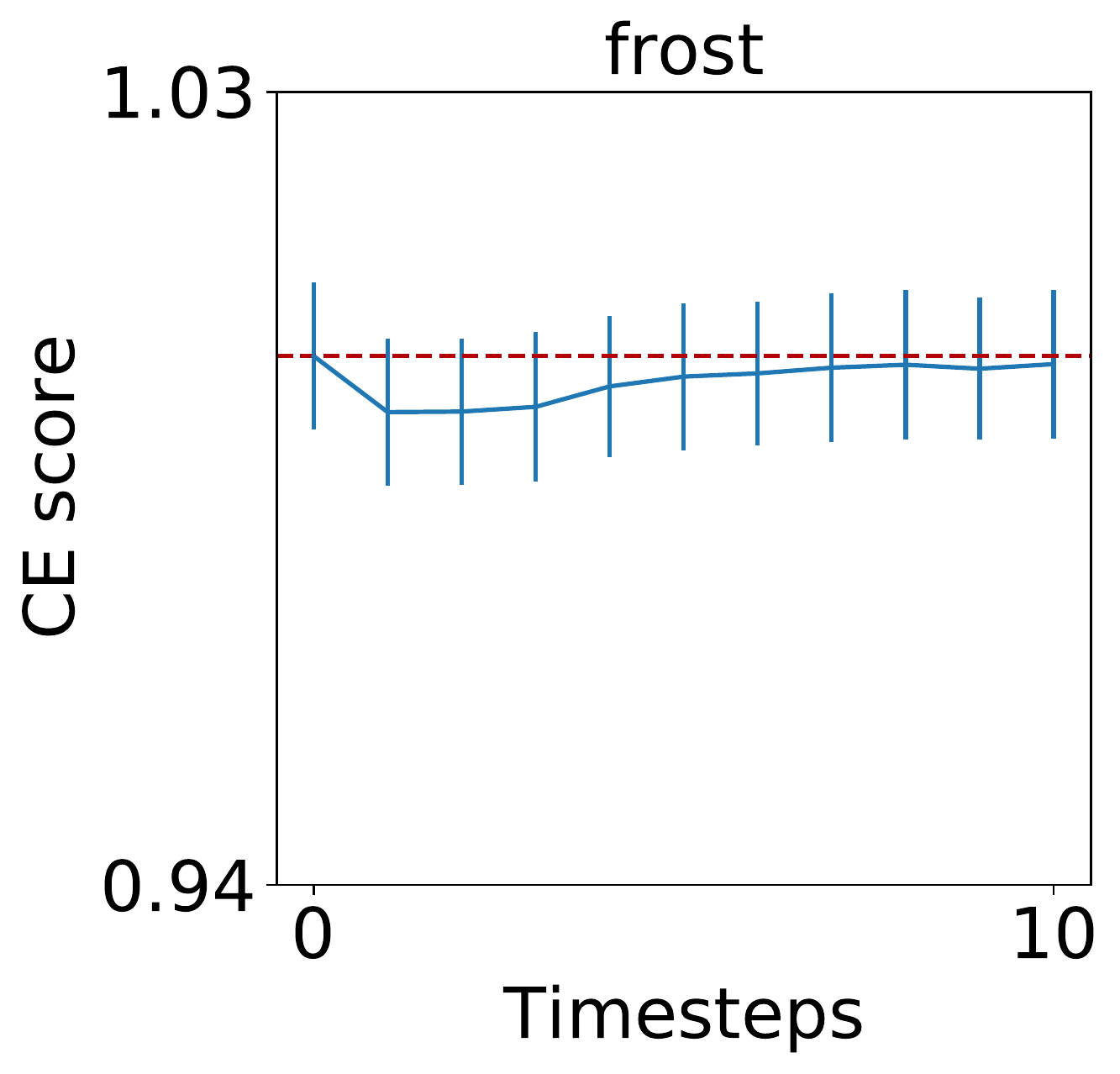}
    \end{subfigure}%
        \begin{subfigure}{0.24\textwidth}
        \includegraphics[scale=0.25]{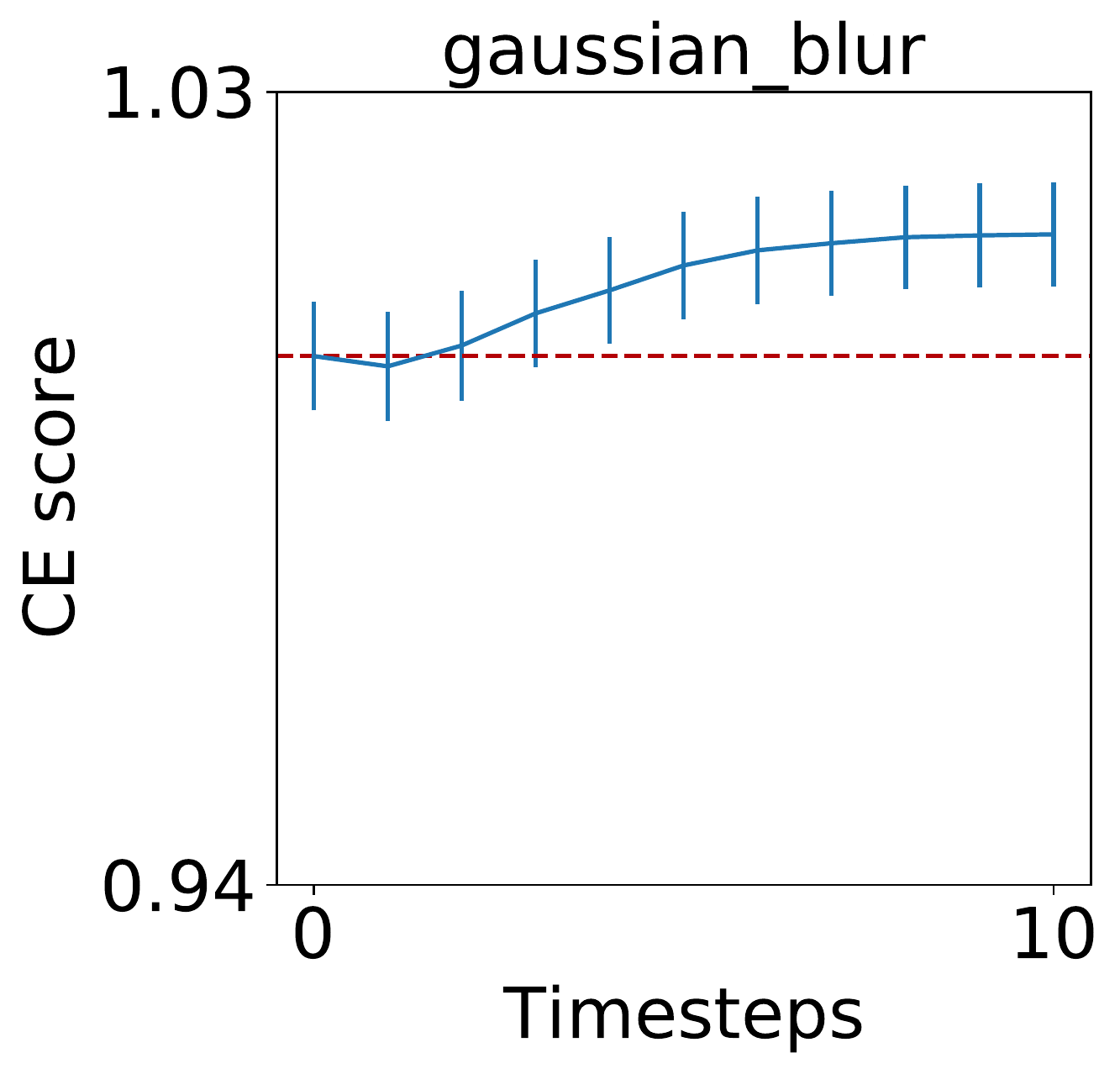}%
    \end{subfigure}%
        \begin{subfigure}{0.24\textwidth}
        \includegraphics[scale=0.25]{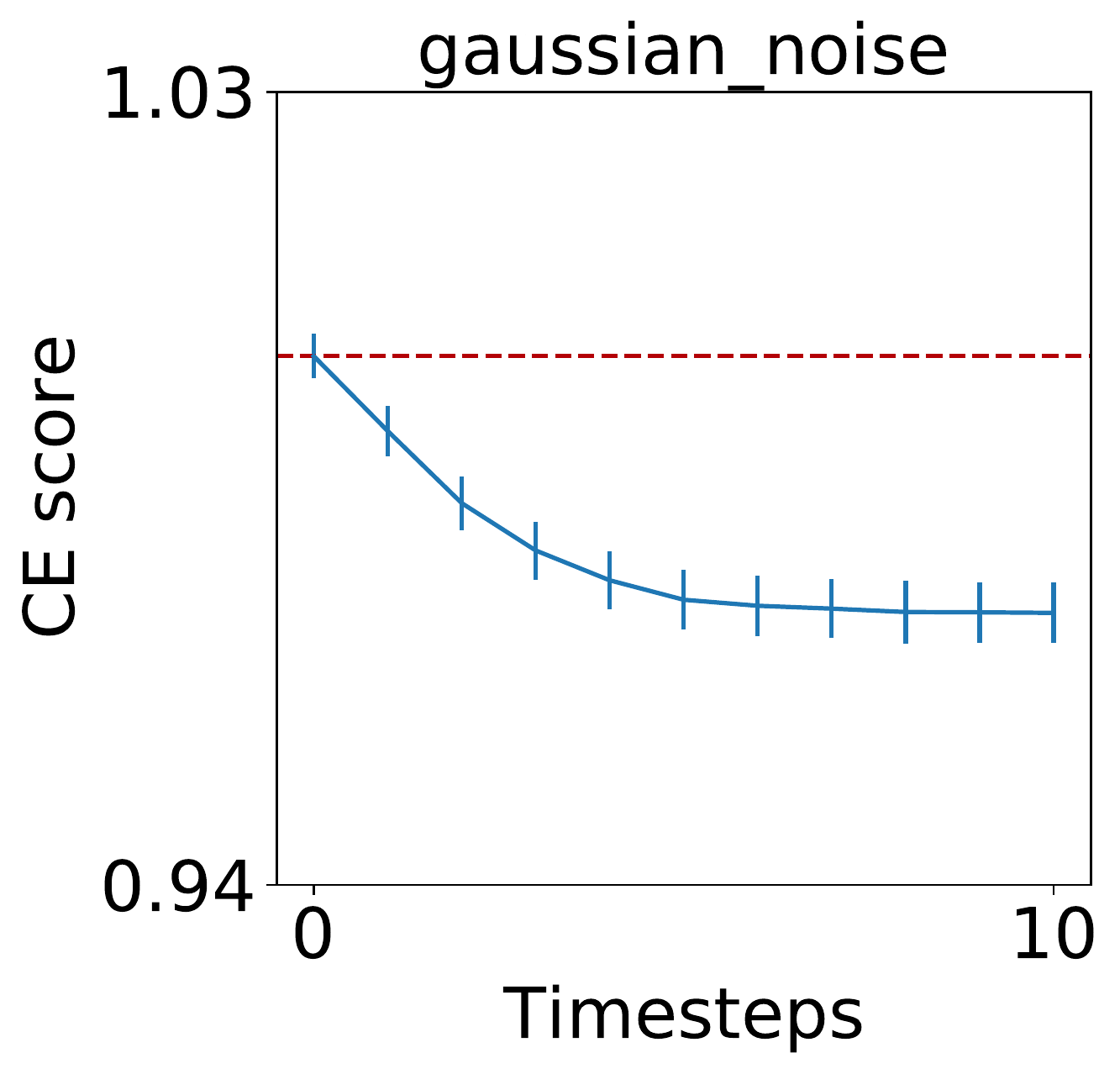}
    \end{subfigure}%
    
    \centering
    \begin{subfigure}{0.24\textwidth}
        \includegraphics[scale=0.25]{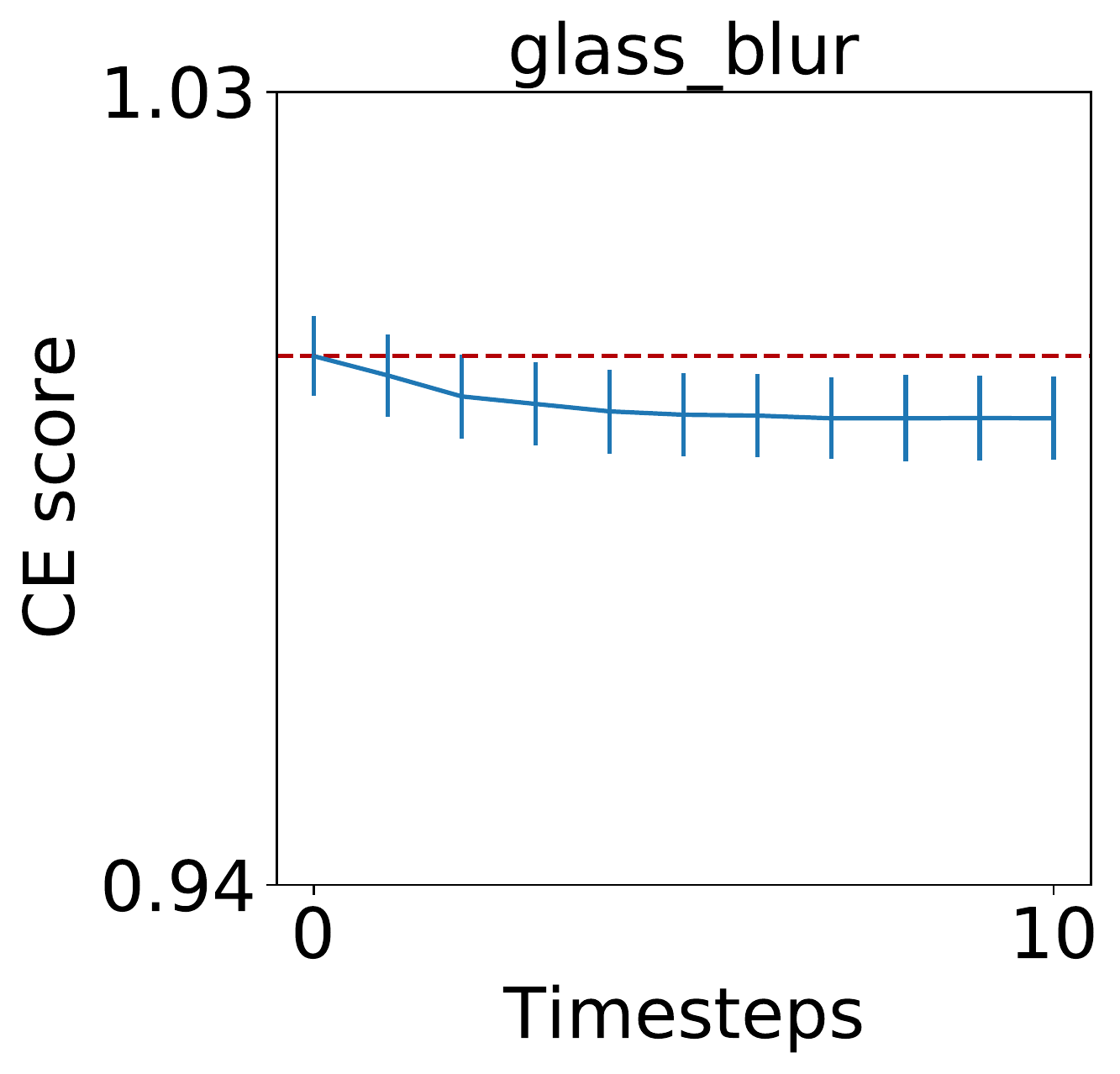}
    \end{subfigure}%
        \begin{subfigure}{0.24\textwidth}
        \includegraphics[scale=0.25]{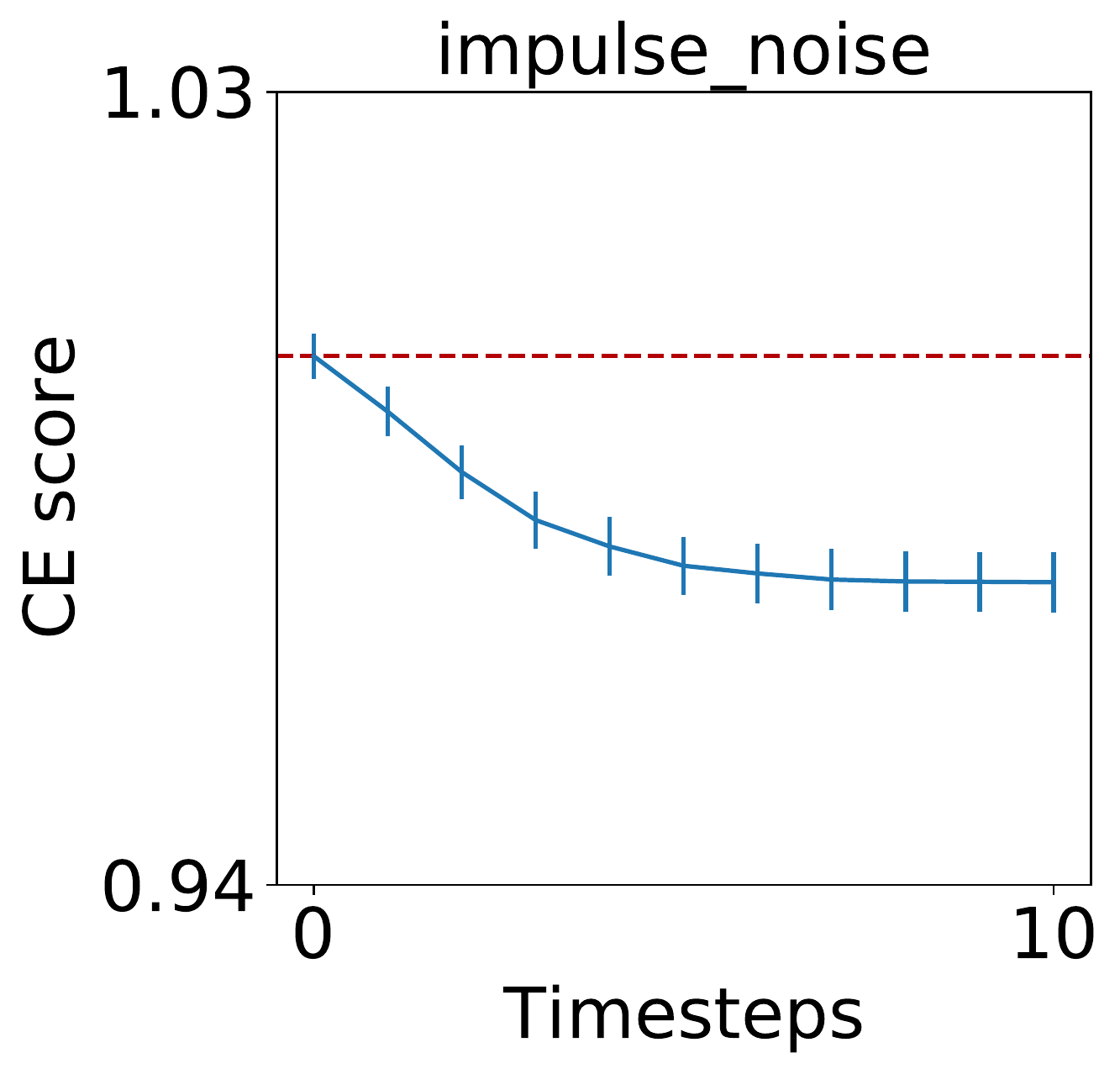}
    \end{subfigure}%
    \begin{subfigure}{0.24\textwidth}
        \includegraphics[scale=0.25]{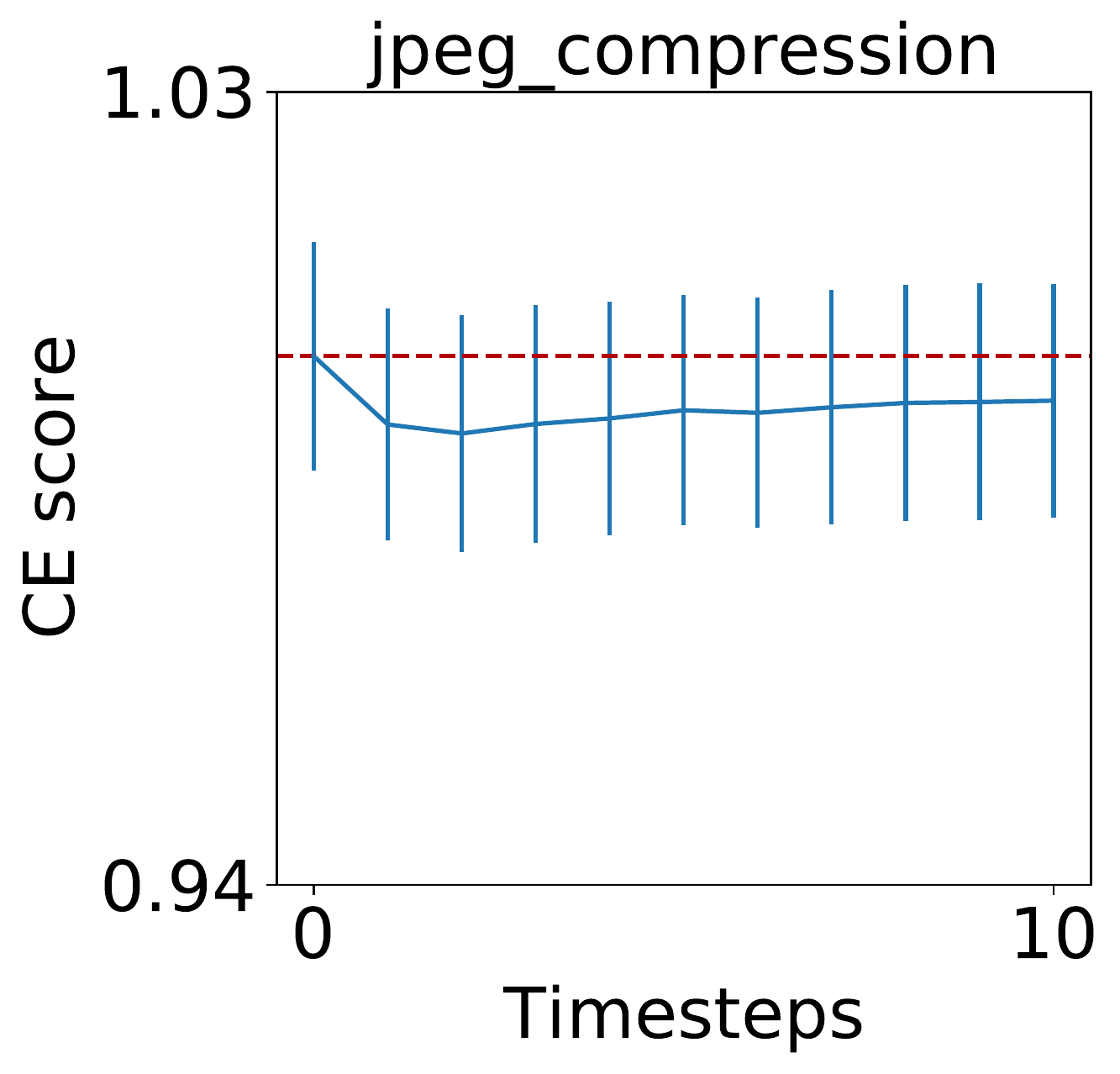}
    \end{subfigure}%
        \begin{subfigure}{0.24\textwidth}
        \includegraphics[scale=0.25]{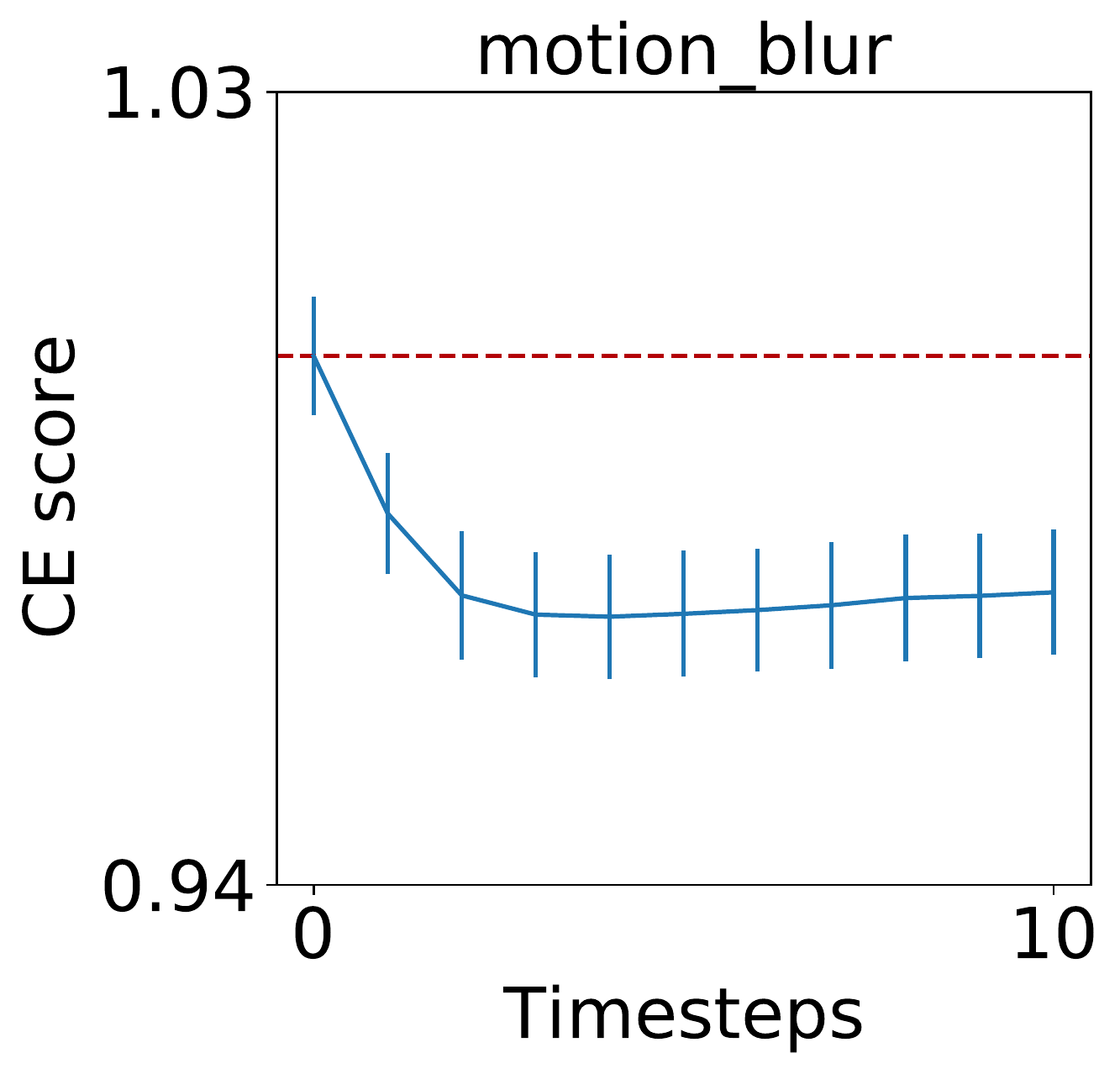}
    \end{subfigure}%

    \centering
    \begin{subfigure}{0.24\textwidth}
        \includegraphics[scale=0.25]{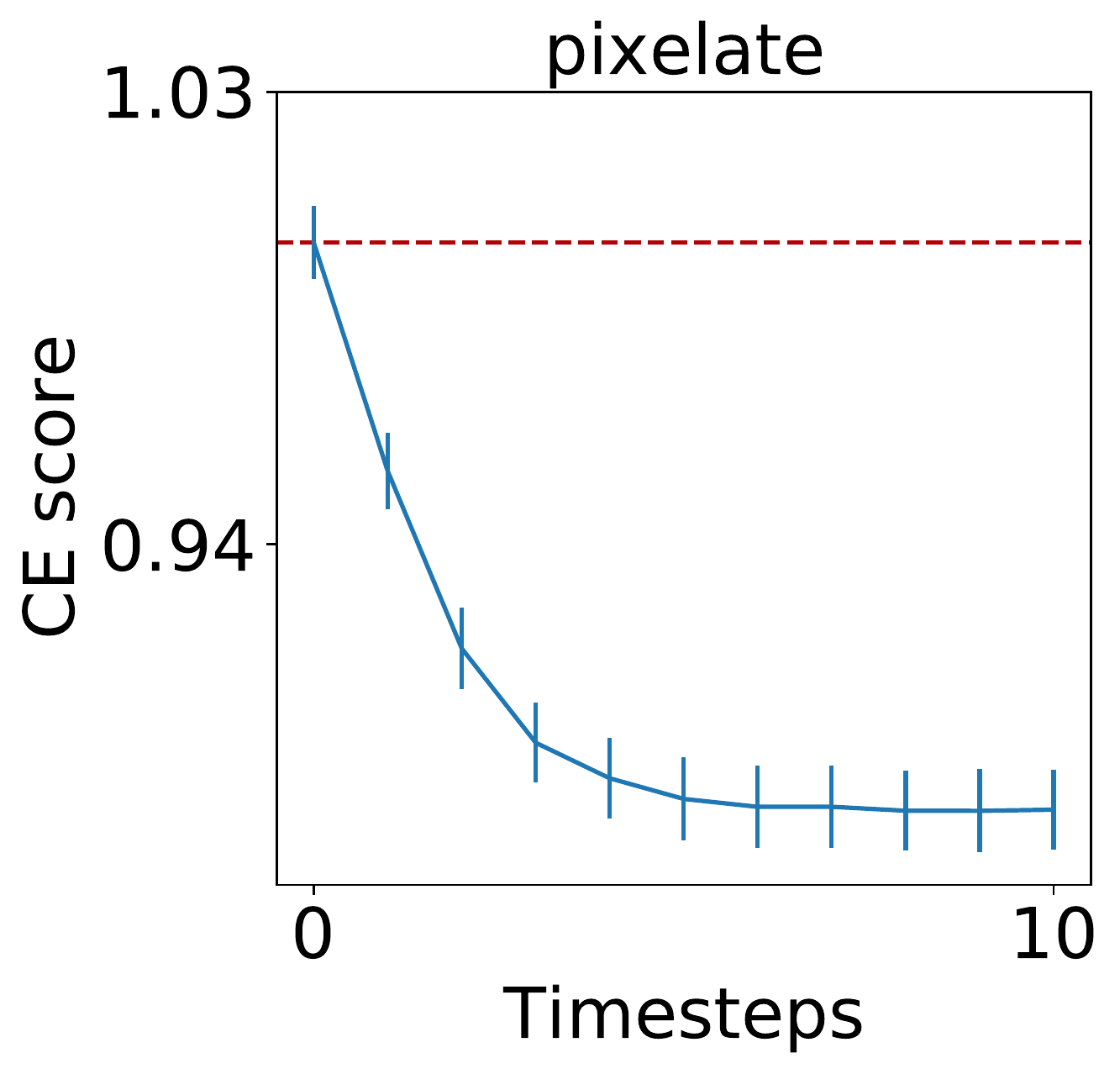}
    \end{subfigure}%
        \begin{subfigure}{0.24\textwidth}
        \includegraphics[scale=0.25]{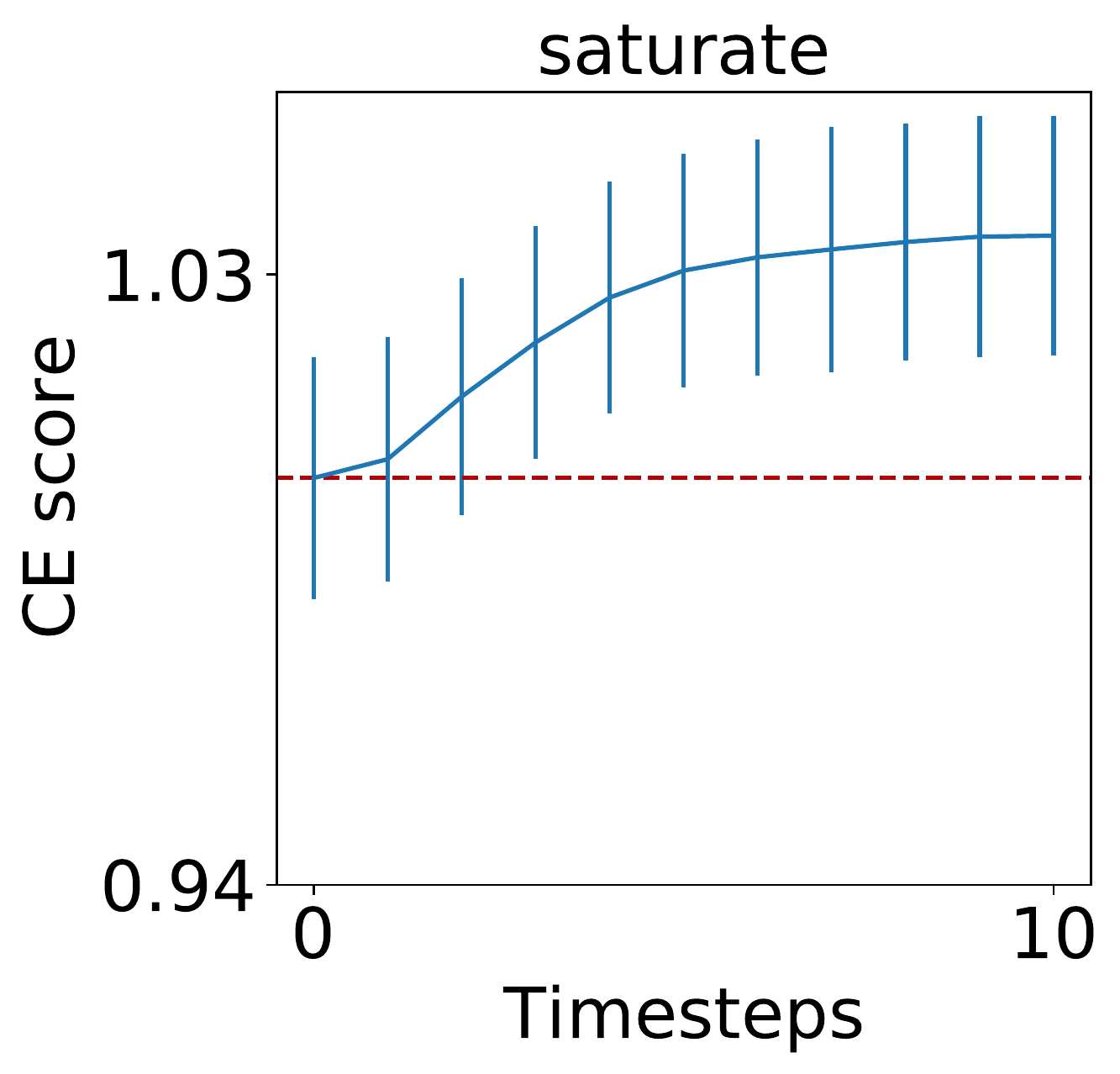}
    \end{subfigure}%
    \begin{subfigure}{0.24\textwidth}
        \includegraphics[scale=0.25]{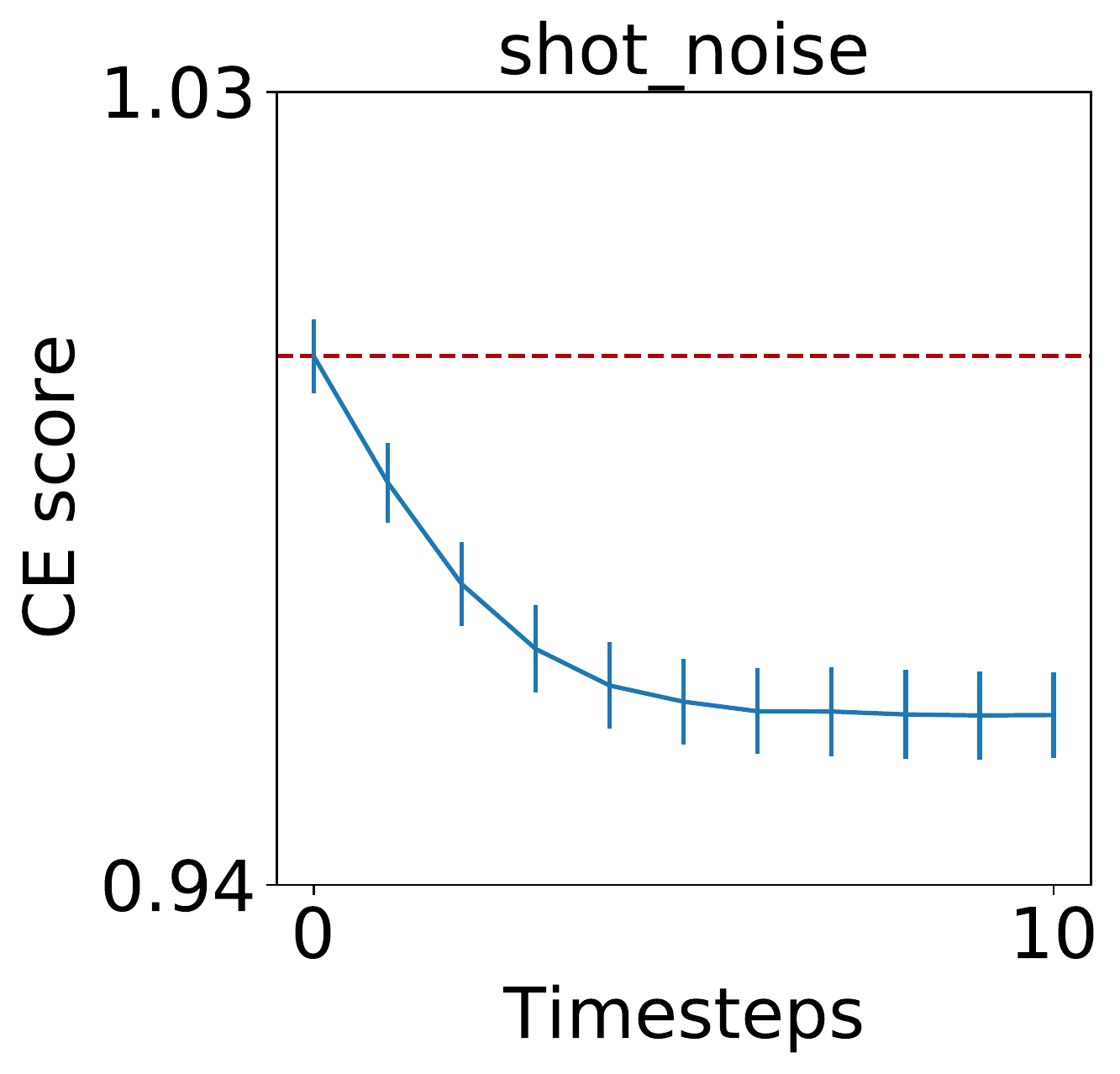}
    \end{subfigure}%
        \begin{subfigure}{0.24\textwidth}
        \includegraphics[scale=0.25]{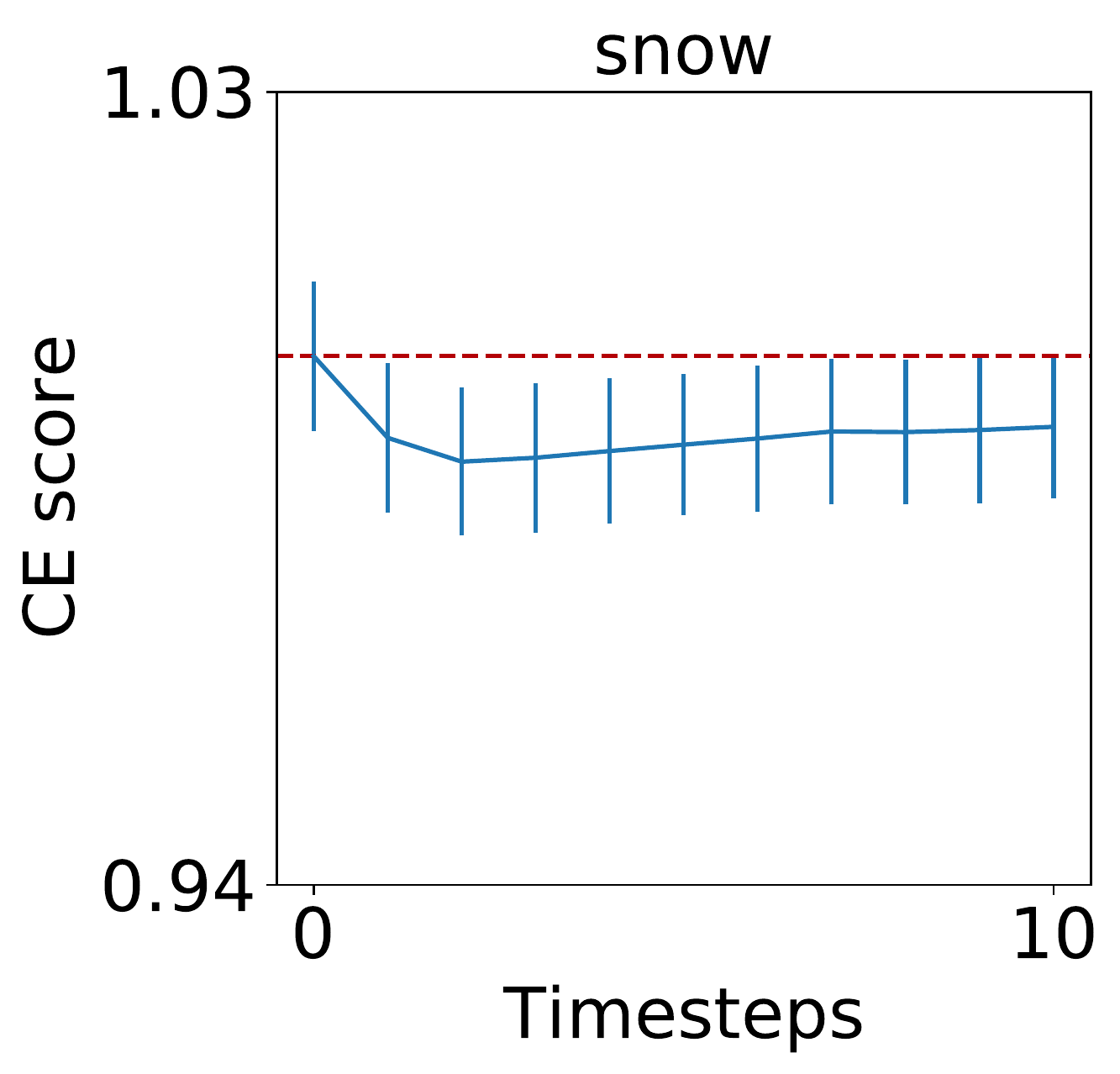}
    \end{subfigure}%
    
    \flushleft
    \hspace{0.19 cm}
    \begin{subfigure}{0.24\textwidth}
        \includegraphics[scale=0.25]{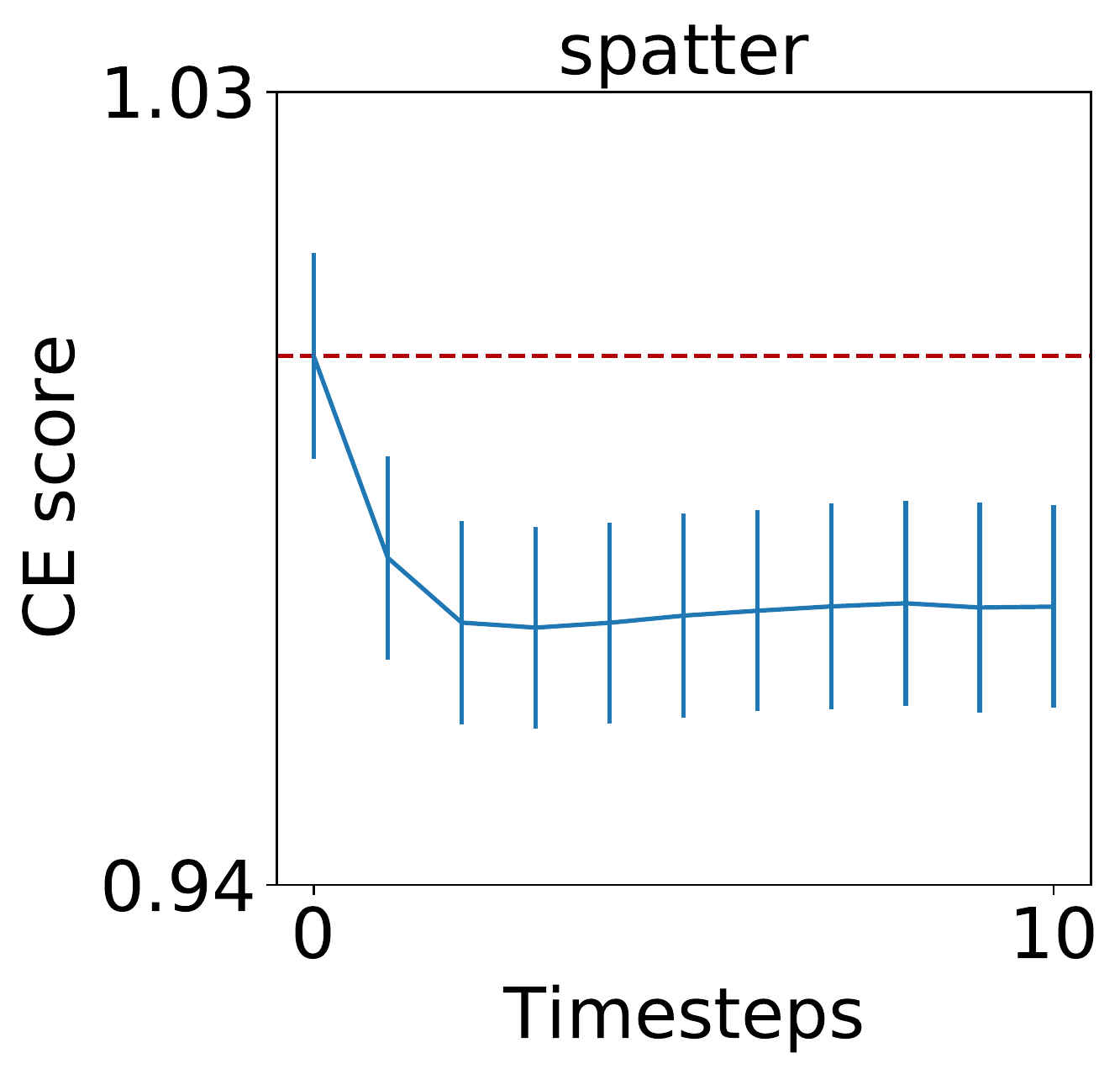}
    \end{subfigure}%
        \begin{subfigure}{0.24\textwidth}
        \includegraphics[scale=0.25]{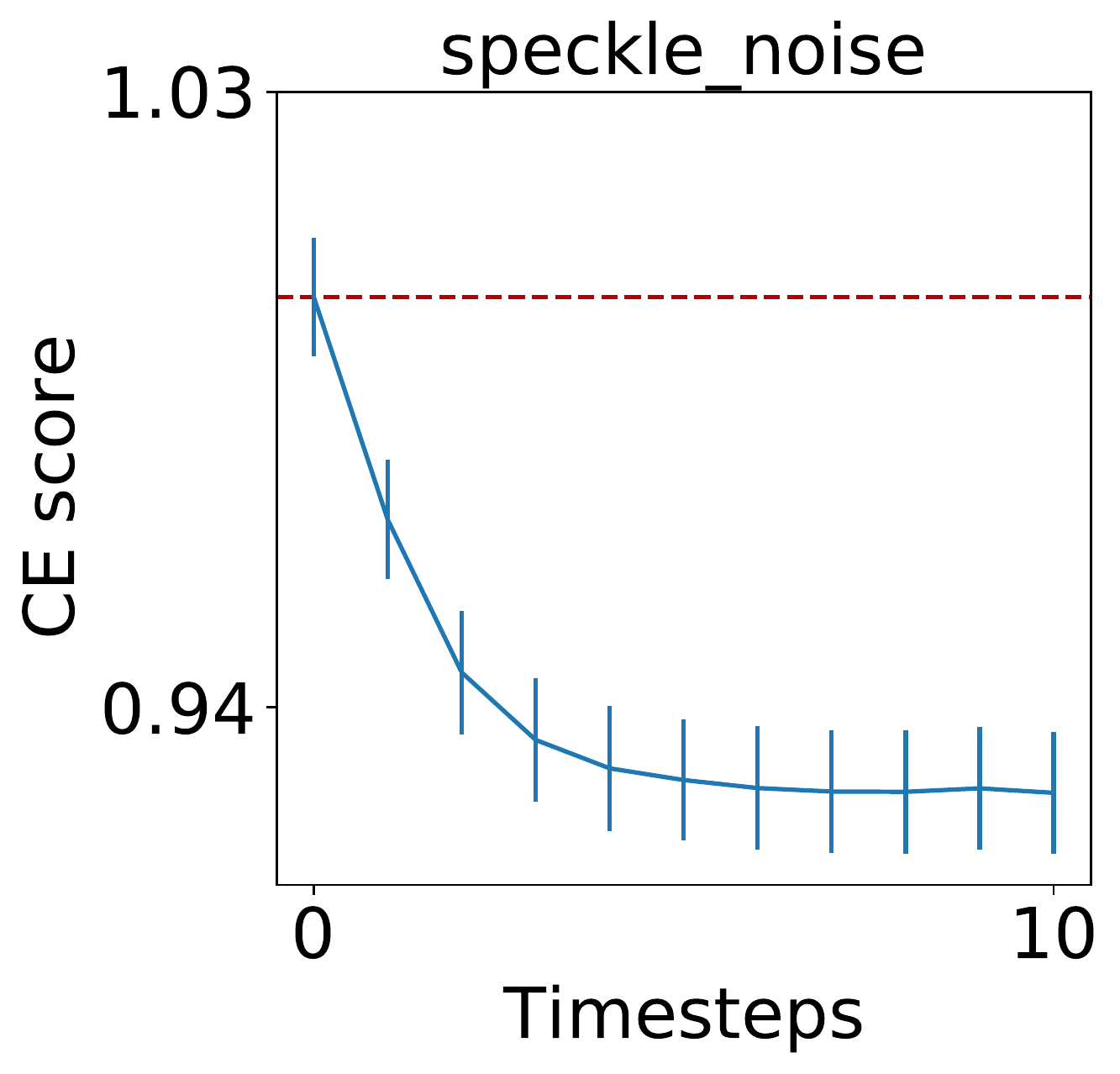}
    \end{subfigure}%
    \begin{subfigure}{0.24\textwidth}
        \includegraphics[scale=0.25]{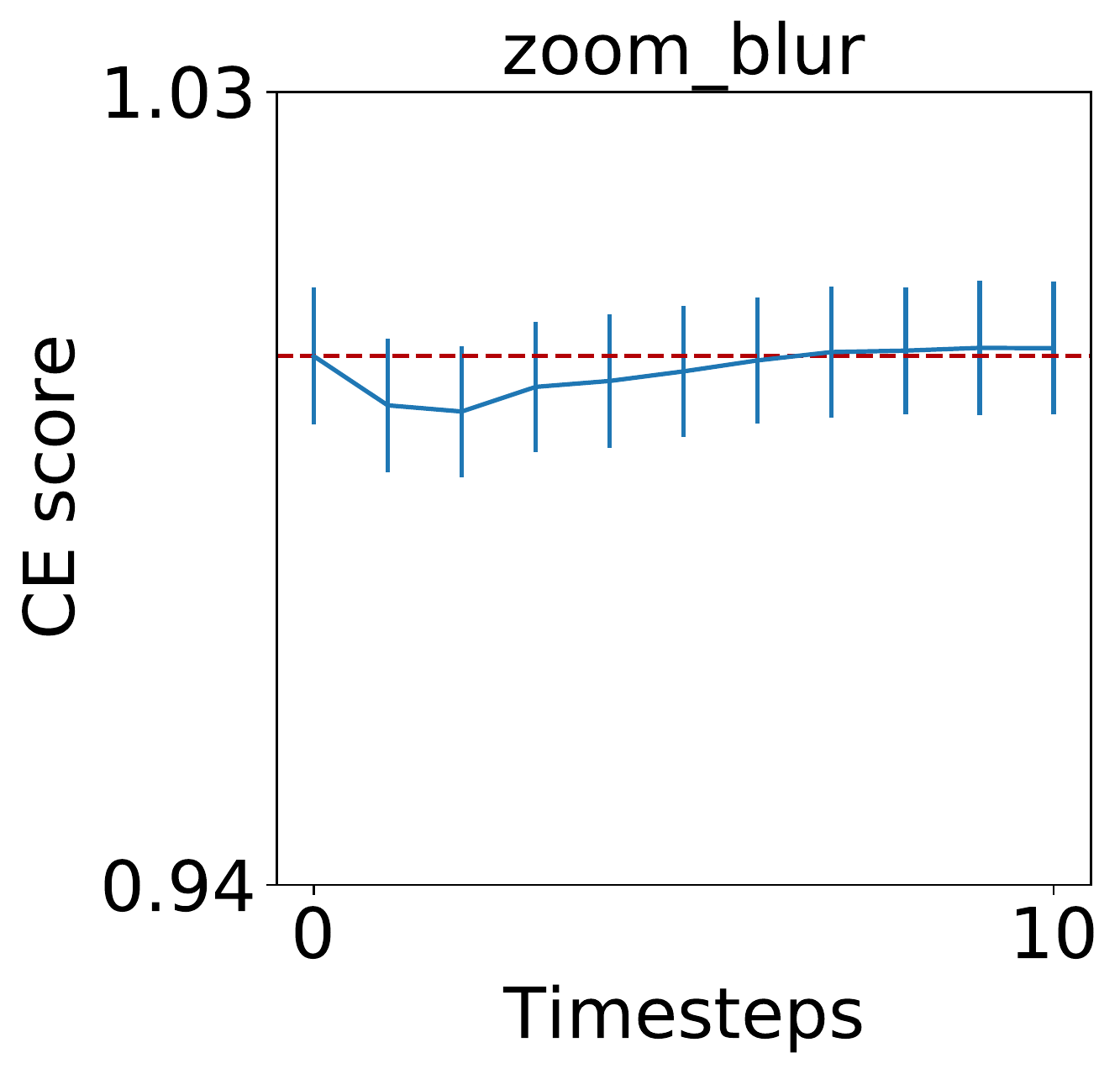}
    \end{subfigure}%

\caption{\textbf{PEfficientNetB0 (optimised) Corruption Error (CE) scores for all distortions:} The panel shows the CE scores calculated on the distorted images provided in the ImageNet-C dataset. The values are normalized with the CE score obtained for the feedforward EfficientNetB0. The error bars denote the standard deviation of the means obtained from bootstrapping (resampling multiple binary populations across all severities.) }    
\label{peffb0_ce_scores_appdx}
\end{figure}

\clearpage
\subsection{mCE scores of the optimized networks using AlexNet as a baseline }
\label{apndx:mce_alexnet}

\begin{figure}[h]
    \centering
    \includegraphics[width=0.3\linewidth]{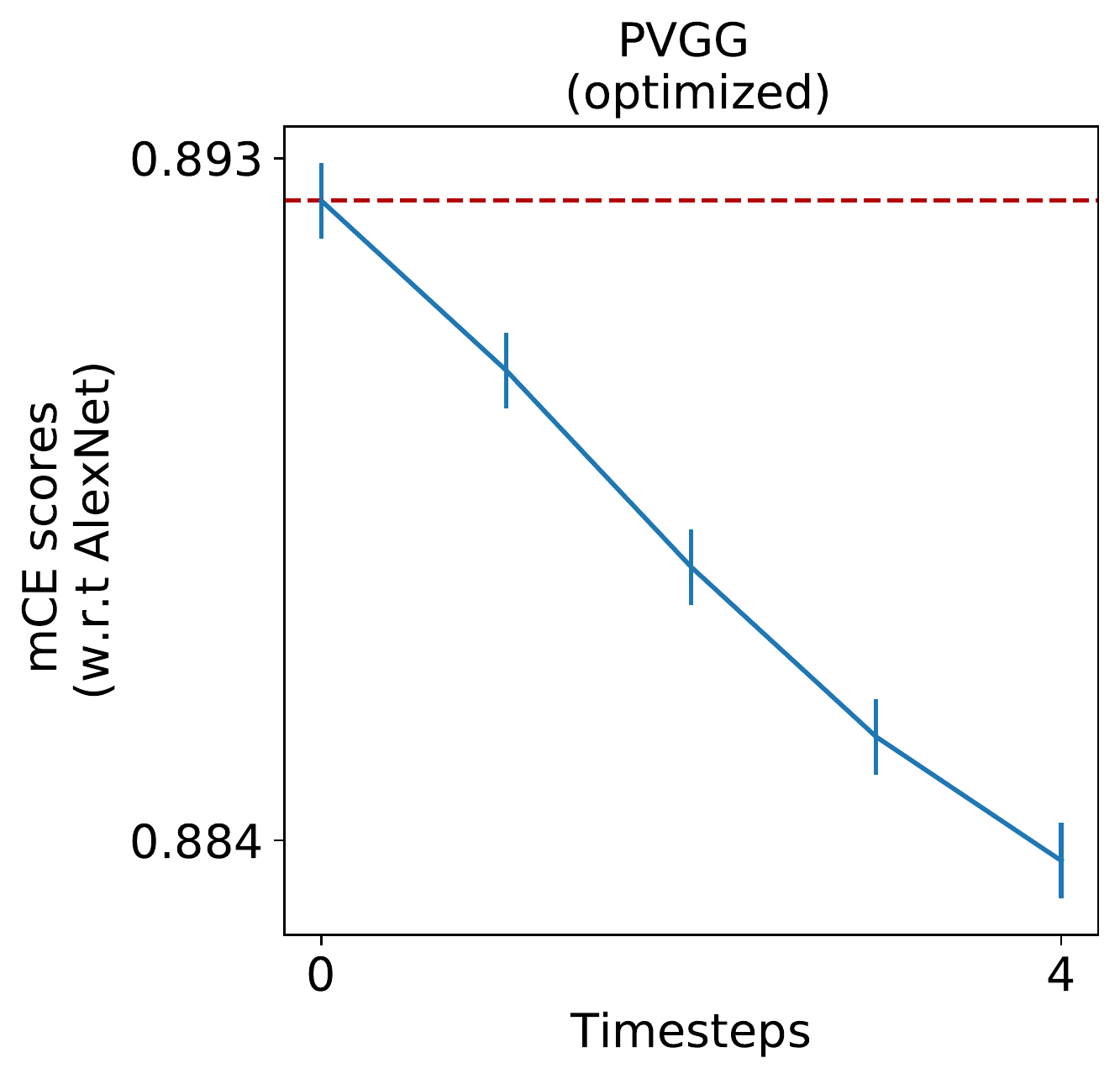}
    \includegraphics[width=0.3\linewidth]{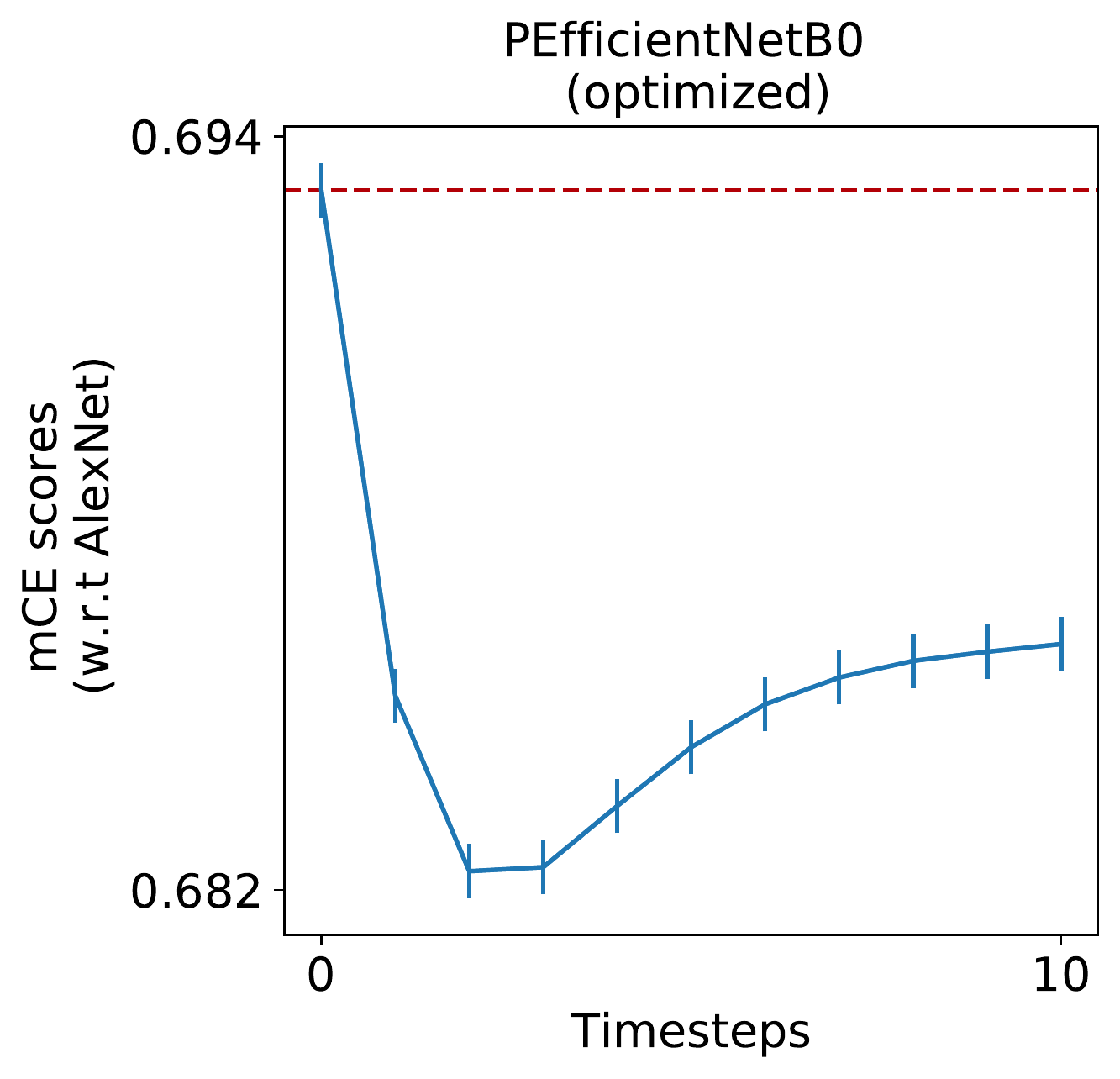}
    \caption{The mCE scores of the optimized networks (as shown in Figure 3) normalized using the score of the AlexNet network. Instead of normalizing using the score for the feedforward version of our recurrent network, to facilitate comparison with other works, we here normalize the scores using the score obtained for AlexNet network.}
    \label{fig:alexnet_as_baseline_plots}
\end{figure}

\begin{figure}[h]
    \centering
    \includegraphics[width=0.3\linewidth]{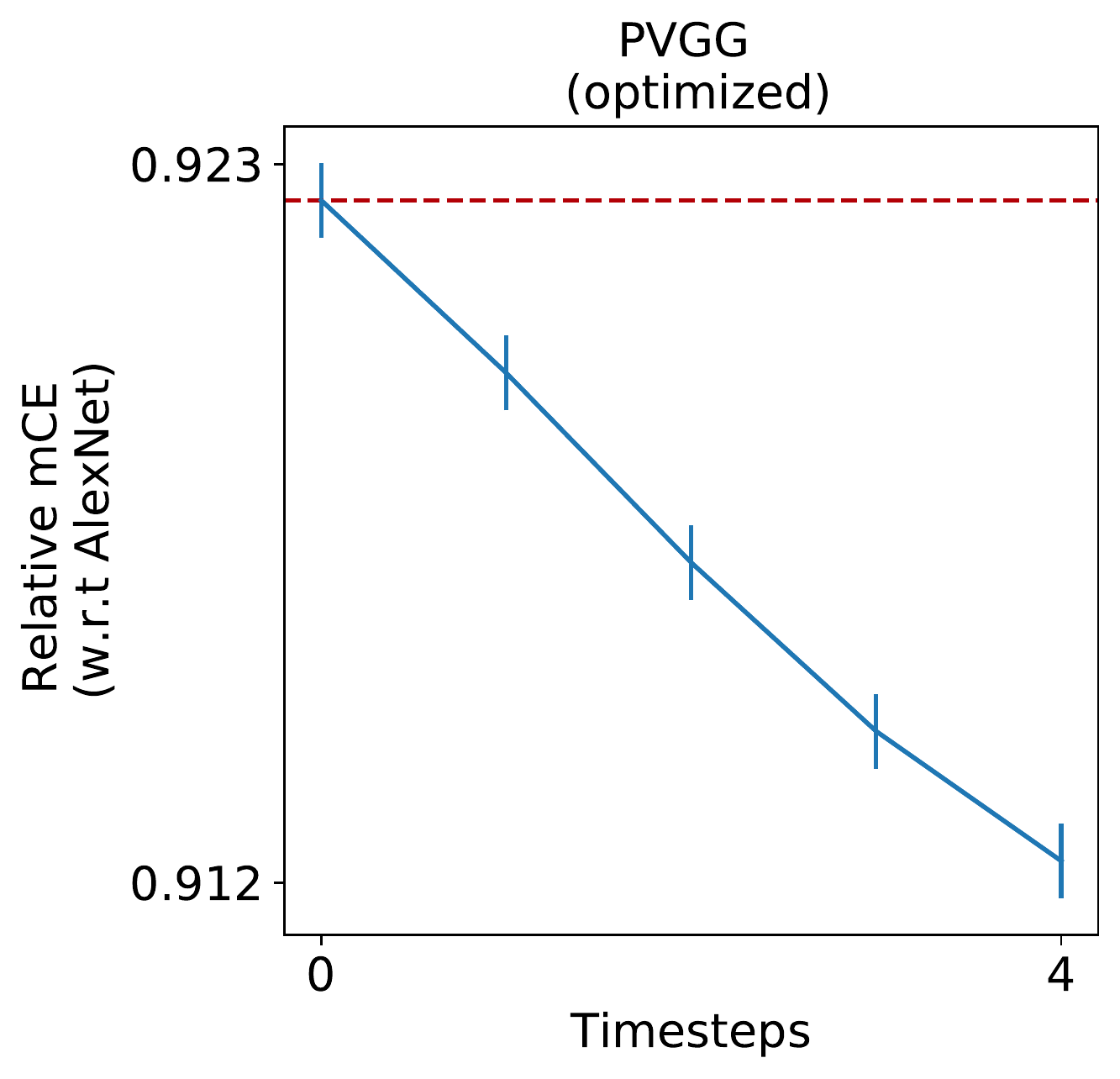}
    \includegraphics[width=0.3\linewidth]{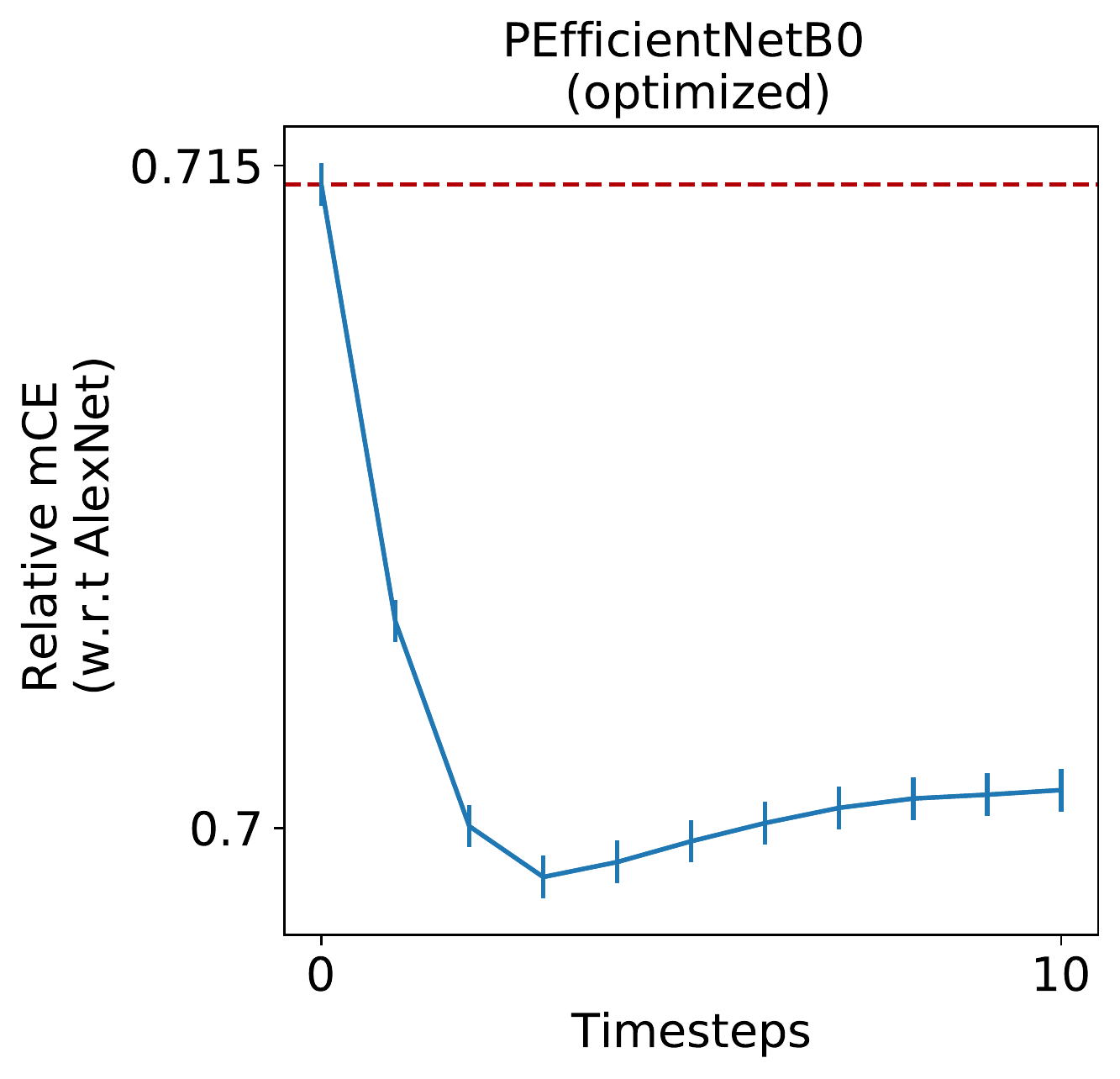}
    \caption{The Relative mCE scores of the optimized networks (as shown in Figure 3) normalized using the score of the AlexNet network. As suggested by~\citep{hendrycks2019benchmarking}, we use Relative mCE score which accounts for the changing baseline accuracy on the clean images over timesteps.} 
    \label{fig:relscores_alexnet_as_baseline_plots}
\end{figure}

\subsection{mCE scores of a predified robust network}
\label{apndx:mce_robust_net}

We also incorporated our recurrent dynamics in an already robust PEfficientNet network. As a simple approach, we just used the hyperparameters ($\alpha$, $\beta$ and $\lambda$) that were optimized for the non-robust version of PEfficientNEtB0 (on 0.25 gaussian noise) and measured its robustness against the corruptions in ImageNet-C dataset.
We observed that the proposed predictive coding dynamics further helped in improving the robustness of this already robust network.

\begin{figure}[h!]
    \centering
    \includegraphics[width=0.3\linewidth]{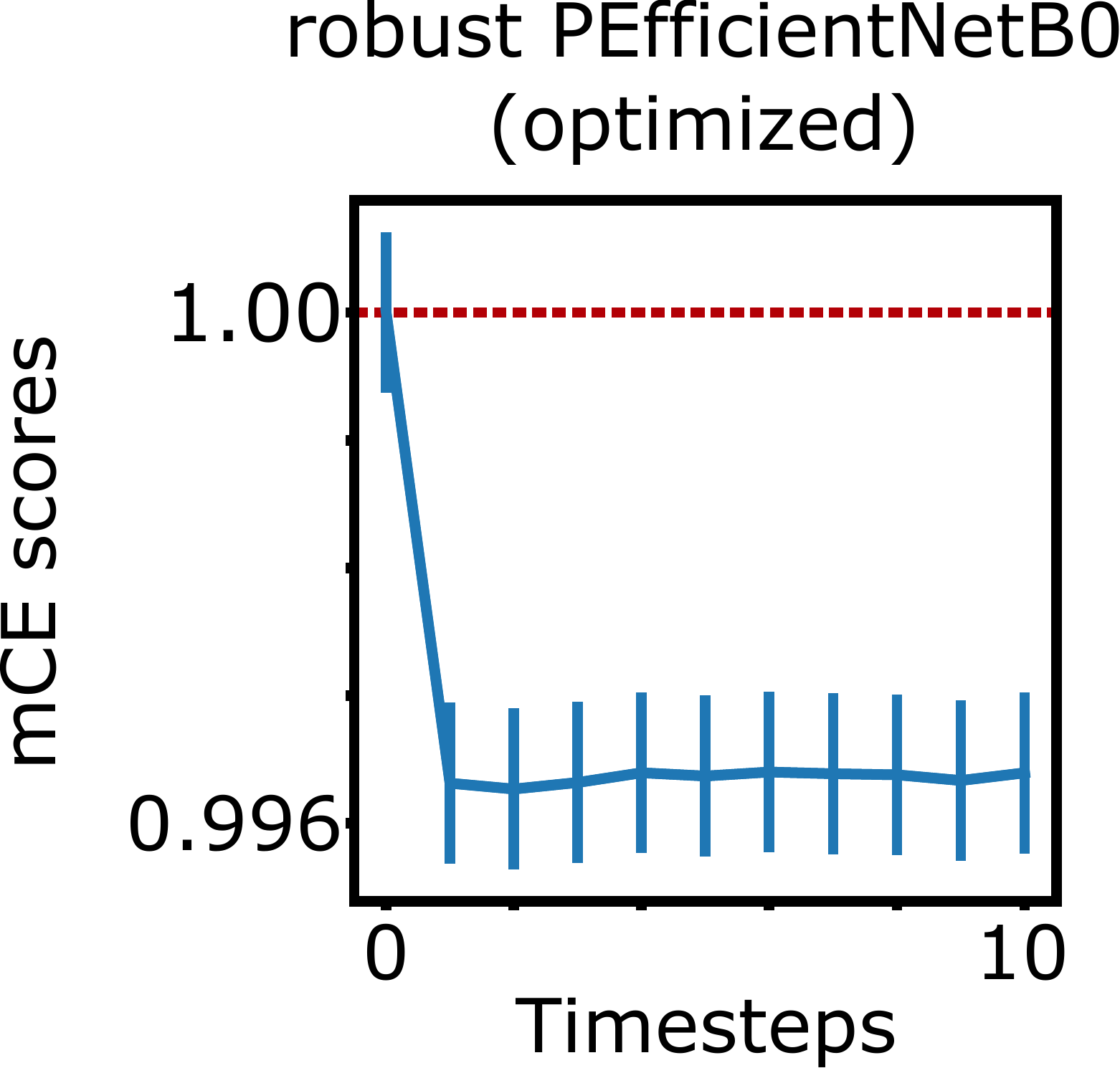}
    \caption{ mCE scores of a predified version of an already robust PEfficientNetB0}
    \label{fig:robust_peffb0_xie}
\end{figure}

\clearpage
\subsection{Original data for Adversarial Attacks}
\label{apndx:adv_attck_orig_data}

We provide here the non-baseline corrected versions of the data presented for adversarial attacks in Figure 4. The panels below show the success rate of the targeted attacks across timesteps calculated on 1000 images. The perturbations allowed ($\epsilon$) and the type of attack are denoted at the top.

\begin{figure}[h!]

    \centering
    \begin{subfigure}{0.23\textwidth}
    \includegraphics[scale=0.23]{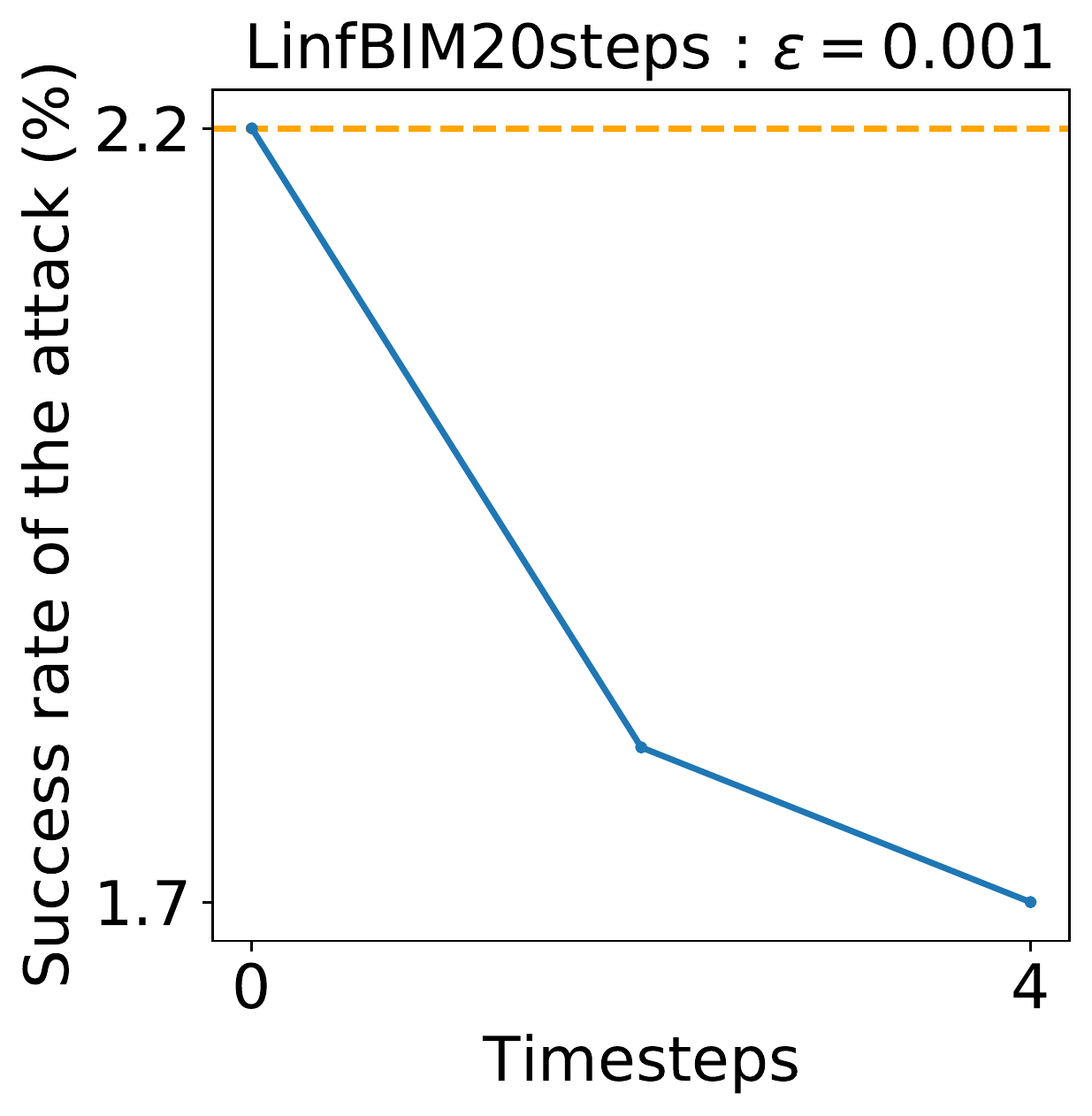}
    \end{subfigure}
    \begin{subfigure}{0.23\textwidth}
    \includegraphics[scale=0.23]{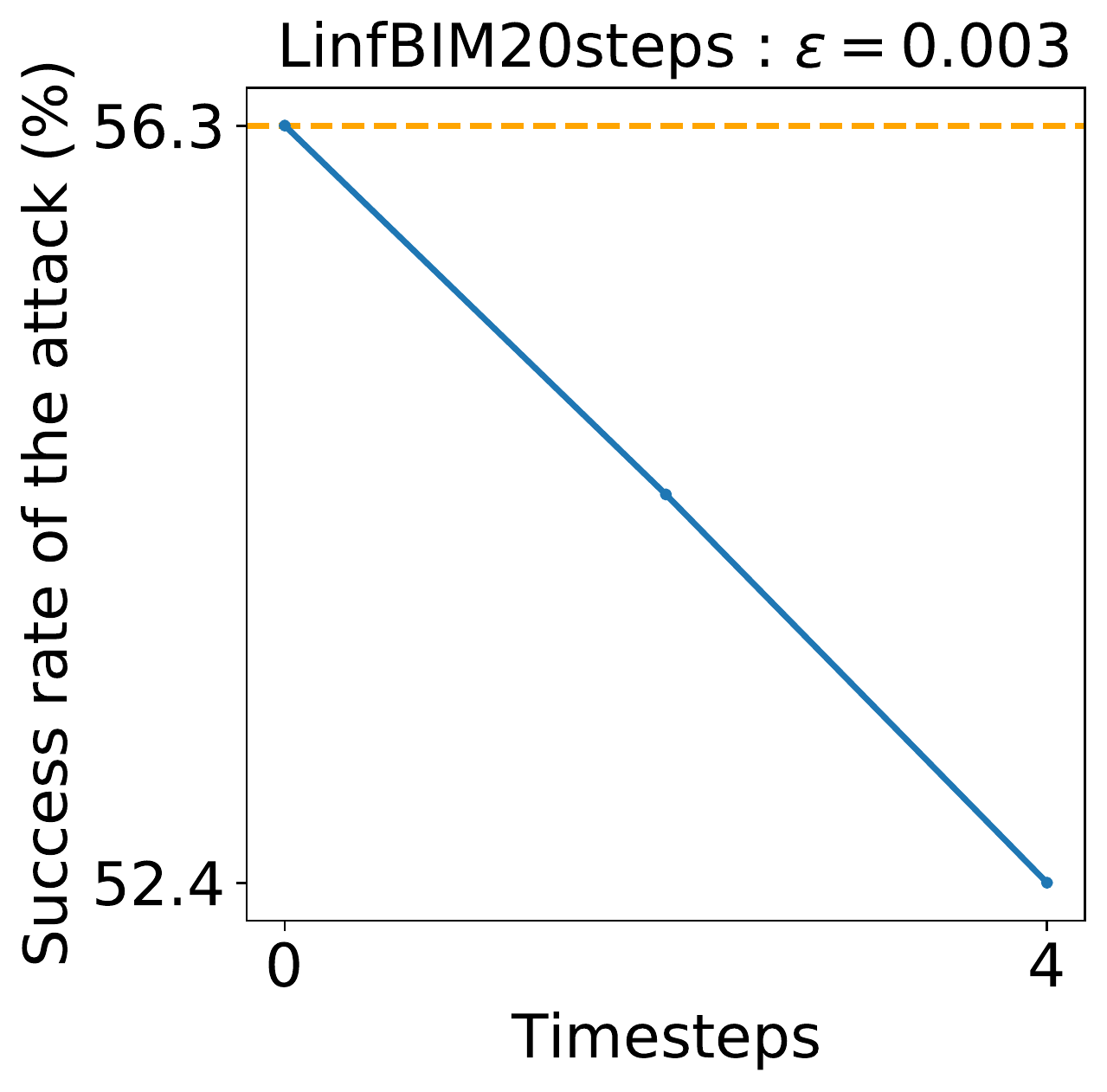}
    \end{subfigure}
    \begin{subfigure}{0.23\textwidth}
    \includegraphics[scale=0.23]{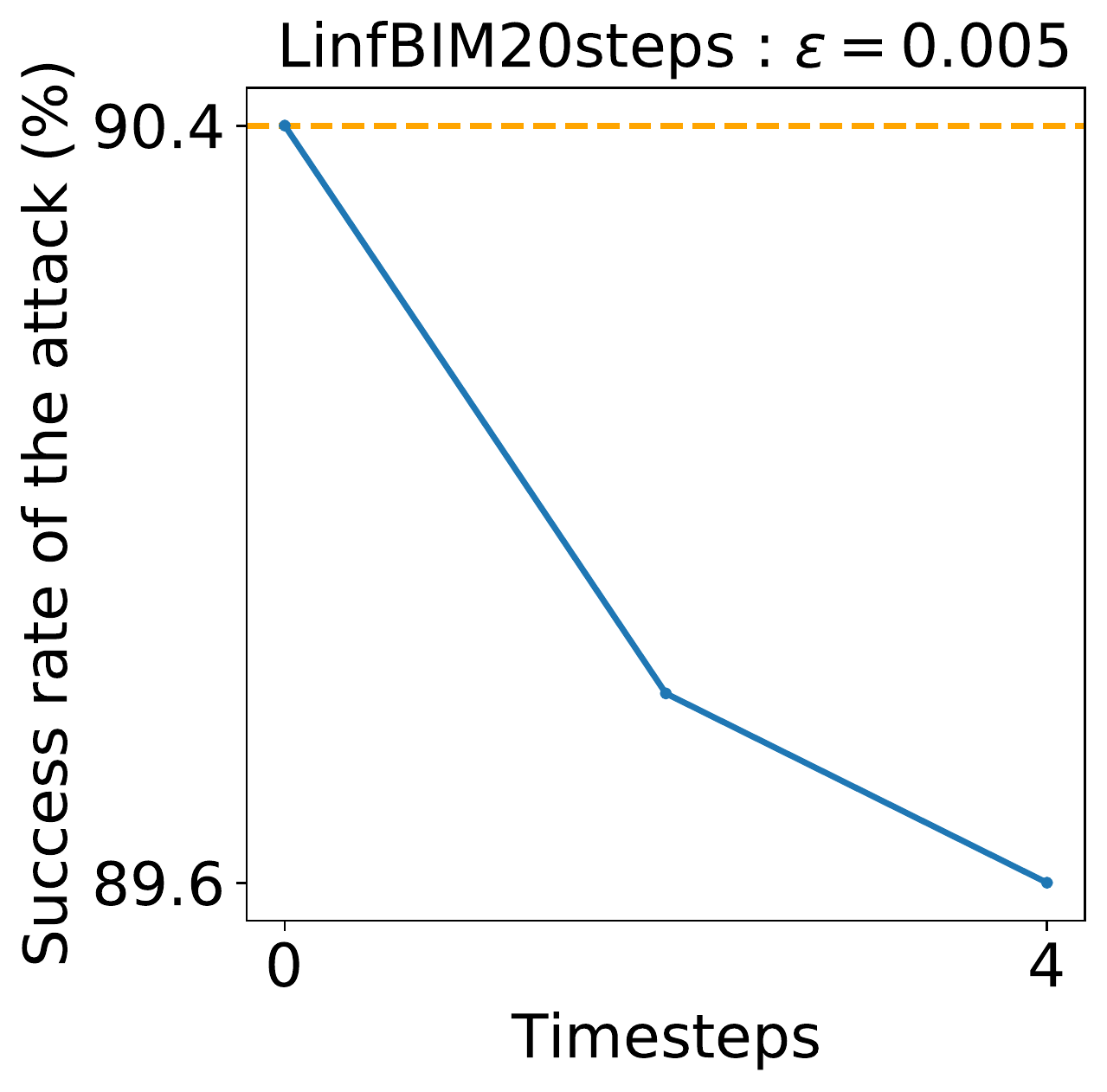}
    \end{subfigure}
    \caption{$L_\infty$BIM attacks on PVGG16 network }

    \begin{subfigure}{0.23\textwidth}
    \includegraphics[scale=0.23]{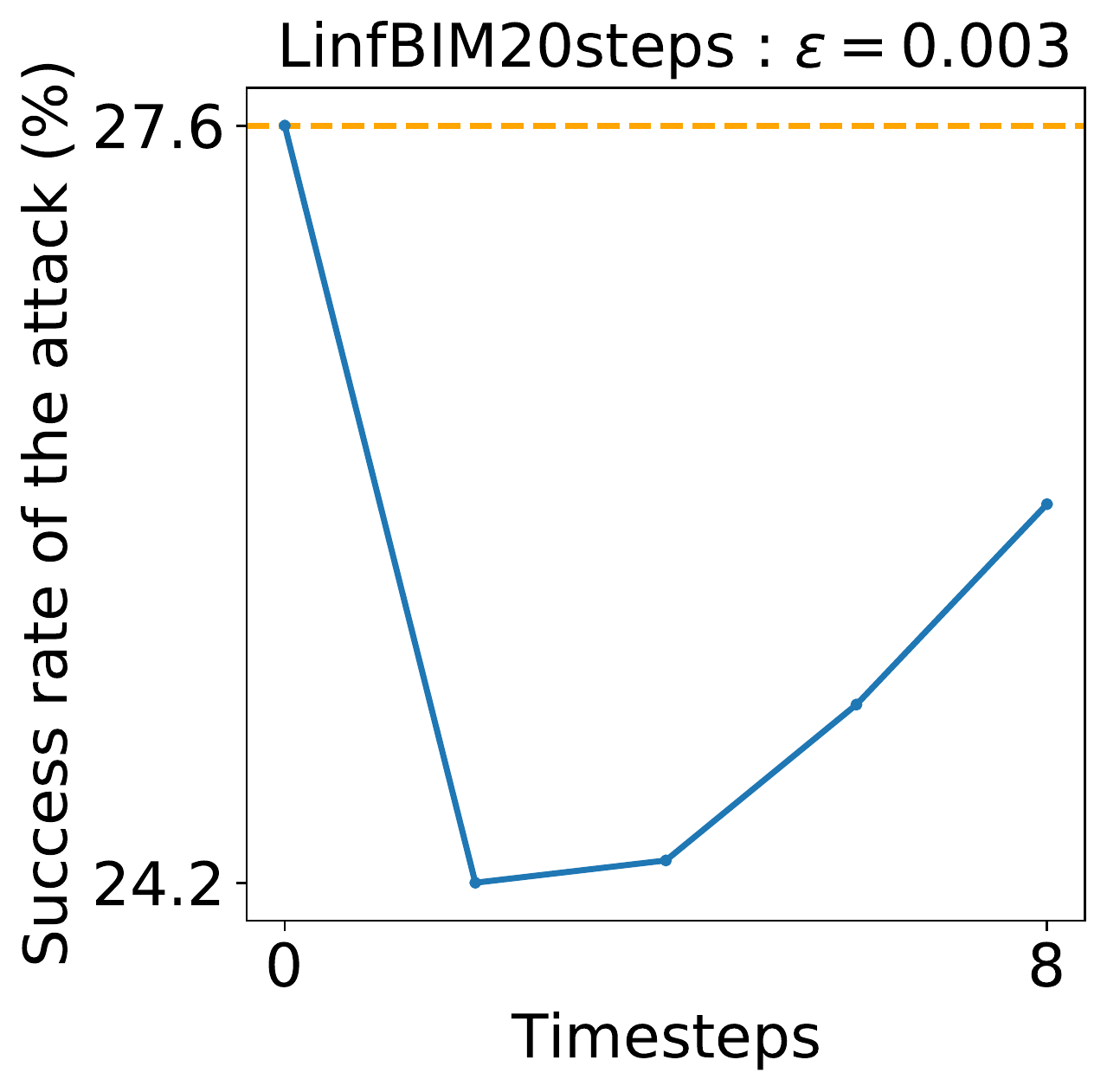}
    \end{subfigure}
    \begin{subfigure}{0.23\textwidth}
    \includegraphics[scale=0.23]{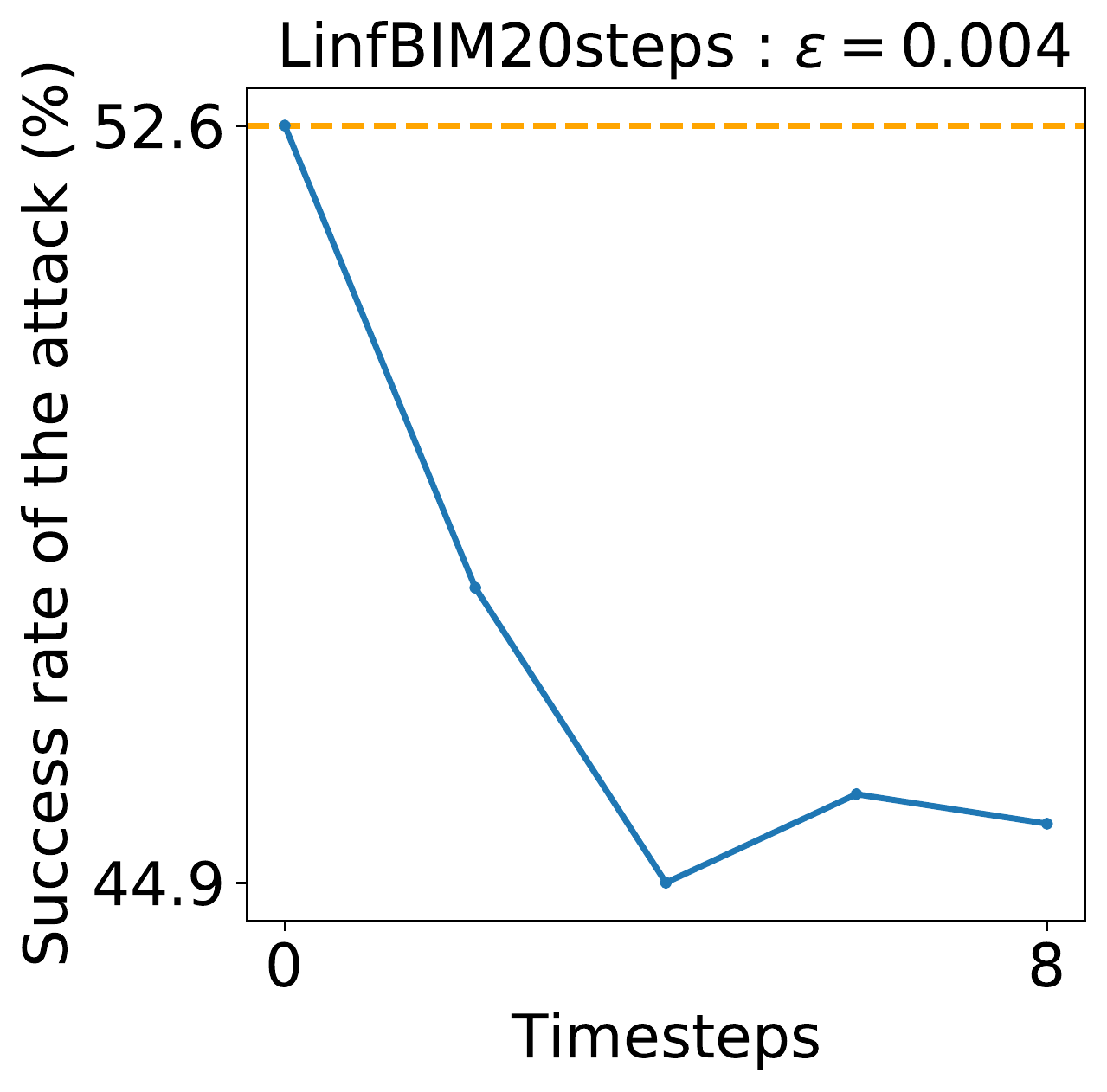}
    \end{subfigure}
    \begin{subfigure}{0.23\textwidth}
    \includegraphics[scale=0.23]{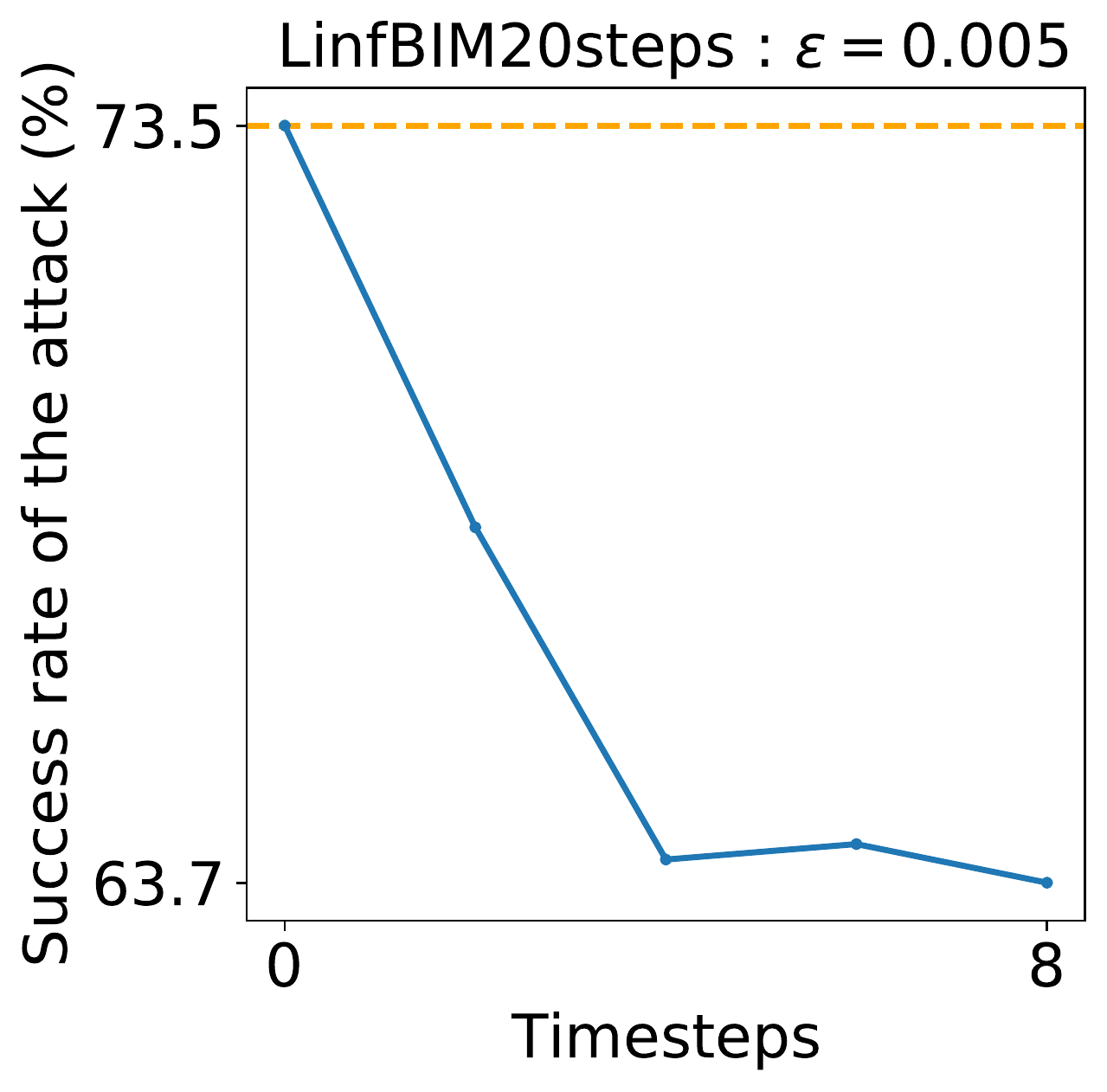}
    \end{subfigure}
    \begin{subfigure}{0.23\textwidth}
    \includegraphics[scale=0.23]{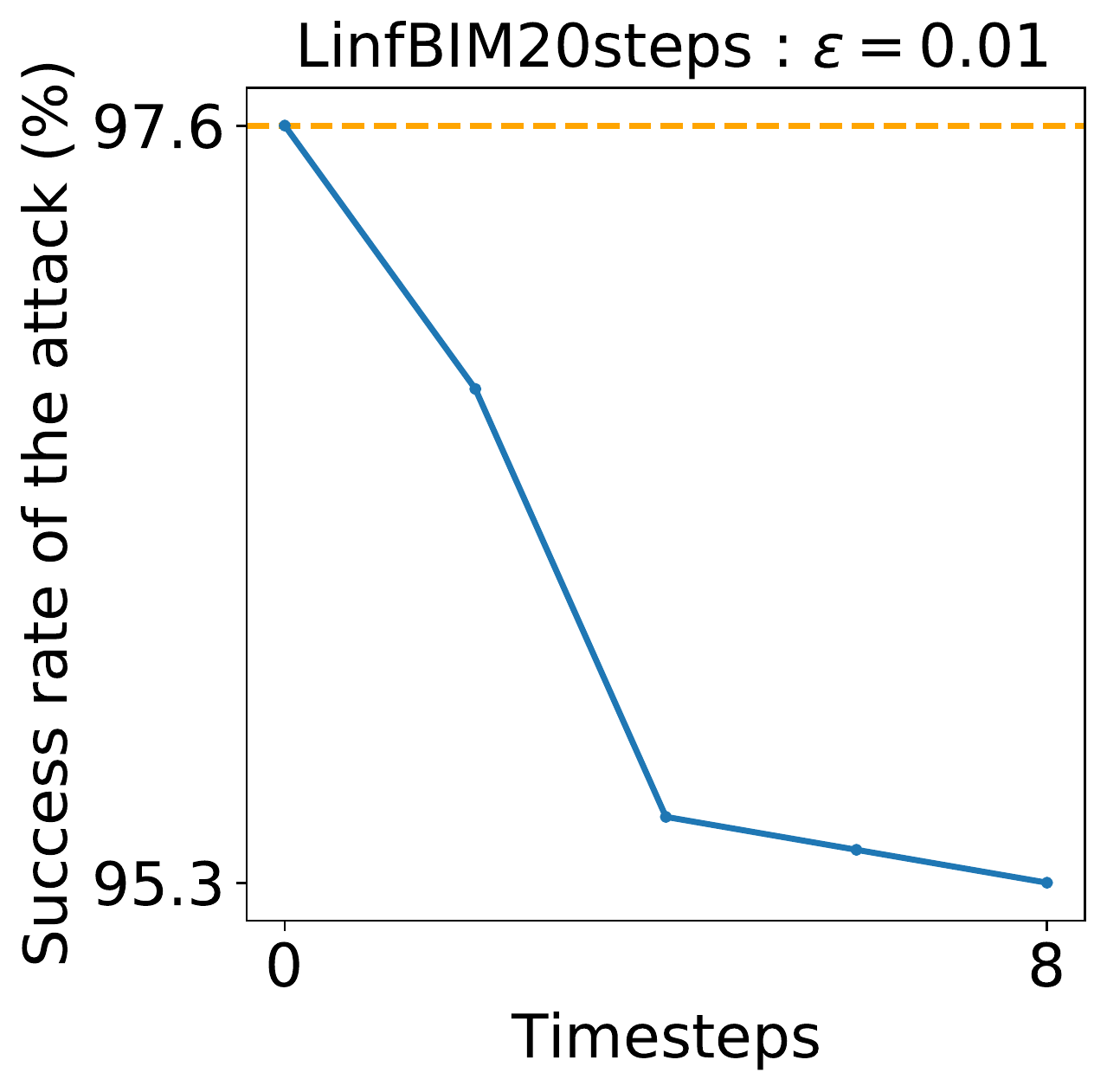}
    \end{subfigure}
    \caption{$L_\infty$BIM attacks on PEfficientNetB0 network}

\end{figure}

\begin{figure}[h!]

    \centering
    \begin{subfigure}{0.23\textwidth}
    \includegraphics[scale=0.23]{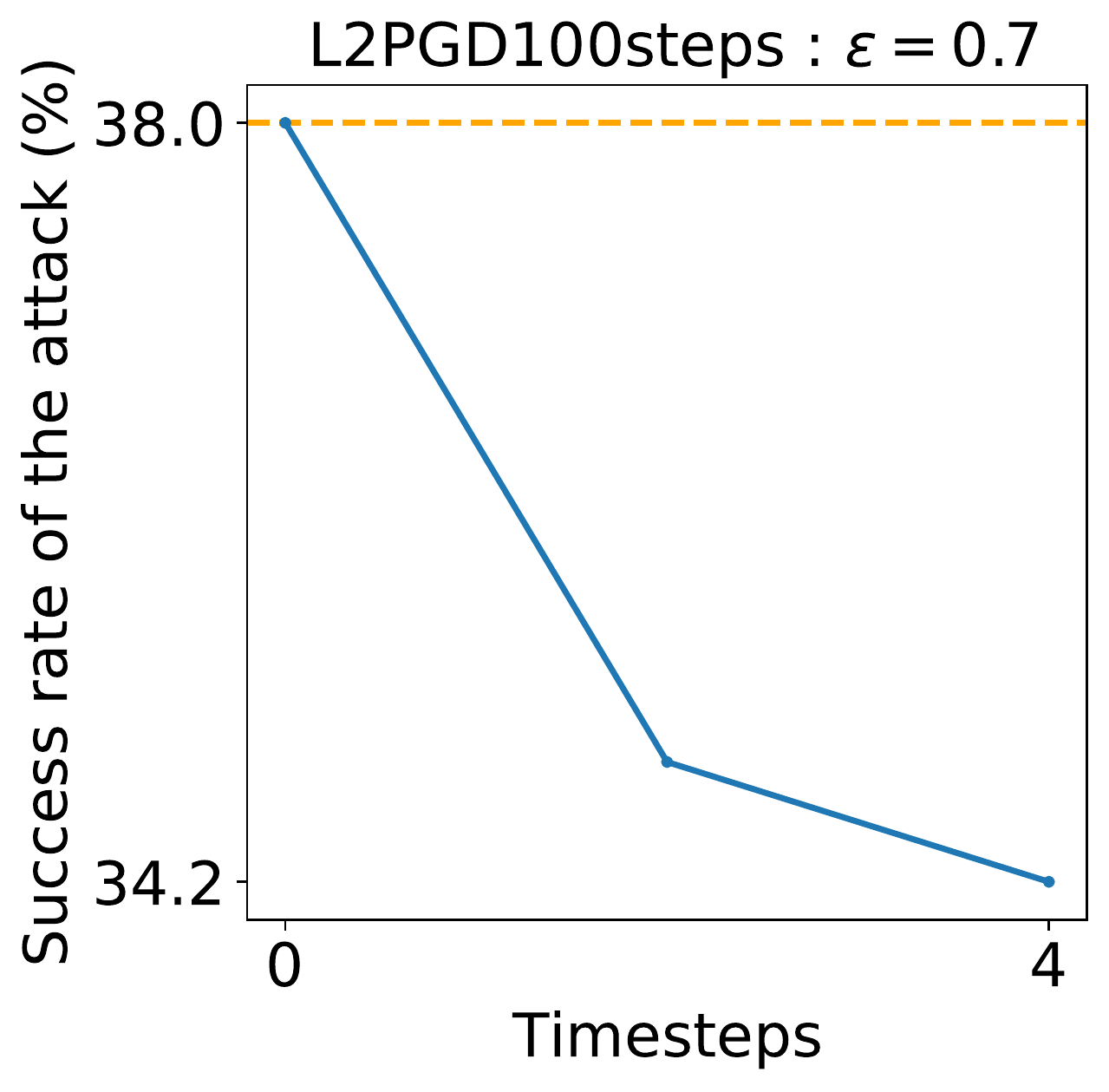}
    \end{subfigure}
    \begin{subfigure}{0.23\textwidth}
    \includegraphics[scale=0.23]{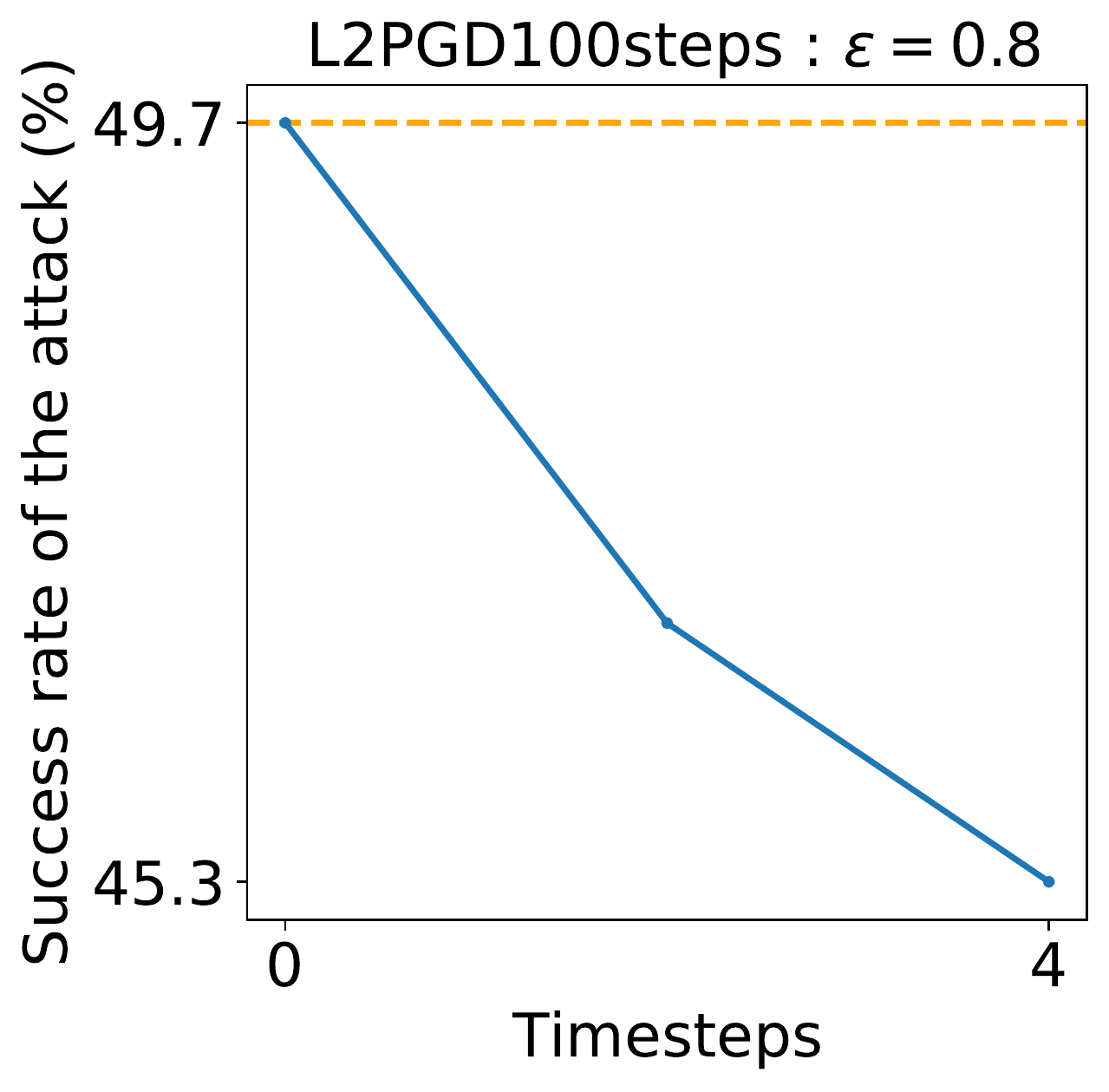}
    \end{subfigure}
    \begin{subfigure}{0.23\textwidth}
    \includegraphics[scale=0.23]{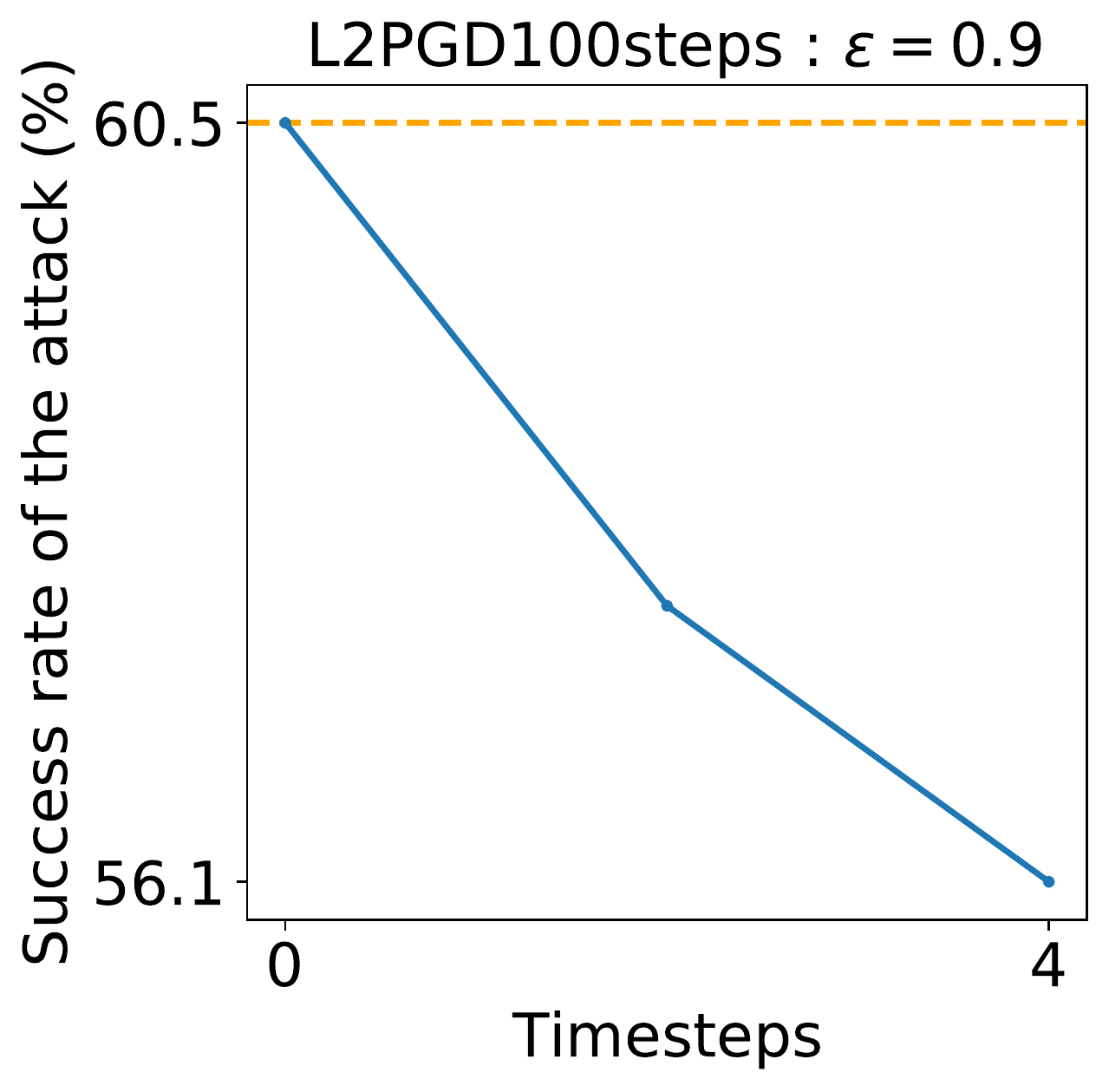}
    \end{subfigure}
    \begin{subfigure}{0.23\textwidth}
    \includegraphics[scale=0.23]{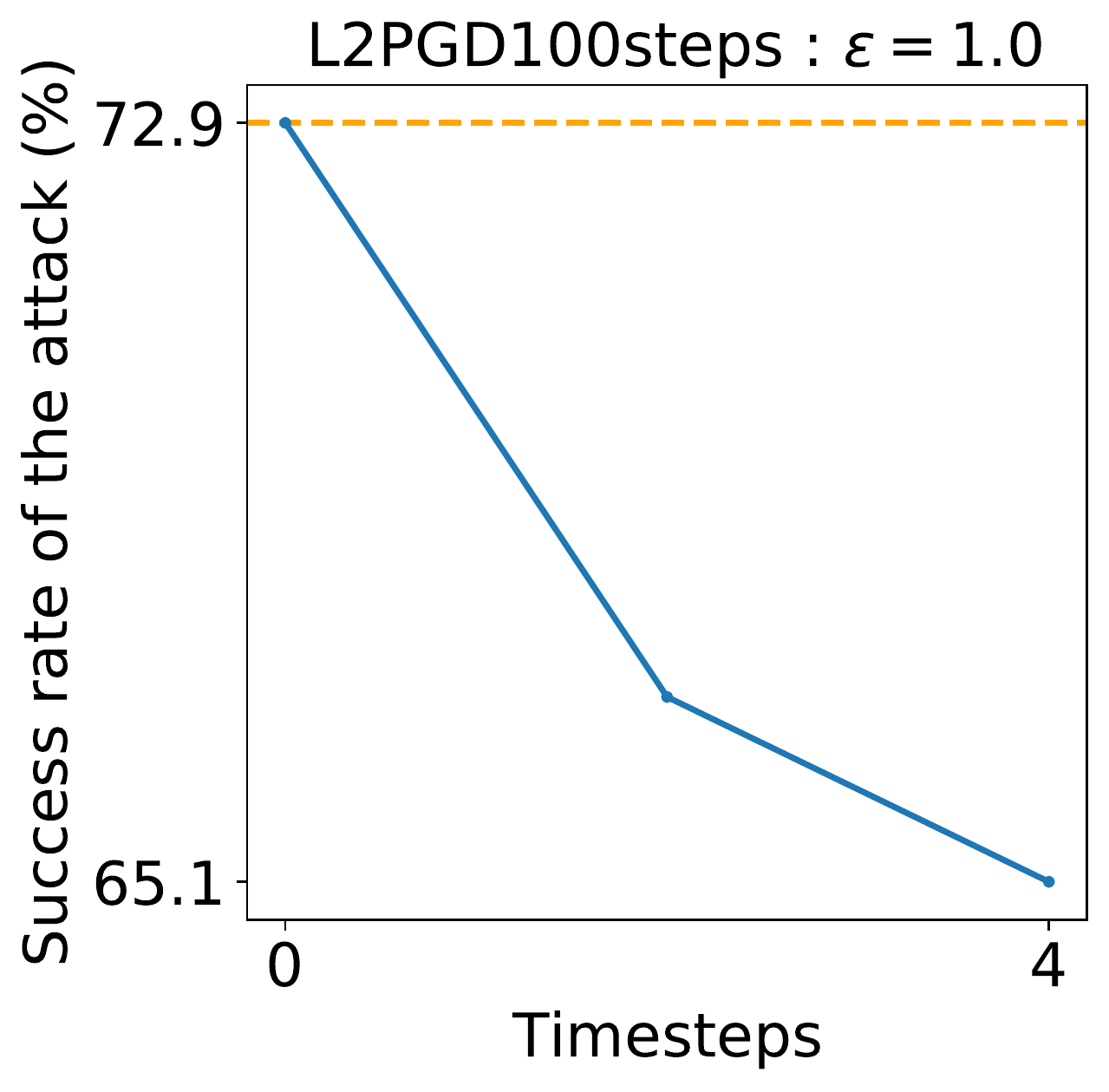}
    \end{subfigure}
    \caption{$L_2$RPGD attacks on PEfficientNetB0 network}
    
\end{figure}
\begin{figure}[h!]
    \centering
    \includegraphics[width=0.7\linewidth]{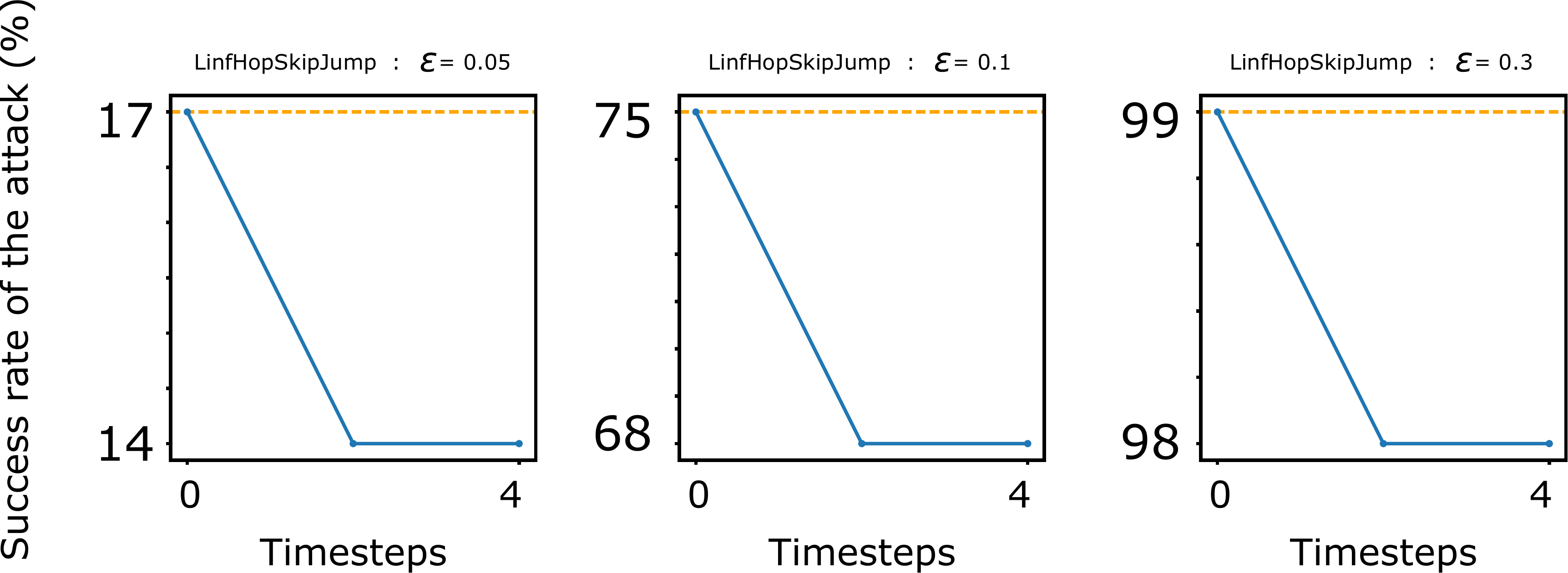}
    \caption{$L_\infty$ HopSkipJump attacks on PEfficientNetB0}
    \label{fig:my_label}
\end{figure}

\begin{figure}[t!]
    \centering
    \includegraphics[width=\linewidth]{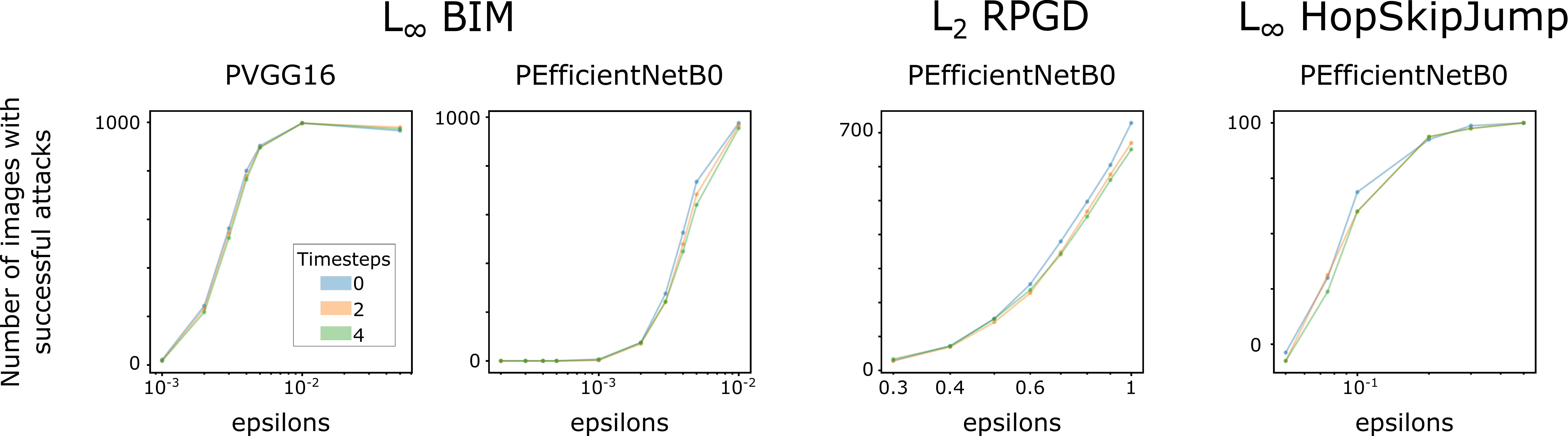}
    \caption{Adversarial Attacks with respect to epsilons. Here we show the number of successful attacks on 1000 (100 for HopSkipJump) images. Increasing the size of the epsilon leads to increase in the success rate of the attack as expected. As predictive coding timesteps increase, the curves shift slightly to the right, meaning that a slightly larger perturbation is required to fool the network. This robustness is more easily seen on Figure~\ref{fig:robustness_adversarial_attacks}, where $\epsilon$ values are sampled near each curve's inflection point. }
    \label{fig:aapdx_advattacks_old_fashion}
\end{figure}

\clearpage
\subsection{Absolute values of the plots shown in the main text}
\label{apndx:abs_val}

\begin{table}[h!]
    \centering
    \begin{tabular}{c|c c | c c}
        \toprule
        Noise Level & \multicolumn{2}{c}{ PVGG16 } & \multicolumn{2}{c}{PEfficientNetB0} \\
        \midrule
         & Accuracy at t=0 & Accuracy at t=15 & Accuracy at t=0 & Accuracy at t=15 \\
        \hline
        $\sigma=0.00$ & 71.63  & 71.47 & 77.29 & 75.35 \\
        $\sigma=0.50$ & 35.61 & 38.59 & 57.66 & 56.24 \\
        $\sigma=0.75$ & 16.69 & 18.46 & 37.11 & 41.05 \\
        $\sigma=1.00$ & 5.59 & 7.05 & 17.03  & 23.59 \\
        \bottomrule
    \end{tabular}
    \caption{Accuracy on gaussian noise-corrupted images. Here we show the accuracy obtained on images corrupted using gaussian noise (at t=0) as shown in figure 2a. All the values are calculated on the corrupted versions of the ImageNet validation dataset.}
    \label{tab:absolute_values}
\end{table}

\begin{table}[h!]
    \centering
    \begin{tabular}{c|c c | c c}
        \toprule
        Noise Level & \multicolumn{2}{c}{ PVGG16 } & \multicolumn{2}{c}{PEfficientNetB0} \\
        \midrule
         & MSE at t=0 & MSE at t=15 & MSE at t=0 & MSE at t=15 \\
        \hline
        $\sigma=0.00$ & 0.224 & 0.220  & 0.186 & 0.184 \\
        $\sigma=0.25$ & 0.342   & 0.324  &0.223 & 0.222 \\
        $\sigma=0.50$ & 0.518  & 0.485  & 0.303 & 0.302 \\
        $\sigma=0.75$ & 0.705  & 0.660  & 0.394 & 0.392  \\
        $\sigma=1.00$ & 0.898  & 0.842  & 0.486 & 0.482   \\
        $\sigma=2.00$ & 1.689  & 1.587  & 0.848 & 0.834 \\
        \bottomrule
    \end{tabular}
    \caption{MSE distances for reconstructions on noisy images. Here we show the MSE distances obtained between the noisy images corrupted using gaussian noises and the reconstructions made by the models as shown in Figure 2b. }
    \label{tab:mse_values_absolute}
\end{table}

\begin{figure}[h!]
    \centering
    \includegraphics[width=\linewidth]{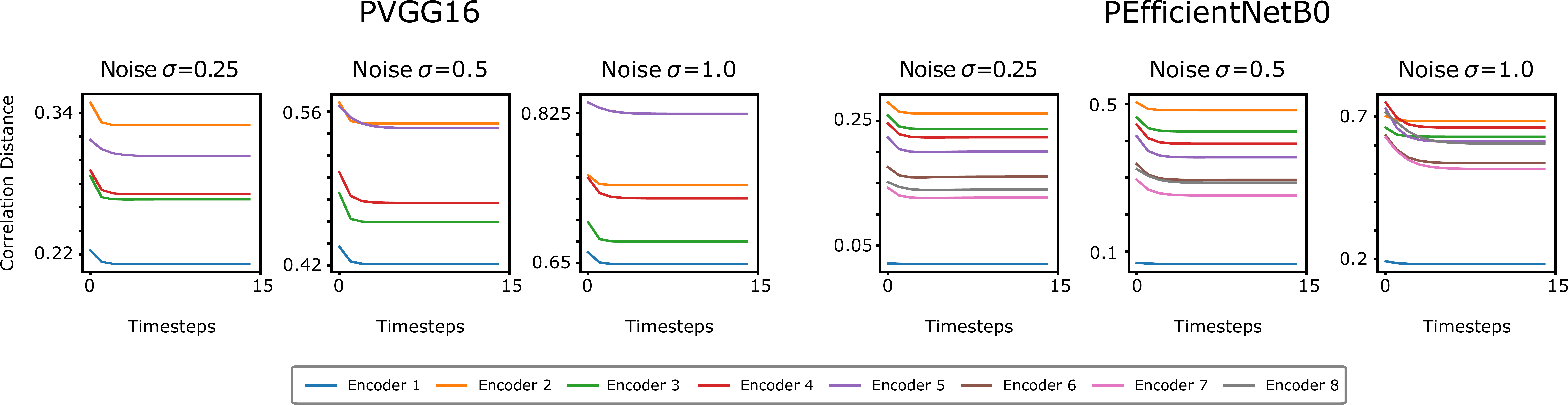}
    \caption{\textbf{Correlation distances for representations obtained on noisy images:} Here we show the absolute correlation distances obtained between clean and noisy representations as shown in Figure 2d in the main text.}
    \label{fig:absolute_corr}
\end{figure}

\end{document}